\documentclass[runningheads]{llncs}

\usepackage[footnotesize]{subfig}
\usepackage{times}
\usepackage{epsfig}
\usepackage{graphicx}
\usepackage{amsmath,amssymb} % define this before the line numbering.
\usepackage{booktabs}
\usepackage{tabulary}
\usepackage{multirow}
\usepackage{url}
\usepackage{footnote}
\usepackage{tablefootnote}
\usepackage[10pt]{moresize}
\usepackage[hang]{footmisc}
\usepackage{pifont}

\usepackage{color}
\usepackage[width=122mm,left=12mm,paperwidth=146mm,height=193mm,top=12mm,paperheight=217mm]{geometry}

\usepackage{array}
\newcolumntype{C}[1]{>{\centering\let\newline\\\arraybackslash\hspace{0pt}}m{#1}}

\def\etal{\emph{et al.}}

\begin{document}
\pagestyle{headings}
\mainmatter

\def\httilde{\mbox{\tt\raisebox{-.5ex}{\symbol{126}}}}

\newcommand{\specialcell}[2][c]{%
  \begin{tabular}[#1]{@{}l@{}}#2\end{tabular}}

%%%%%%%%% TITLE
\title{SSD: Single Shot MultiBox Detector}

\titlerunning{SSD: Single Shot MultiBox Detector}

\authorrunning{Liu \etal}
% \authorrunning{Wei Liu, Dragomir Anguelov, Dumitru Erhan, Christian Szegedy, Scott Reed, Cheng-Yang Fu, Alexander C. Berg}

\author{Wei Liu$^1$, Dragomir Anguelov$^2$, Dumitru Erhan$^3$, Christian Szegedy$^3$,\\Scott Reed$^4$, Cheng-Yang Fu$^1$, Alexander C. Berg$^1$}
\institute{$^1$UNC Chapel Hill $^2$Zoox Inc. $^3$Google Inc. $^4$University of Michigan, Ann-Arbor\\
{\tt\small $^1$wliu@cs.unc.edu, $^2$drago@zoox.com, $^3$\{dumitru,szegedy\}@google.com, $^4$reedscot@umich.edu, $^1$\{cyfu,aberg\}@cs.unc.edu}}

\maketitle

%%%%%%%%% ABSTRACT
\begin{abstract}
We present a method for detecting objects in images using a single deep neural network. Our approach, named SSD, discretizes the output space of bounding boxes into a set of default boxes over different aspect ratios and scales per feature map location. At prediction time, the network generates scores for the presence of each object category in each default box and produces adjustments to the box to better match the object shape. Additionally, the network combines predictions from multiple feature maps with different resolutions to naturally handle objects of various sizes. SSD is simple relative to methods that require object proposals because it completely eliminates proposal generation and subsequent pixel or feature resampling stages and encapsulates all computation in a single network. This makes SSD easy to train and straightforward to integrate into systems that require a detection component. Experimental results on the PASCAL VOC, COCO, and ILSVRC datasets confirm that SSD has competitive accuracy to methods that utilize an additional object proposal step and is much faster, while providing a unified framework for both training and inference. For $300\times 300$ input, SSD achieves 74.3\% mAP\footnote{We achieved even better results using an improved data augmentation scheme in follow-on experiments: 77.2\% mAP for $300\times 300$ input and 79.8\% mAP for $512 \times 512$ input on VOC2007. Please see Sec.~\ref{sec:data-aug-new} for details.} on VOC2007 \texttt{test} at 59 FPS on a Nvidia Titan X and for $512\times 512$ input, SSD achieves 76.9\% mAP, outperforming a comparable state-of-the-art Faster R-CNN model. Compared to other single stage methods, SSD has much better accuracy even with a smaller input image size. Code is available at: \url{https://github.com/weiliu89/caffe/tree/ssd} .
\end{abstract}

%%%%%%%%% BODY TEXT
\section{Introduction}
Current state-of-the-art object detection systems are variants of the following approach: hypothesize bounding boxes, resample pixels or features for each box, and apply a high-quality classifier. This pipeline has prevailed on detection benchmarks since the Selective Search work~\cite{uijlings2013selective} through the current leading results on PASCAL VOC, COCO, and ILSVRC detection all based on Faster R-CNN\cite{ren2015faster} albeit with deeper features such as \cite{kaiming2015resiual}. While accurate, these approaches have been too computationally intensive for embedded systems and, even with high-end hardware, too slow for real-time applications. Often detection speed for these approaches is measured in seconds per frame (SPF), and even the fastest high-accuracy detector, Faster R-CNN, operates at only 7 frames per second (FPS). There have been many attempts to build faster detectors by attacking each stage of the detection pipeline (see related work in Sec.~\ref{sec:relatedwork}), but so far, significantly increased speed comes only at the cost of significantly decreased detection accuracy.

This paper presents the first deep network based object detector that does not resample pixels or features for bounding box hypotheses {\em and} and is as accurate as approaches that do. This results in a significant improvement in speed for high-accuracy detection (59 FPS with mAP 74.3\% on VOC2007 \texttt{test}, vs. Faster R-CNN 7 FPS with mAP 73.2\% or YOLO 45 FPS with mAP 63.4\%). The fundamental improvement in speed comes from eliminating bounding box proposals and the subsequent pixel or feature resampling stage. We are not the first to do this (cf \cite{sermanet2013overfeat,redmon2015you}), but by adding a series of improvements, we manage to increase the accuracy significantly over previous attempts. Our improvements include using a small convolutional filter to predict object categories and offsets in bounding box locations, using separate predictors (filters) for different aspect ratio detections, and applying these filters to multiple feature maps from the later stages of a network in order to perform detection at multiple scales. With these modifications---especially using multiple layers for prediction at different scales---we can achieve high-accuracy using relatively low resolution input, further increasing detection speed. While these contributions may seem small independently, we note that the resulting system improves accuracy on real-time detection for PASCAL VOC from 63.4\% mAP for YOLO to 74.3\% mAP for our SSD. This is a larger relative improvement in detection accuracy than that from the recent, very high-profile work on residual networks~\cite{kaiming2015resiual}. Furthermore, significantly improving the speed of high-quality detection can broaden the range of settings where computer vision is useful.

We summarize our contributions as follows:
\begin{itemize}
\item We introduce SSD, a single-shot detector for multiple categories that is faster than the previous state-of-the-art for single shot detectors (YOLO), and significantly more accurate, in fact as accurate as slower techniques that perform explicit region proposals and pooling (including Faster R-CNN).

\item The core of SSD is predicting category scores and box offsets for a fixed set of default bounding boxes using small convolutional filters applied to feature maps.

\item To achieve high detection accuracy we produce predictions of different scales from feature maps of different scales, and explicitly separate predictions by aspect ratio.

\item These design features lead to simple end-to-end training and high accuracy, even on low resolution input images, further improving the speed vs accuracy trade-off.

\item Experiments include timing and accuracy analysis on models with varying input size evaluated on PASCAL VOC, COCO, and ILSVRC and are compared to a range of recent state-of-the-art approaches.
\end{itemize}

\begin{figure}[tbp]
    \centering
    \includegraphics[width=0.9\linewidth]{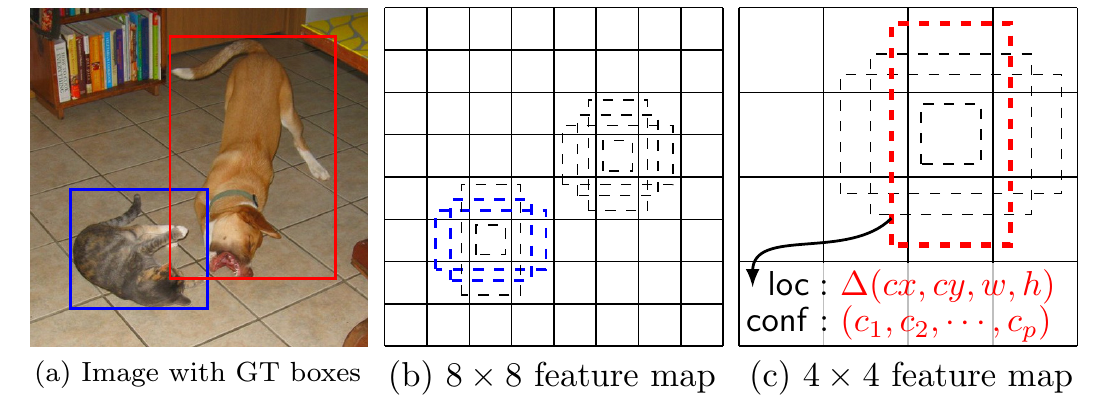}
    \caption{\textbf{SSD framework.} (a) SSD only needs an input image and ground truth boxes for each object during training. In a convolutional fashion, we evaluate a small set (e.g. 4) of default boxes of different aspect ratios at each location in several feature maps with different scales (e.g. $8\times 8$ and $4\times 4$ in (b) and (c)). For each default box, we predict both the shape offsets and the confidences for all object categories ($(c_1, c_2, \cdots, c_p)$). At training time, we first match these default boxes to the ground truth boxes. For example, we have matched two default boxes with the cat and one with the dog, which are treated as positives and the rest as negatives. The model loss is a weighted sum between localization loss (e.g. Smooth L1~\cite{girshick2015fast}) and confidence loss (e.g. Softmax).}
    \label{fig:ssdframework}
\end{figure}

%------------------------------------------------------------------------
\section{The Single Shot Detector (SSD)}
\label{sec:ssd}

This section describes our proposed SSD framework for detection (Sec.~\ref{sec:ssdmodel}) and the associated training methodology (Sec.~\ref{sec:ssdtraining}). Afterwards, Sec.~\ref{sec:exp} presents dataset-specific model details and experimental results.

\subsection{Model}
\label{sec:ssdmodel}

The SSD approach is based on a feed-forward convolutional network that produces a fixed-size collection of bounding boxes and scores for the presence of object class instances in those boxes, followed by a non-maximum suppression step to produce the final detections. The early network layers are based on a standard architecture used for high quality image classification (truncated before any classification layers), which we will call the base network\footnote{We use the VGG-16 network as a base, but other networks should also produce good results.}. We then add auxiliary structure to the network to produce detections with the following key features:

\begin{figure}
	\centering
	\includegraphics[width=0.8\linewidth]{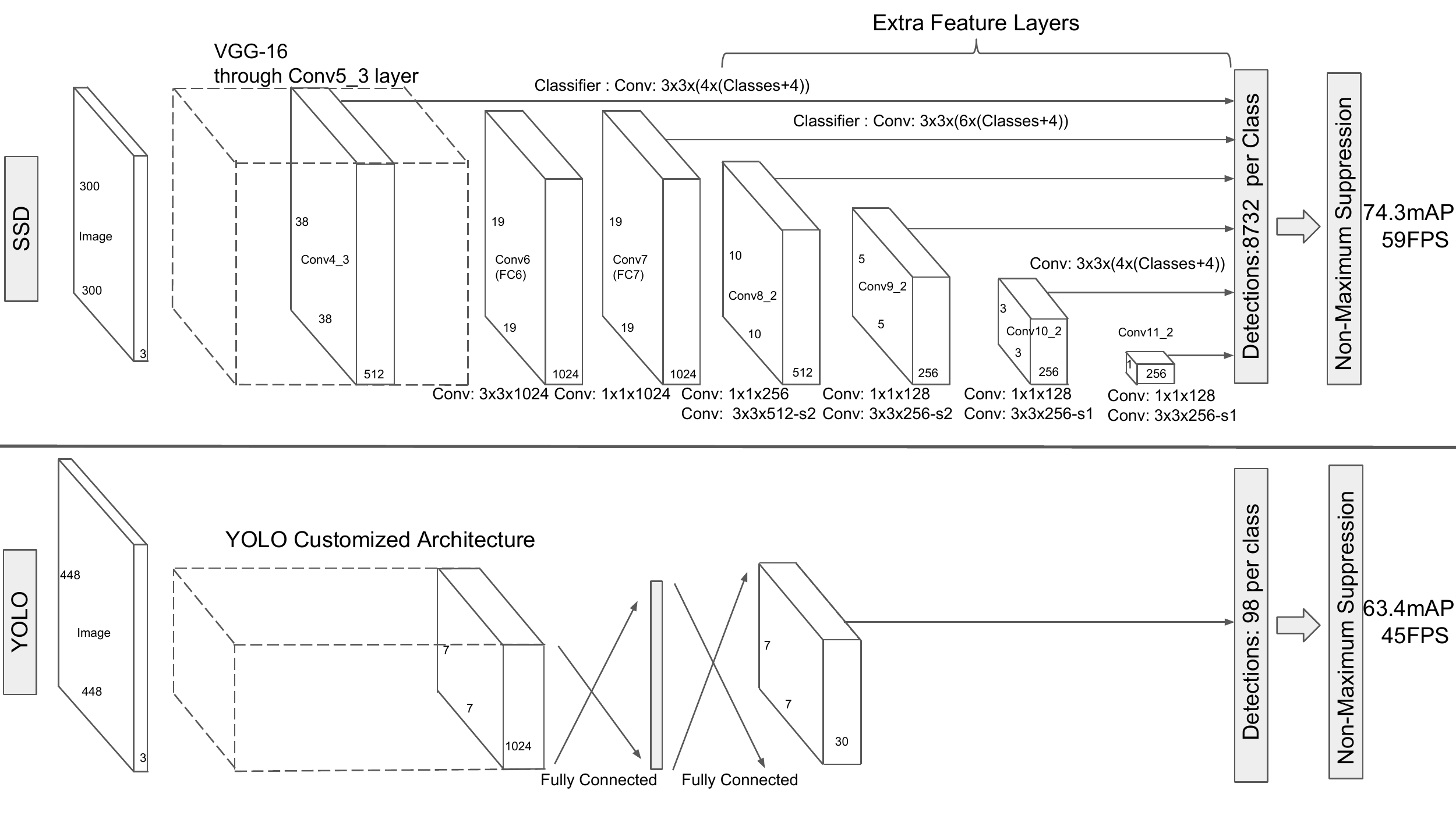}
    \caption{A comparison between two single shot detection models: SSD and YOLO~\cite{redmon2015you}. Our SSD model adds several feature layers to the end of a base network, which predict the offsets to default boxes of different scales and aspect ratios and their associated confidences. SSD with a $300\times 300$ input size significantly outperforms its $448\times 448$ YOLO counterpart in accuracy on VOC2007 \texttt{test} while also improving the speed.}
    \label{fig:architecture}
\end{figure}

\smallskip
\noindent {\bf Multi-scale feature maps for detection} We add convolutional feature layers to the end of the truncated base network. These layers decrease in size progressively and allow predictions of detections at multiple scales.  The convolutional model for predicting detections is different for each feature layer ({\em cf} Overfeat\cite{sermanet2013overfeat} and YOLO\cite{redmon2015you} that operate on a single scale feature map).

\smallskip
\noindent {\bf Convolutional predictors for detection} Each added feature layer (or optionally an existing feature layer from the base network) can produce a fixed set of detection predictions using a set of convolutional filters. These are indicated on top of the SSD network architecture in Fig.~\ref{fig:architecture}.  For a feature layer of size $m \times n$ with $p$ channels, the basic element for predicting parameters of a potential detection is a $3 \times 3 \times p$ {\em small kernel} that produces either a score for a category, or a shape offset relative to the default box coordinates. At each of the $m \times n$ locations where the kernel is applied, it produces an output value. The bounding box offset output values are measured relative to a default box position relative to each feature map location ({\em cf} the architecture of YOLO\cite{redmon2015you} that uses an intermediate fully connected layer instead of a convolutional filter for this step).

\smallskip
\noindent {\bf Default boxes and aspect ratios} We associate a set of default bounding boxes with each feature map cell, for multiple feature maps at the top of the network. The default boxes tile the feature map in a convolutional manner, so that the position of each box relative to its corresponding cell is fixed. At each feature map cell, we predict the offsets relative to the default box shapes in the cell, as well as the per-class scores that indicate the presence of a class instance in each of those boxes. Specifically, for each box out of $k$ at a given location, we compute $c$ class scores and the $4$ offsets relative to the original default box shape. This results in a total of $(c+4)k$ filters that are applied around each location in the feature map, yielding $(c+4)kmn$ outputs for a $m\times n$ feature map. For an illustration of default boxes, please refer to Fig.~\ref{fig:ssdframework}. Our default boxes are similar to the {\it anchor boxes} used in Faster R-CNN~\cite{ren2015faster}, however we apply them to several feature maps of different resolutions. Allowing different default box shapes in several feature maps let us efficiently discretize the space of possible output box shapes.

\subsection{Training}
\label{sec:ssdtraining}

The key difference between training SSD and training a typical detector that uses region proposals, is that ground truth information needs to be assigned to specific outputs in the fixed set of detector outputs. Some version of this is also required for training in YOLO\cite{redmon2015you} and for the region proposal stage of Faster R-CNN\cite{ren2015faster} and MultiBox\cite{erhan2014scalable}. Once this assignment is determined, the loss function and back propagation are applied end-to-end. Training also involves choosing the set of default boxes and scales for detection as well as the hard negative mining and data augmentation strategies.

\subsubsection{Matching strategy}
\label{sec:matchingstrategy}
During training we need to determine which default boxes correspond to a ground truth detection and train the network accordingly.  For each ground truth box we are selecting from default boxes that vary over location, aspect ratio, and scale. We begin by matching each ground truth box to the default box with the best jaccard overlap (as in MultiBox~\cite{erhan2014scalable}). Unlike MultiBox, we then match default boxes to any ground truth with jaccard overlap higher than a threshold (0.5). This simplifies the learning problem, allowing the network to predict high scores for multiple overlapping default boxes rather than requiring it to pick only the one with maximum overlap.

\subsubsection{Training objective}
\label{sec:trainingobjective}
The SSD training objective is derived from the MultiBox objective~\cite{erhan2014scalable,szegedy2014scalable} but is extended to handle multiple object categories. Let $x_{ij}^p = \{1,0\}$ be an indicator for matching the $i$-th default box to the $j$-th ground truth box of category $p$.
%, and $x_{ij}^p = 0$ otherwise. 
In the matching strategy above, we can have $\sum_i x_{ij}^p \geq 1$.
%, meaning there can be more than one default box matched to the $j$-th ground truth box. 
The overall objective loss function is a weighted sum of the localization loss (loc) and the confidence loss (conf):
\begin{equation}
L(x, c, l, g) = \frac{1}{N}(L_{conf}(x, c) + \alpha L_{loc}(x, l, g))
\label{eq:loss}
\end{equation}
where N is the number of matched default boxes. If $N = 0$, wet set the loss to 0. The localization loss is a Smooth L1 loss~\cite{girshick2015fast} between the predicted box ($l$) and the ground truth box ($g$) parameters. Similar to Faster R-CNN~\cite{ren2015faster}, we regress to offsets for the center ($cx, cy$) of the default bounding box ($d$) and for its width ($w$) and height ($h$).
\begin{equation}
\begin{split}
L_{loc}(x,l,g) = \sum_{i\in Pos}^N \sum_{m\in\{cx, cy, w, h\}}& x_{ij}^k\text{smooth}_{\text{L1}}(l_i^m - \hat{g}_j^m)\\
\hat{g}_j^{cx} = (g_j^{cx} - d_i^{cx}) / d_i^w\quad\quad&
\hat{g}_j^{cy} = (g_j^{cy} - d_i^{cy}) / d_i^h\\
\hat{g}_j^{w} = \log\Big(\frac{g_j^{w}}{d_i^w}\Big)\quad\quad&
\hat{g}_j^{h} = \log\Big(\frac{g_j^{h}}{d_i^h}\Big)
\end{split}
\label{eq:locloss}
\end{equation}
%%\begin{equation}
%%L_{loc}(x, l, g) = \frac{1}{2}\sum_{i,j} x_{ij}^p||l_i - g_j^p||_2^2
%%\end{equation}
The confidence loss is the softmax loss over multiple classes confidences ($c$).
\begin{equation}
L_{conf}(x, c) = - \sum_{i\in Pos}^N x_{ij}^p log(\hat{c}_i^p) - \sum_{i\in Neg} log(\hat{c}_i^0)\quad\text{where}\quad\hat{c}_i^p = \frac{\exp(c_i^p)}{\sum_p \exp(c_i^p)}
\label{eq:confloss}
\end{equation}
and the weight term $\alpha$ is set to 1 by cross validation.
\subsubsection{Choosing scales and aspect ratios for default boxes}
\label{sec:defaultboxes}
To handle different object scales, some methods~\cite{sermanet2013overfeat,he2014spatial} suggest processing the image at different sizes and combining the results afterwards. However, by utilizing feature maps from several different layers in a single network for prediction we can mimic the same effect, while also sharing parameters across all object scales. Previous works~\cite{long2014fully,hariharan2014hypercolumns} have shown that using feature maps from the lower layers can improve semantic segmentation quality because the lower layers capture more fine details of the input objects. Similarly, ~\cite{liu2015parsenet} showed that adding global context pooled from a feature map can help smooth the segmentation results. Motivated by these methods, we use both the lower and upper feature maps for detection. Figure~\ref{fig:ssdframework} shows two exemplar feature maps ($8\times 8$ and $4\times 4$) which are used in the framework. In practice, we can use many more with small computational overhead.

Feature maps from different levels within a network are known to have different (empirical) receptive field sizes~\cite{zhou2014object}. Fortunately, within the SSD framework, the default boxes do not necessary need to correspond to the actual receptive fields of each layer.
We design the tiling of default boxes so that specific feature maps learn to be responsive to particular scales of the objects. Suppose we want to use $m$ feature maps for prediction. The scale of the default boxes for each feature map is computed as:
\begin{equation}
s_k = s_\text{min} + \frac{s_\text{max} - s_\text{min}}{m - 1} (k - 1),\quad k\in [1, m]
\end{equation}
where $s_\text{min}$ is 0.2 and $s_\text{max}$ is 0.9, meaning the lowest layer has a scale of 0.2 and the highest layer has a scale of 0.9, and all layers in between are regularly spaced. We impose different aspect ratios for the default boxes, and denote them as $a_r \in \{1, 2, 3, \frac{1}{2}, \frac{1}{3}\}$. We can compute the width ($w_k^a = s_k\sqrt{a_r}$) and height ($h_k^a = s_k / \sqrt{a_r}$) for each default box. For the aspect ratio of 1, we also add a default box whose scale is $s'_k = \sqrt{s_k s_{k+1}}$, resulting in 6 default boxes per feature map location. We set the center of each default box to $(\frac{i+0.5}{|f_k|}, \frac{j+0.5}{|f_k|})$, where $|f_k|$ is the size of the $k$-th square feature map, $i, j\in [0, |f_k|)$. In practice, one can also design a distribution of default boxes to best fit a specific dataset. How to design the optimal tiling is an open question as well.

By combining predictions for all default boxes with different scales and aspect ratios from all locations of many feature maps, we have a diverse set of predictions, covering various input object sizes and shapes. For example, in Fig.~\ref{fig:ssdframework}, the dog is matched to a default box in the $4\times 4$ feature map, but not to any default boxes in the $8\times 8$ feature map. This is because those boxes have different scales and do not match the dog box, and therefore are considered as negatives during training.
\vspace{-1em}
\subsubsection{Hard negative mining}
\label{sec:hardnegative}
After the matching step, most of the default boxes are negatives, especially when the number of possible default boxes is large. This introduces a significant imbalance between the positive and negative training examples. Instead of using all the negative examples, we sort them using the highest confidence loss for each default box and pick the top ones so that the ratio between the negatives and positives is at most 3:1. We found that this leads to faster optimization and a more stable training.
\vspace{-1em}
\subsubsection{Data augmentation}
\label{sec:dataaugmentation}
To make the model more robust to various input object sizes and shapes, each training image is randomly sampled by one of the following options:
\begin{itemize}
\setlength\itemsep{0em}
\item Use the entire original input image.
\item Sample a patch so that the \emph{minimum} jaccard overlap with the objects is 0.1, 0.3, 0.5, 0.7, or 0.9.
\item Randomly sample a patch.
\end{itemize}
The size of each sampled patch is [0.1, 1] of the original image size, and the aspect ratio is between $\frac{1}{2}$ and 2. We keep the overlapped part of the ground truth box if the center of it is in the sampled patch. After the aforementioned sampling step, each sampled patch is resized to fixed size and is horizontally flipped with probability of 0.5, in addition to applying some photo-metric distortions similar to those described in~\cite{howard2013some}.

%------------------------------------------------------------------------
\section{Experimental Results} \label{sec:exp}
\subsubsection{Base network} Our experiments are all based on VGG16~\cite{simonyan2014very}, which is pre-trained on the ILSVRC CLS-LOC dataset~\cite{russakovsky2014imagenet}. Similar to DeepLab-LargeFOV~\cite{chen2014semantic}, we convert fc6 and fc7 to convolutional layers, subsample parameters from fc6 and fc7, change pool5 from $2\times 2-s2$ to $3\times 3-s1$, and use the \emph{\`{a} trous} algorithm~\cite{holschneider1990real} to fill the "holes". We remove all the dropout layers and the fc8 layer. We fine-tune the resulting model using SGD with initial learning rate $10^{-3}$, 0.9 momentum, 0.0005 weight decay, and batch size 32. The learning rate decay policy is slightly different for each dataset, and we will describe details later. The full training and testing code is built on Caffe~\cite{jia2014caffe} and is open source at: \url{https://github.com/weiliu89/caffe/tree/ssd} .
%We also provide various SSD models trained on different datasets.

\subsection{PASCAL VOC2007}
On this dataset, we compare against Fast R-CNN~\cite{girshick2015fast} and Faster R-CNN~\cite{ren2015faster} on VOC2007 \texttt{test} (4952 images). All methods fine-tune on the same pre-trained VGG16 network.

Figure~\ref{fig:architecture} shows the architecture details of the SSD300 model. We use conv4\_3, conv7 (fc7), conv8\_2, conv9\_2, conv10\_2, and conv11\_2 to predict both location and confidences. We set default box with scale 0.1 on conv4\_3\footnote{For SSD512 model, we add extra conv12\_2 for prediction, set $s_\text{min}$ to 0.15, and 0.07 on conv4\_3.}. We initialize the parameters for all the newly added convolutional layers with the "xavier" method~\cite{glorot2010understanding}. For conv4\_3, conv10\_2 and conv11\_2, we only associate 4 default boxes at each feature map location -- omitting aspect ratios of $\frac{1}{3}$ and 3. For all other layers, we put 6 default boxes as described in Sec.~\ref{sec:defaultboxes}. Since, as pointed out in~\cite{liu2015parsenet}, conv4\_3 has a different feature scale compared to the other layers, we use the L2 normalization technique introduced in~\cite{liu2015parsenet} to scale the feature norm at each location in the feature map to 20 and learn the scale during back propagation. We use the $10^{-3}$ learning rate for 40k iterations, then continue training for 10k iterations with $10^{-4}$ and $10^{-5}$. When training on VOC2007 \texttt{trainval}, Table~\ref{tab:voc07} shows that our low resolution SSD300 model is already more accurate than Fast R-CNN. When we train SSD on a larger $512\times 512$ input image, it is even more accurate, surpassing Faster R-CNN by 1.7\% mAP. If we train SSD with more (i.e. 07+12) data, we see that SSD300 is already better than Faster R-CNN by 1.1\% and that SSD512 is 3.6\% better. If we take models trained on COCO \texttt{trainval35k} as described in Sec.~\ref{sec:expcoco} and fine-tuning them on the 07+12 dataset with SSD512, we achieve the best results: 81.6\% mAP.
\begin{table}[ht]\ssmall
	\centering
	\setlength{\tabcolsep}{1.55pt}
	\begin{tabular*}{\textwidth}{l|c|c|cccccccccccccccccccc}
		%\hline
		%\noalign{\smallskip}
		\tiny Method & data & \tiny mAP & \tiny aero & \tiny bike & \tiny bird & \tiny boat & \tiny bottle & \tiny bus & \tiny car & \tiny cat & \tiny chair & \tiny cow & \tiny table & \tiny dog & \tiny horse & \tiny mbike & \tiny person & \tiny plant & \tiny sheep & \tiny sofa & \tiny train & \tiny tv \\
        \hline
		%\noalign{\smallskip}
        \tiny Fast~\cite{girshick2015fast} & 07 & 66.9 & 74.5 & 78.3 & 69.2 & 53.2 & 36.6 & 77.3 & 78.2 & 82.0 & 40.7 & 72.7 & 67.9 & 79.6 & 79.2 & 73.0 & 69.0 & 30.1 & 65.4 & 70.2 & 75.8 & 65.8\\
		\tiny Fast~\cite{girshick2015fast} & 07+12 & 70.0 & 77.0 & 78.1 & 69.3 & 59.4 & 38.3 & 81.6 & 78.6 & 86.7 & 42.8 & 78.8 & 68.9 & 84.7 & 82.0 & 76.6 & 69.9 & 31.8 & 70.1 & 74.8 & 80.4 & 70.4\\
        \tiny Faster~\cite{ren2015faster} & 07 & 69.9 & 70.0 & 80.6 & 70.1 & 57.3 & 49.9 & 78.2 & 80.4 & 82.0 & 52.2 & 75.3 & 67.2 & 80.3 & 79.8 & 75.0 & 76.3 & 39.1 & 68.3 & 67.3 & 81.1 & 67.6\\
		\tiny Faster~\cite{ren2015faster} & 07+12 & 73.2 & 76.5 & 79.0 & 70.9 & 65.5 & 52.1 & 83.1 & 84.7 & 86.4 & 52.0 & 81.9 & 65.7 & 84.8 & 84.6 & 77.5 & 76.7 & 38.8 & 73.6 & 73.9 & 83.0 & 72.6\\
        \tiny Faster~\cite{ren2015faster} & \tiny 07+12+COCO & 78.8 & 84.3 & 82.0 & 77.7 & 68.9 & 65.7 & 88.1 & 88.4 & 88.9 & 63.6 & 86.3 & 70.8 & 85.9 & 87.6 & 80.1 & 82.3 & 53.6 & 80.4 & 75.8 & 86.6 & 78.9\\
        %\tiny ION~\cite{bell2015inside} & 78.2 & 79.1 & 76.8 & 61.5 & 54.7 & 81.9 & 84.3 & 88.3 & 53.1 & 78.3 & 71.6 & 85.9 & 84.8 & 81.6 & 74.3 & 45.6 & 75.3 & 72.1 & 82.6 & 81.4 & 74.6\\
		\hline
        \tiny SSD300 & 07 & 68.0 & 73.4 & 77.5 & 64.1 & 59.0 & 38.9 & 75.2 & 80.8 & 78.5 & 46.0 & 67.8 & 69.2 & 76.6 & 82.1 & 77.0 & 72.5 & 41.2 & 64.2 & 69.1 & 78.0 & 68.5\\
		\tiny SSD300 & 07+12 & 74.3 & 75.5 & 80.2 & 72.3 & 66.3 & 47.6 & 83.0 & 84.2 & 86.1 & 54.7 & 78.3 & 73.9 & 84.5 & 85.3 & 82.6 & 76.2 & 48.6 & 73.9 & 76.0 & 83.4 & 74.0\\
        \tiny SSD300 & \tiny 07+12+COCO & 79.6 & 80.9 & 86.3 & 79.0 & \textbf{76.2} & 57.6 & 87.3 & 88.2 & 88.6 & 60.5 & 85.4 & \textbf{76.7} & \textbf{87.5} & \textbf{89.2} & 84.5 & 81.4 & 55.0 & 81.9 & \textbf{81.5} & 85.9 & 78.9\\
        \tiny SSD512 & 07 & 71.6 & 75.1 & 81.4 & 69.8 & 60.8 & 46.3 & 82.6 & 84.7 & 84.1 & 48.5 & 75.0 & 67.4 & 82.3 & 83.9 & 79.4 & 76.6 & 44.9 & 69.9 & 69.1 & 78.1 & 71.8\\
        \tiny SSD512 & 07+12 & 76.8 & 82.4 & 84.7 & 78.4 & 73.8 & 53.2 & 86.2 & 87.5 & 86.0 & 57.8 & 83.1 & 70.2 & 84.9 & 85.2 & 83.9 & 79.7 & 50.3 & 77.9 & 73.9 & 82.5 & 75.3\\
        \tiny SSD512 & \tiny 07+12+COCO & \textbf{81.6} & \textbf{86.6} & \textbf{88.3} & \textbf{82.4} & 76.0 & \textbf{66.3} & \textbf{88.6} & \textbf{88.9} & \textbf{89.1} & \textbf{65.1} & \textbf{88.4} & 73.6 & 86.5 & 88.9 & \textbf{85.3} & \textbf{84.6} & \textbf{59.1} & \textbf{85.0} & 80.4 & \textbf{87.4} & \textbf{81.2}\\
		%\bottomrule
        \noalign{\smallskip}
	\end{tabular*}
	\caption{\textbf{PASCAL VOC2007 \texttt{test} detection results.} Both Fast and Faster R-CNN use input images whose minimum dimension is 600. The two SSD models have exactly the same settings except that they have different input sizes ($300\times 300$ vs. $512\times 512$). It is obvious that larger input size leads to better results, and more data always helps. Data: "07": VOC2007 \texttt{trainval}, "07+12": union of VOC2007 and VOC2012 \texttt{trainval}. "07+12+COCO": first train on COCO \texttt{trainval35k} then fine-tune on 07+12.}
    \label{tab:voc07}
\end{table}

\begin{figure}[tbp]
	\centering
    \includegraphics[width=0.285\linewidth]{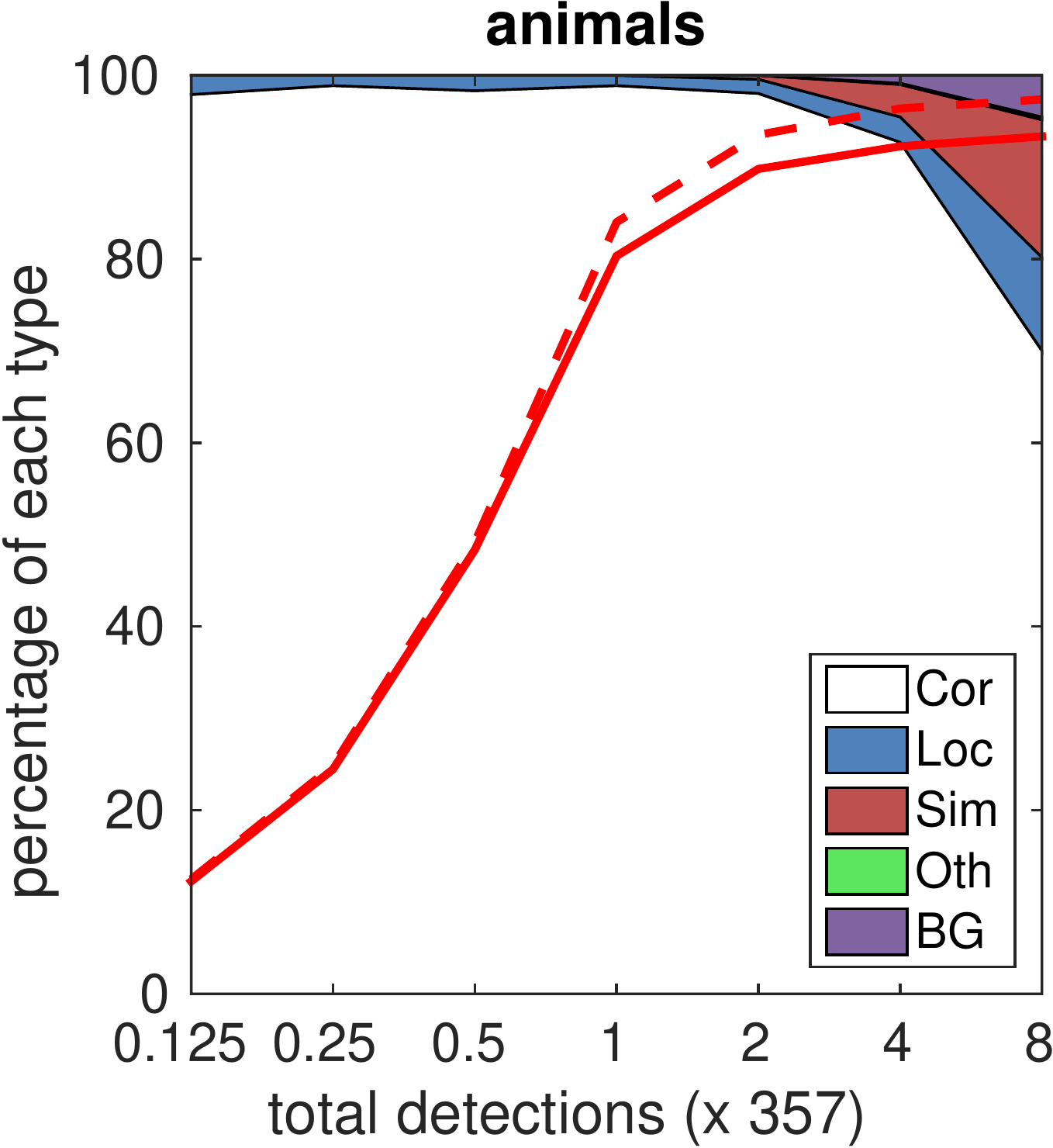}
    \hspace{0.4cm}
    \includegraphics[width=0.285\linewidth]{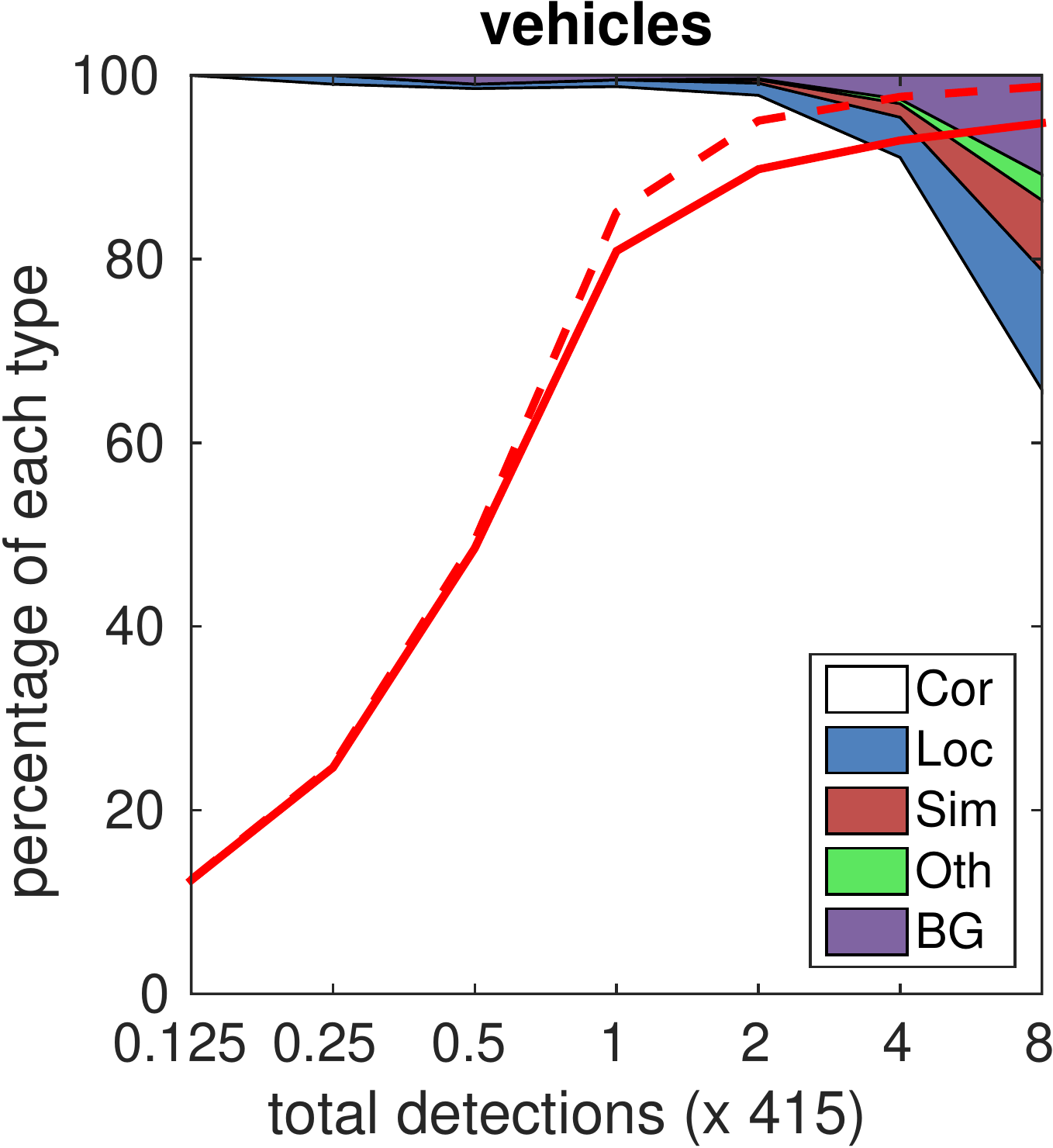}
    \hspace{0.4cm}
    \includegraphics[width=0.285\linewidth]{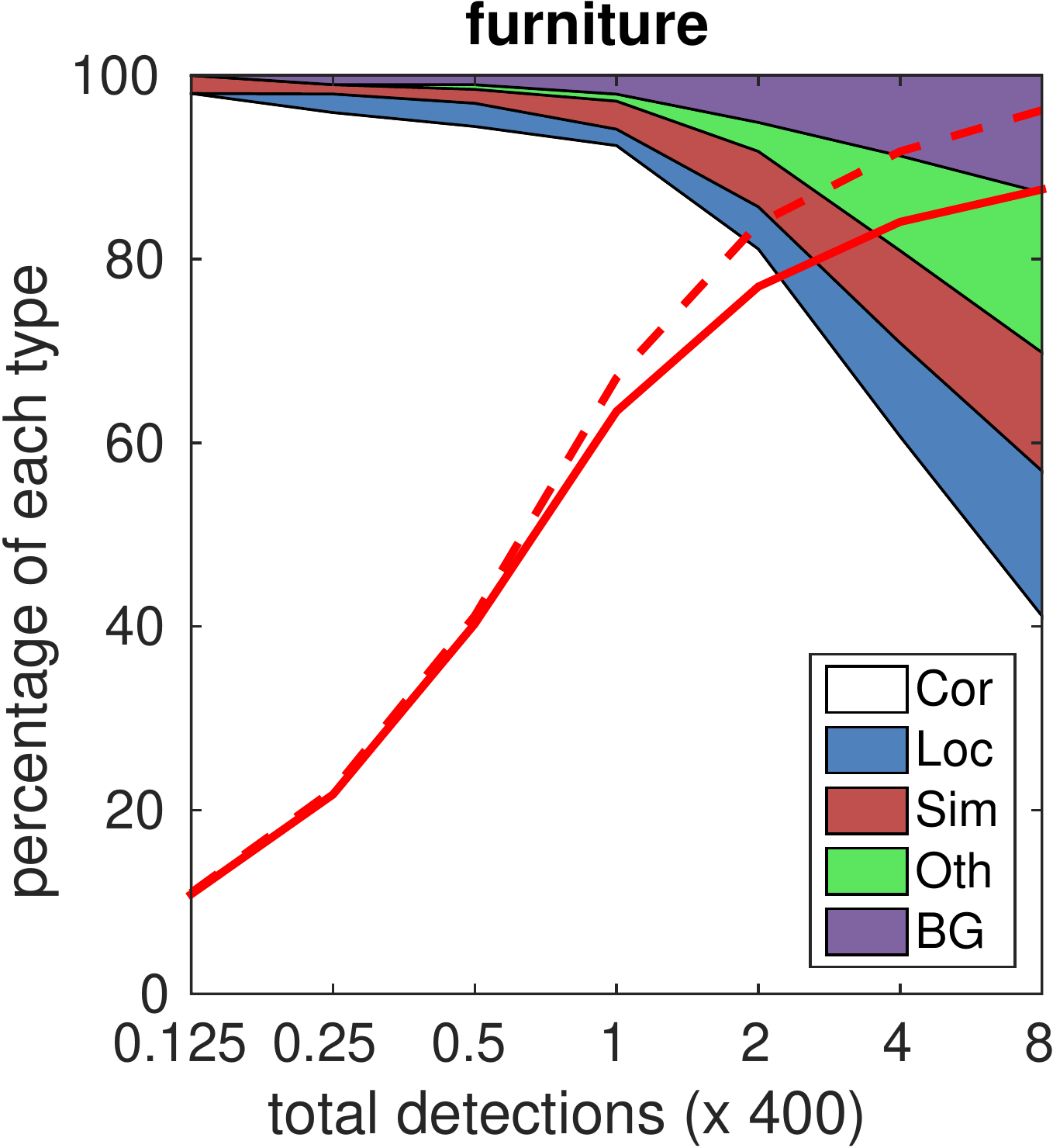}\\
    \includegraphics[width=0.285\linewidth]{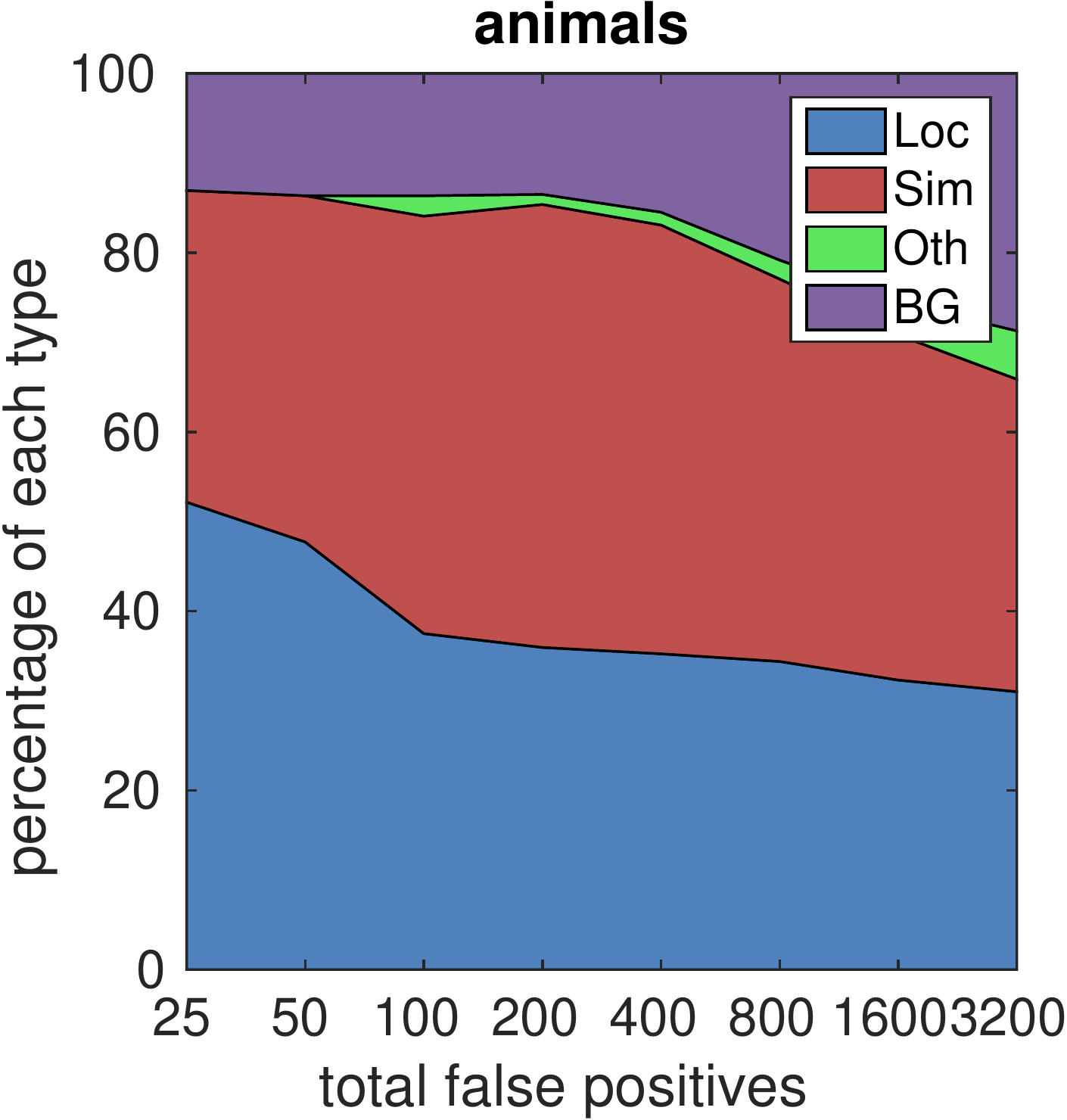}
    \hspace{0.4cm}
    \includegraphics[width=0.285\linewidth]{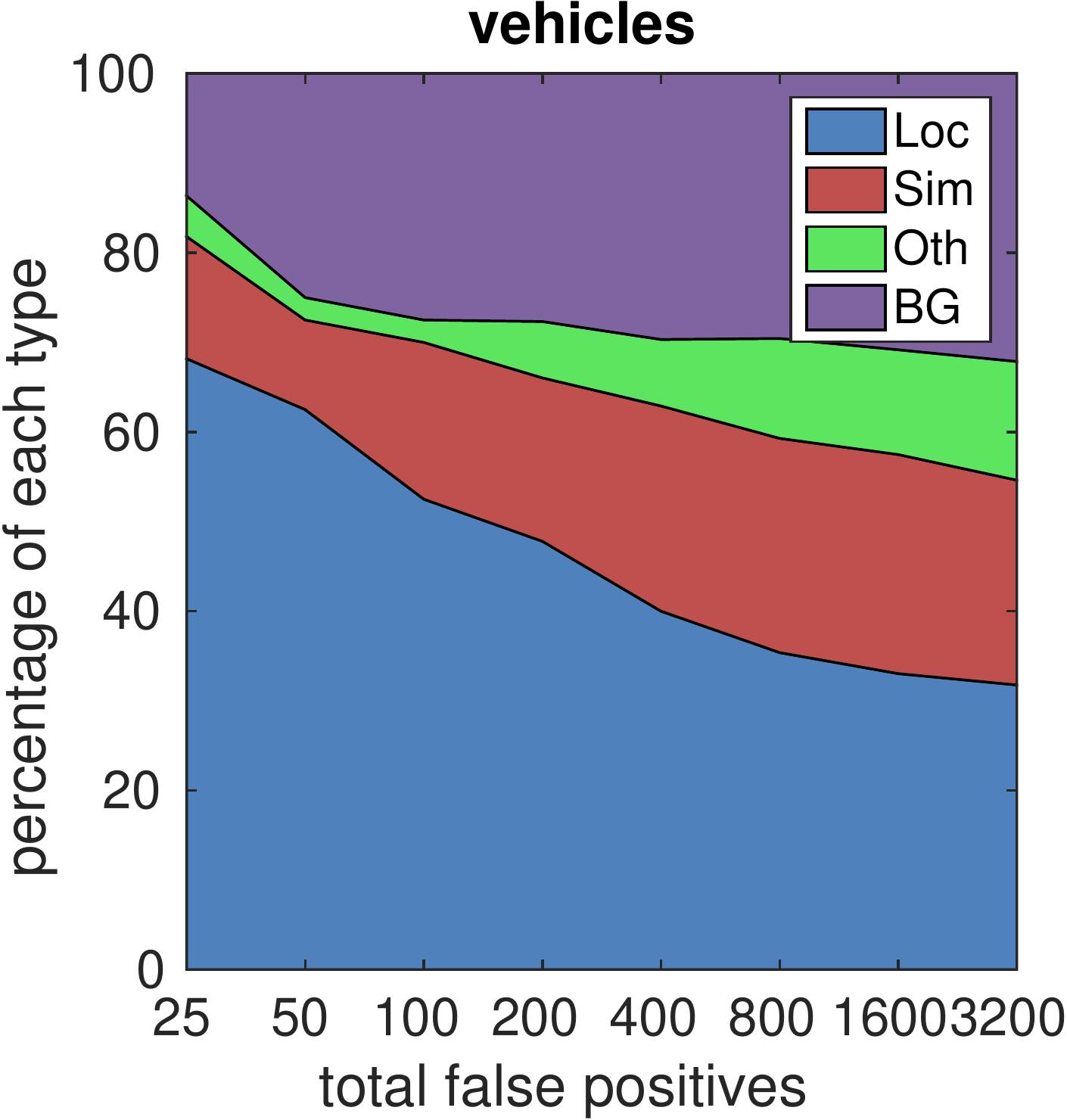}
    \hspace{0.4cm}
    \includegraphics[width=0.285\linewidth]{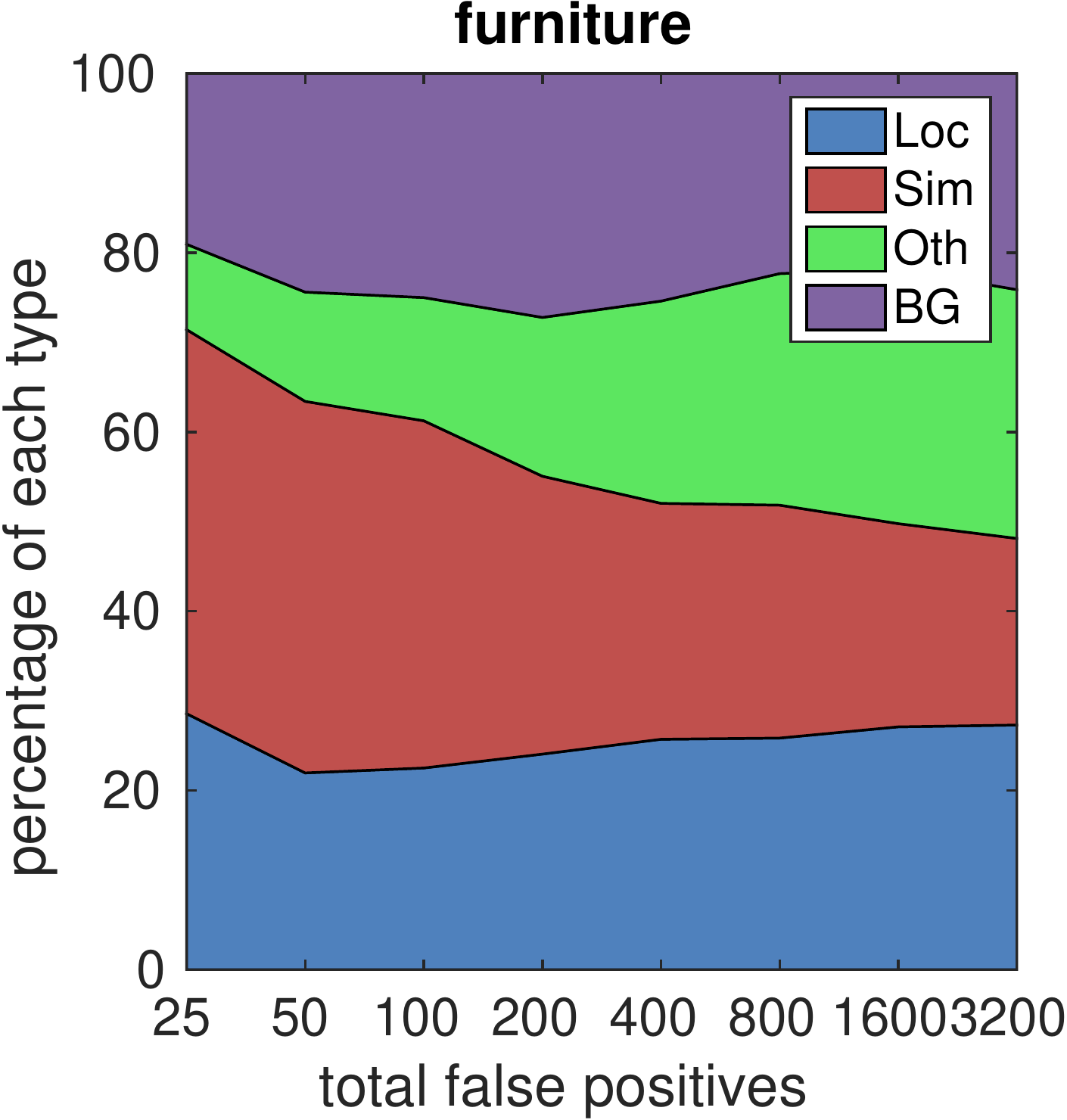}\\
    \caption{\textbf{Visualization of performance for SSD512 on animals, vehicles, and furniture from VOC2007 \texttt{test}.} The top row shows the cumulative fraction of detections that are correct (Cor) or false positive due to poor localization (Loc), confusion with similar categories (Sim), with others (Oth), or with background (BG). The solid red line reflects the change of recall with âstrongâ criteria (0.5 jaccard overlap) as the num- ber of detections increases. The dashed red line is using the âweakâ criteria (0.1 jaccard overlap). The bottom row shows the distribution of top-ranked false positive types.}
    \label{fig:fpanalysis}
\end{figure}

\begin{figure*}[htbp]
	\centering
    \includegraphics[width=0.495\linewidth]{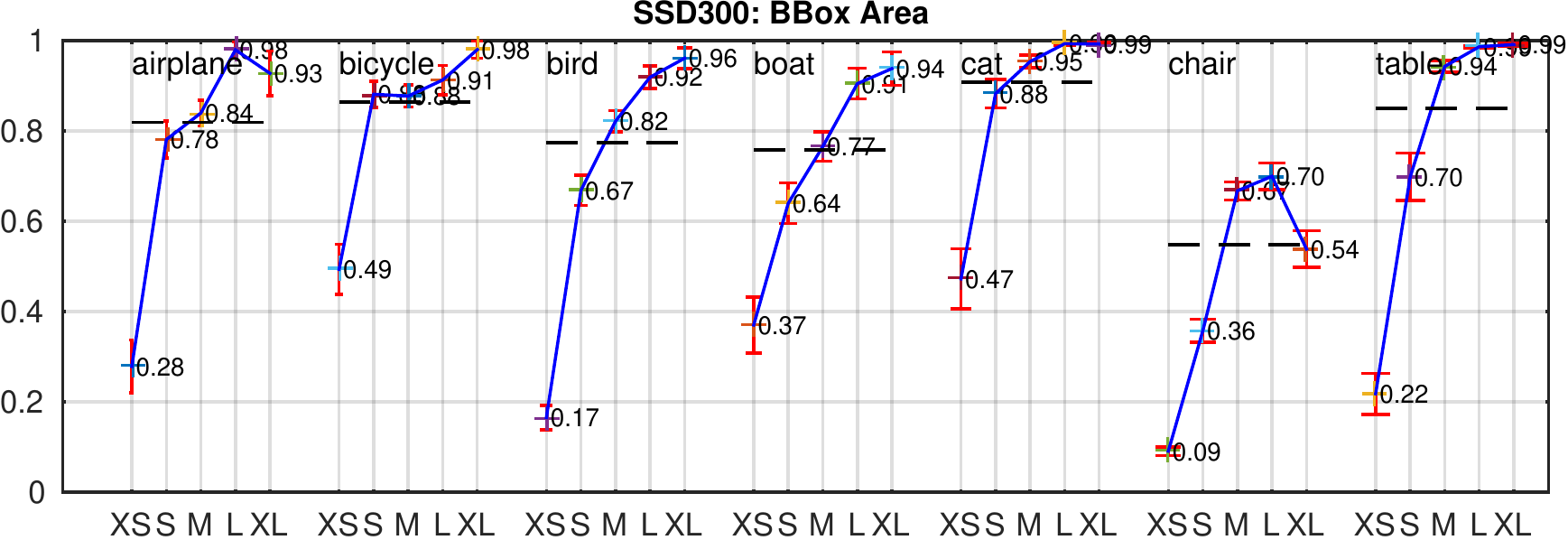}
    \includegraphics[width=0.495\linewidth]{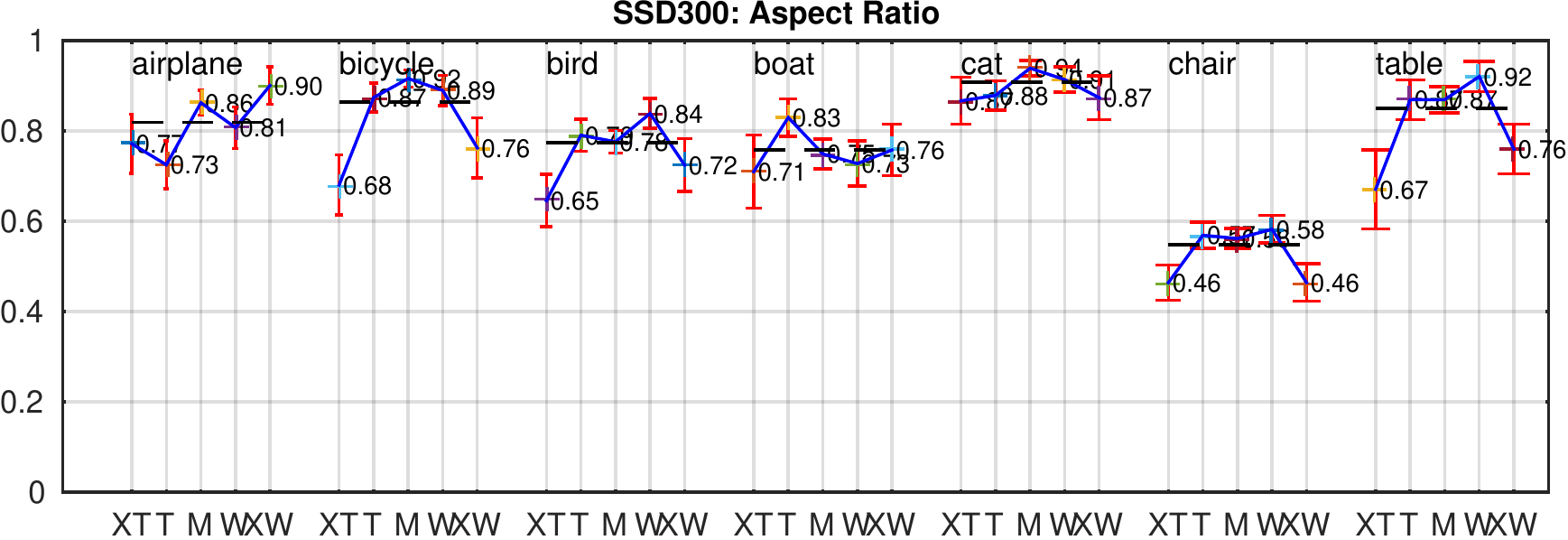}\\
    \includegraphics[width=0.495\linewidth]{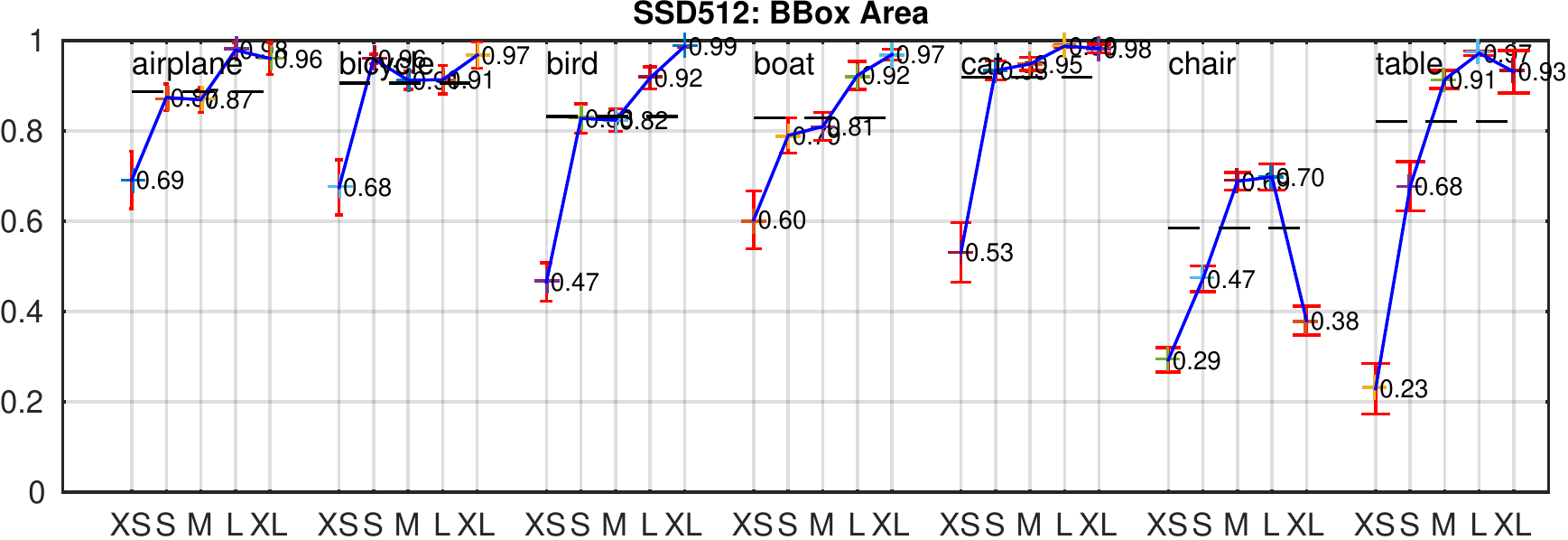}
    \includegraphics[width=0.495\linewidth]{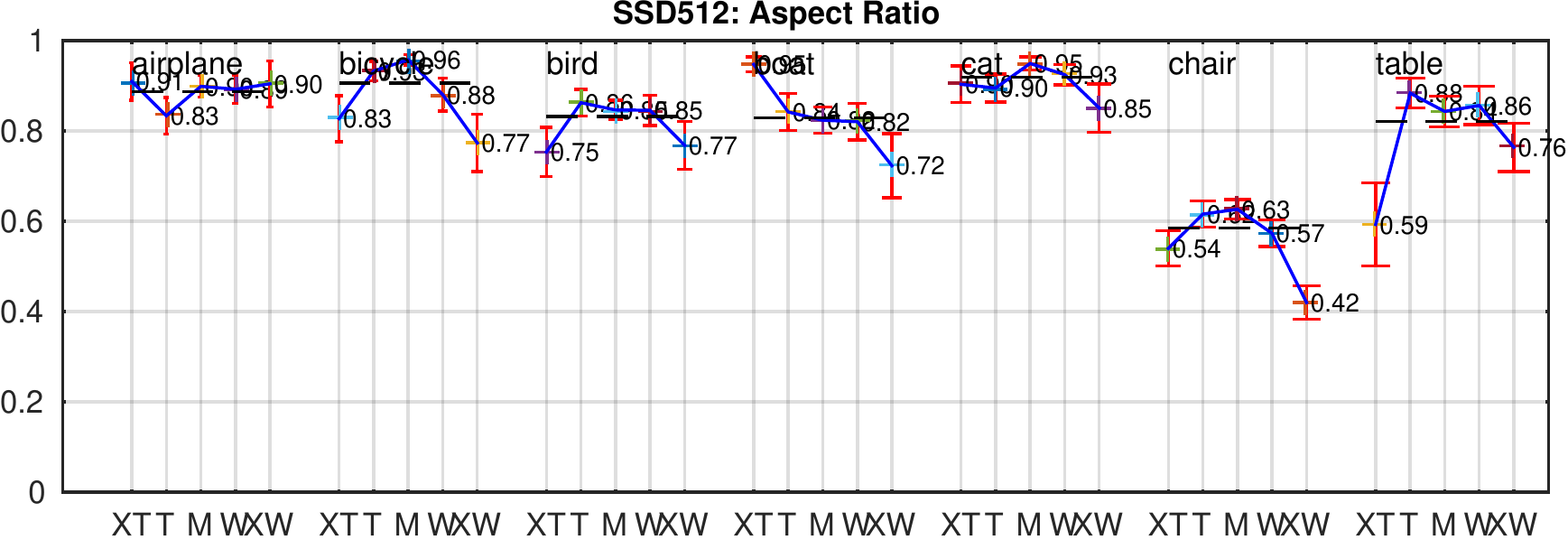}
    \caption{\textbf{Sensitivity and impact of different object characteristics on VOC2007 \texttt{test} set using~\cite{hoiem2012diagnosing}.} The plot on the left shows the effects of BBox Area per category, and the right plot shows the effect of Aspect Ratio. Key: BBox Area: XS=extra-small; S=small; M=medium; L=large; XL =extra-large. Aspect Ratio: XT=extra-tall/narrow; T=tall; M=medium; W=wide; XW =extra-wide.}
    \label{fig:sensitivityanalysis}
\end{figure*}

To understand the performance of our two SSD models in more details, we used the detection analysis tool from~\cite{hoiem2012diagnosing}. Figure~\ref{fig:fpanalysis} shows that SSD can detect various object categories with high quality (large white area). The majority of its confident detections are correct. The recall is around 85-90\%, and is much higher with ``weak" (0.1 jaccard overlap) criteria. Compared to R-CNN~\cite{girshick2014rich}, SSD has less localization error, indicating that SSD can localize objects better because it directly learns to regress the object shape and classify object categories instead of using two decoupled steps. However, SSD has more confusions with similar object categories (especially for animals), partly because we share locations for multiple categories. Figure~\ref{fig:sensitivityanalysis} shows that SSD is very sensitive to the bounding box size. In other words, it has much worse performance on smaller objects than bigger objects. This is not surprising because those small objects may not even have any information at the very top layers. Increasing the input size (e.g. from $300\times 300$ to $512\times 512$) can help improve detecting small objects, but there is still a lot of room to improve. On the positive side, we can clearly see that SSD performs really well on large objects. And it is very robust to different object aspect ratios because we use default boxes of various aspect ratios per feature map location.

\subsection{Model analysis}
To understand SSD better, we carried out controlled experiments to examine how each component affects performance. For all the experiments, we use the same settings and input size ($300\times 300$), except for specified changes to the settings or component(s).
\begin{table}
	\centering
	\setlength{\tabcolsep}{4pt}
	\begin{tabular}{r|cccccc}
    	& \multicolumn{5}{c}{SSD300} \\
    	\hline
        more data augmentation? & & \ding{52} & \ding{52} & \ding{52} & \ding{52}\\
        include \{$\frac{1}{2},2$\} box? & \ding{52} & & \ding{52} & \ding{52} & \ding{52}\\
        include \{$\frac{1}{3},3$\} box? & \ding{52} & & & \ding{52} & \ding{52}\\
        use atrous? & \ding{52} & \ding{52} & \ding{52} & & \ding{52}\\
        \hline
        VOC2007 \texttt{test} mAP & 65.5 & 71.6 & 73.7 & 74.2 & \textbf{74.3}\\
	\end{tabular}
    \caption{\textbf{Effects of various design choices and components on SSD performance.}}
    \label{tab:voc07analysis}
\end{table}

\smallskip
%\subsubsection{Data augmentation is crucial}
\noindent{\bf Data augmentation is crucial.} Fast and Faster R-CNN use the original image and the horizontal flip to train. We use a more extensive sampling strategy, similar to YOLO~\cite{redmon2015you}. Table~\ref{tab:voc07analysis} shows that we can improve 8.8\% mAP with this sampling strategy. We do not know how much our sampling strategy will benefit Fast and Faster R-CNN, but they are likely to benefit less because they use a feature pooling step during classification that is relatively robust to object translation by design.

%\subsubsection{More default box shapes is better}
\noindent{\bf More default box shapes is better.} As described in Sec.~\ref{sec:defaultboxes}, by default we use 6 default boxes per location. If we remove the boxes with $\frac{1}{3}$ and 3 aspect ratios, the performance drops by 0.6\%. By further removing the boxes with $\frac{1}{2}$ and 2 aspect ratios, the performance drops another 2.1\%. Using a variety of default box shapes seems to make the task of predicting boxes easier for the network.

%\subsubsection{Atrous is faster}
\noindent{\bf Atrous is faster.} As described in Sec.~\ref{sec:exp}, we used the atrous version of a subsampled VGG16, following DeepLab-LargeFOV~\cite{chen2014semantic}. If we use the full VGG16, keeping pool5 with $2\times 2-s2$ and not subsampling parameters from fc6 and fc7, and add conv5\_3 for prediction, the result is about the same while the speed is about 20\% slower.
\begin{table}\footnotesize
	\centering
    \setlength{\tabcolsep}{2pt}
    \begin{tabular}{cccccc|>{\hspace{1.5pc}}cc|c}
    	\multicolumn{6}{c|}{Prediction source layers from:} & \multicolumn{2}{C{2.8cm}|}{mAP\newline use boundary boxes?} & \multirow{2}{*}{\# Boxes}\\
        conv4\_3 & conv7 & conv8\_2 & conv9\_2 & conv10\_2 & conv11\_2 & Yes & No & \\
        \hline
        \ding{52} & \ding{52} & \ding{52} & \ding{52} & \ding{52} & \ding{52} & 74.3 & 63.4 & 8732\\
        \ding{52} & \ding{52} & \ding{52} & \ding{52} & \ding{52} &  & \textbf{74.6} & 63.1  & 8764\\
        \ding{52} & \ding{52} & \ding{52} & \ding{52} &  &  & 73.8 & 68.4 & 8942\\
        \ding{52} & \ding{52} & \ding{52} &  &  &  & 70.7 & 69.2 & 9864\\
        \ding{52} & \ding{52} &  &  &  &  & 64.2 & 64.4 & 9025\\
        & \ding{52} &  &  &  &  & 62.4 & 64.0 & 8664\\
        \hline
	\end{tabular}
    \caption{\textbf{Effects of using multiple output layers.}}
    \label{tab:multilayers}
\end{table}

%\subsubsection{Different layers for different scales}
\noindent{\bf Multiple output layers at different resolutions is better.} A major contribution of SSD is using default boxes of different scales on different output layers. To measure the advantage gained, we progressively remove layers and compare results. For a fair comparison, every time we remove a layer, we adjust the default box tiling to keep the total number of boxes similar to the original (8732).  This is done by stacking more scales of boxes on remaining layers and adjusting scales of boxes if needed. We do not exhaustively optimize the tiling for each setting. Table~\ref{tab:multilayers} shows a decrease in accuracy with fewer layers, dropping monotonically from 74.3 to 62.4.
%We conjecture that it is because it imposes too much burden on a single layer if we put default boxes of multiple scales on it. But by spreading out default boxes of multiple scales to different layers can let each layer focus on detecting objects of certain scale, which make it easier to learn. 
When we stack boxes of multiple scales on a layer, many are on the image boundary and need to be handled carefully. We tried the strategy used in Faster R-CNN~\cite{ren2015faster}, ignoring boxes which are on the boundary. We observe some interesting trends. For example, it hurts the performance by a large margin if we use very coarse feature maps (e.g. conv11\_2 ($1\times 1$) or conv10\_2 ($3\times 3$)). The reason might be that we do not have enough large boxes to cover large objects after the pruning. When we use primarily finer resolution maps, the performance starts increasing again because even after pruning a sufficient number of large boxes remains. If we only use conv7 for prediction, the performance is the worst, reinforcing the message that it is critical to spread boxes of different scales over different layers. Besides, since our predictions do not rely on ROI pooling as in~\cite{girshick2015fast}, we do not have the \emph{collapsing bins} problem in low-resolution feature maps~\cite{zhang2016is}.  The SSD architecture combines predictions from feature maps of various resolutions to achieve comparable accuracy to Faster R-CNN, while using lower resolution input images.

\subsection{PASCAL VOC2012}
We use the same settings as those used for our basic VOC2007 experiments above, except that we use VOC2012 \texttt{trainval} and VOC2007 \texttt{trainval} and \texttt{test} (21503 images) for training, and test on VOC2012 \texttt{test} (10991 images). We train the models with $10^{-3}$ learning rate for 60k iterations, then $10^{-4}$ for 20k iterations. Table~\ref{tab:voc12} shows the results of our SSD300 and SSD512\footnote{\ssmall\url{http://host.robots.ox.ac.uk:8080/leaderboard/displaylb.php?cls=mean&challengeid=11&compid=4}} model. We see the same performance trend as we observed on VOC2007 test. Our SSD300 improves accuracy over Fast/Faster R-CNN. By increasing the training and testing image size to $512\times 512$, we are 4.5\% more accurate than Faster R-CNN. Compared to YOLO, SSD is significantly more accurate, likely due to  the use of convolutional default boxes from multiple feature maps and our matching strategy during training. When fine-tuned from models trained on COCO, our SSD512 achieves 80.0\% mAP, which is 4.1\% higher than Faster R-CNN.
\begin{table}[ht]\ssmall
	\centering
	\setlength{\tabcolsep}{1.45pt}
	\begin{tabular*}{\textwidth}{l|c|c|cccccccccccccccccccc}
		\tiny Method & data & \tiny mAP & \tiny aero & \tiny bike & \tiny bird & \tiny boat & \tiny bottle & \tiny bus & \tiny car & \tiny cat & \tiny chair & \tiny cow & \tiny table & \tiny dog & \tiny horse & \tiny mbike & \tiny person & \tiny plant & \tiny sheep & \tiny sofa & \tiny train & \tiny tv \\
        \hline
		\tiny Fast\cite{girshick2015fast} & 07++12 & 68.4 & 82.3 & 78.4 & 70.8 & 52.3 & 38.7 & 77.8 & 71.6 & 89.3 & 44.2 & 73.0 & 55.0 & 87.5 & 80.5 & 80.8 & 72.0 & 35.1 & 68.3 & 65.7 & 80.4 & 64.2\\
		\tiny Faster\cite{ren2015faster} & 07++12 & 70.4 & 84.9 & 79.8 & 74.3 & 53.9 & 49.8 & 77.5 & 75.9 & 88.5 & 45.6 & 77.1 & 55.3 & 86.9 & 81.7 & 80.9 & 79.6 & 40.1 & 72.6 & 60.9 & 81.2 & 61.5\\
        \tiny Faster\cite{ren2015faster} & \tiny 07++12+COCO & 75.9 & 87.4 & 83.6 & 76.8 & 62.9 & 59.6 & 81.9 & 82.0 & 91.3 & 54.9 & 82.6 & 59.0 & 89.0 & 85.5 & 84.7 & 84.1 & 52.2 & 78.9 & 65.5 & 85.4 & 70.2\\
		\tiny YOLO\cite{redmon2015you} & 07++12 & 57.9 & 77.0 & 67.2 & 57.7 & 38.3 & 22.7 & 68.3 & 55.9 & 81.4 & 36.2 & 60.8 & 48.5 & 77.2 & 72.3 & 71.3 & 63.5 & 28.9 & 52.2 & 54.8 & 73.9 & 50.8\\
		% \tiny Fast+YOLO~\cite{redmon2015you} & 70.7 & 83.4 & 78.5 & 73.5 & 55.8 & 43.4 & 79.1 & 73.1 & 89.4 & 49.4 & 75.5 & 57.0 & 87.5 & 80.9 & 81.0 & 74.7 & 41.8 & 71.5 & 68.5 & 82.1 & 67.2\\
		\hline
        \tiny SSD300 & 07++12 & 72.4 & 85.6 & 80.1 & 70.5 & 57.6 & 46.2 & 79.4 & 76.1 & 89.2 & 53.0 & 77.0 & 60.8 & 87.0 & 83.1 & 82.3 & 79.4 & 45.9 & 75.9 & 69.5 & 81.9 & 67.5\\
        \tiny SSD300 & \tiny 07++12+COCO & 77.5 & 90.2 & 83.3 & 76.3 & 63.0 & 53.6 & 83.8 & 82.8 & 92.0 & 59.7 & 82.7 & 63.5 & 89.3 & 87.6 & 85.9 & 84.3 & 52.6 & 82.5 & \textbf{74.1} & \textbf{88.4} & 74.2\\
        \tiny SSD512 & 07++12 & 74.9 & 87.4 & 82.3 & 75.8 & 59.0 & 52.6 & 81.7 & 81.5 & 90.0 & 55.4 & 79.0 & 59.8 & 88.4 & 84.3 & 84.7 & 83.3 & 50.2 & 78.0 & 66.3 & 86.3 & 72.0\\
        \tiny SSD512 & \tiny 07++12+COCO & \textbf{80.0} & \textbf{90.7} & \textbf{86.8} & \textbf{80.5} & \textbf{67.8} & \textbf{60.8} & \textbf{86.3} & \textbf{85.5} & \textbf{93.5} & \textbf{63.2} & \textbf{85.7} & \textbf{64.4} & \textbf{90.9} & \textbf{89.0} & \textbf{88.9} & \textbf{86.8} & \textbf{57.2} & \textbf{85.1} & 72.8 & \textbf{88.4} & \textbf{75.9}\\
        \noalign{\smallskip}
	\end{tabular*}
	\caption{\textbf{PASCAL VOC2012 \texttt{test} detection results.} Fast and Faster R-CNN use images with minimum dimension 600, while the image size for YOLO is $448\times 448$. data: "07++12": union of VOC2007 \texttt{trainval} and \texttt{test} and VOC2012 \texttt{trainval}. "07++12+COCO": first train on COCO \texttt{trainval35k} then fine-tune on 07++12.}
    \label{tab:voc12}
\end{table}
\vspace{-2em}
\subsection{COCO}
\label{sec:expcoco}
To further validate the SSD framework, we trained our SSD300 and SSD512 architectures on the COCO dataset. Since objects in COCO tend to be smaller than PASCAL VOC, we use smaller default boxes for all layers. We follow the strategy mentioned in Sec.~\ref{sec:defaultboxes}, but now our smallest default box has a scale of 0.15 instead of 0.2, and the scale of the default box on conv4\_3 is 0.07 (e.g. 21 pixels for a $300\times 300$ image)\footnote{For SSD512 model, we add extra conv12\_2 for prediction, set $s_\text{min}$ to 0.1, and 0.04 on conv4\_3.}.

\begin{table}
	\centering
	\setlength{\tabcolsep}{1pt}
	\begin{tabular*}{\textwidth}{l|c|C{3.4em}C{2.2em}C{2.2em}|C{2em}C{2em}C{2em}|C{2.2em}C{2.2em}C{2.2em}|C{2em}C{2em}C{2em}}
    	\multirow{2}{*}{Method} & \multirow{2}{*}{data} & \multicolumn{3}{c|}{\scriptsize{Avg. Precision, IoU:}} & \multicolumn{3}{c|}{\scriptsize{Avg. Precision, Area:}} & \multicolumn{3}{c|}{\scriptsize{Avg. Recall, \#Dets:}} & \multicolumn{3}{c}{\scriptsize{Avg. Recall, Area:}}\\
        & & 0.5:0.95 & 0.5 & 0.75 & S & M & L & 1 & 10 & 100 & S & M & L\\
        \hline
        Fast~\cite{girshick2015fast} & train & 19.7 & 35.9 & - & - & - & - & - & - & - & - & - & -\\
        Fast~\cite{bell2015inside} & train & 20.5 & 39.9 & 19.4 & 4.1 & 20.0 & 35.8 & 21.3 & 29.5 & 30.1 & 7.3 & 32.1 & 52.0\\
        %Faster~\cite{ren2015faster} & train & 21.5 & 42.1 & - & - & - & - & - & - & - & - & - & -\\
        Faster~\cite{ren2015faster} & trainval & 21.9 & 42.7 & - & - & - & - & - & - & - & - & - & -\\
        ION~\cite{bell2015inside} & train & 23.6 & 43.2 & 23.6 & 6.4 & 24.1 & 38.3 & 23.2 & 32.7 & 33.5 & 10.1 & 37.7 & 53.6\\
        Faster~\cite{cocoleaderboard} & trainval & 24.2 & 45.3 & 23.5 & 7.7 & 26.4 & 37.1 & 23.8 & 34.0 & 34.6 & 12.0 & 38.5 & 54.4\\
        \hline
        SSD300 & trainval35k & 23.2 & 41.2 & 23.4 & 5.3 & 23.2 & 39.6 & 22.5 & 33.2 & 35.3 & 9.6 & 37.6 & 56.5\\
        SSD512 & trainval35k & \textbf{26.8} & \textbf{46.5} & \textbf{27.8} & \textbf{9.0} & \textbf{28.9} & \textbf{41.9} & \textbf{24.8} & \textbf{37.5} & \textbf{39.8} & \textbf{14.0} & \textbf{43.5} & \textbf{59.0}\\
    \end{tabular*}
    \caption{\textbf{COCO \texttt{test-dev2015} detection results.}}
    \label{tab:coco}
\end{table}

\begin{figure}
	\centering
	\includegraphics[width=0.19\linewidth]{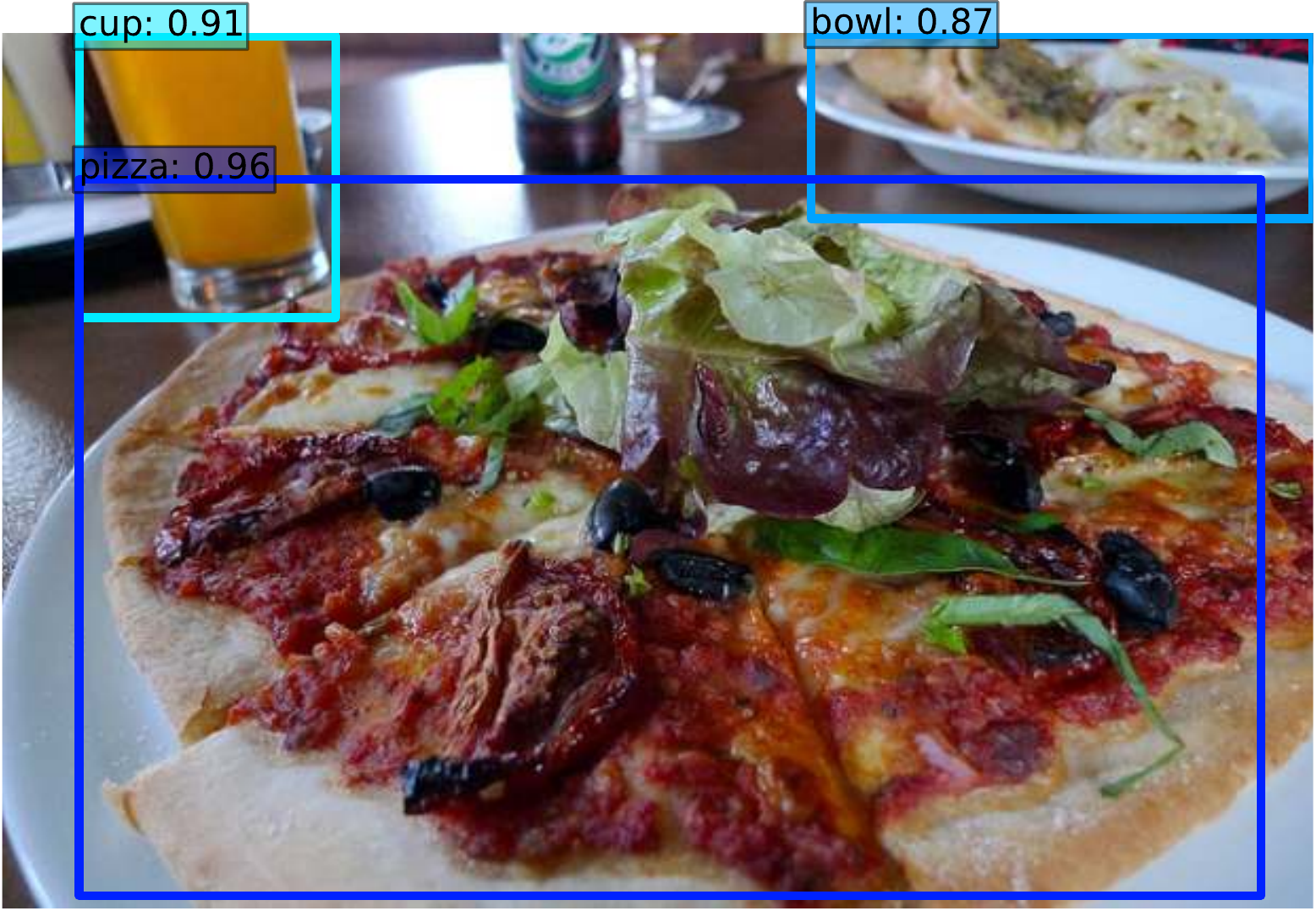}
	\includegraphics[width=0.19\linewidth]{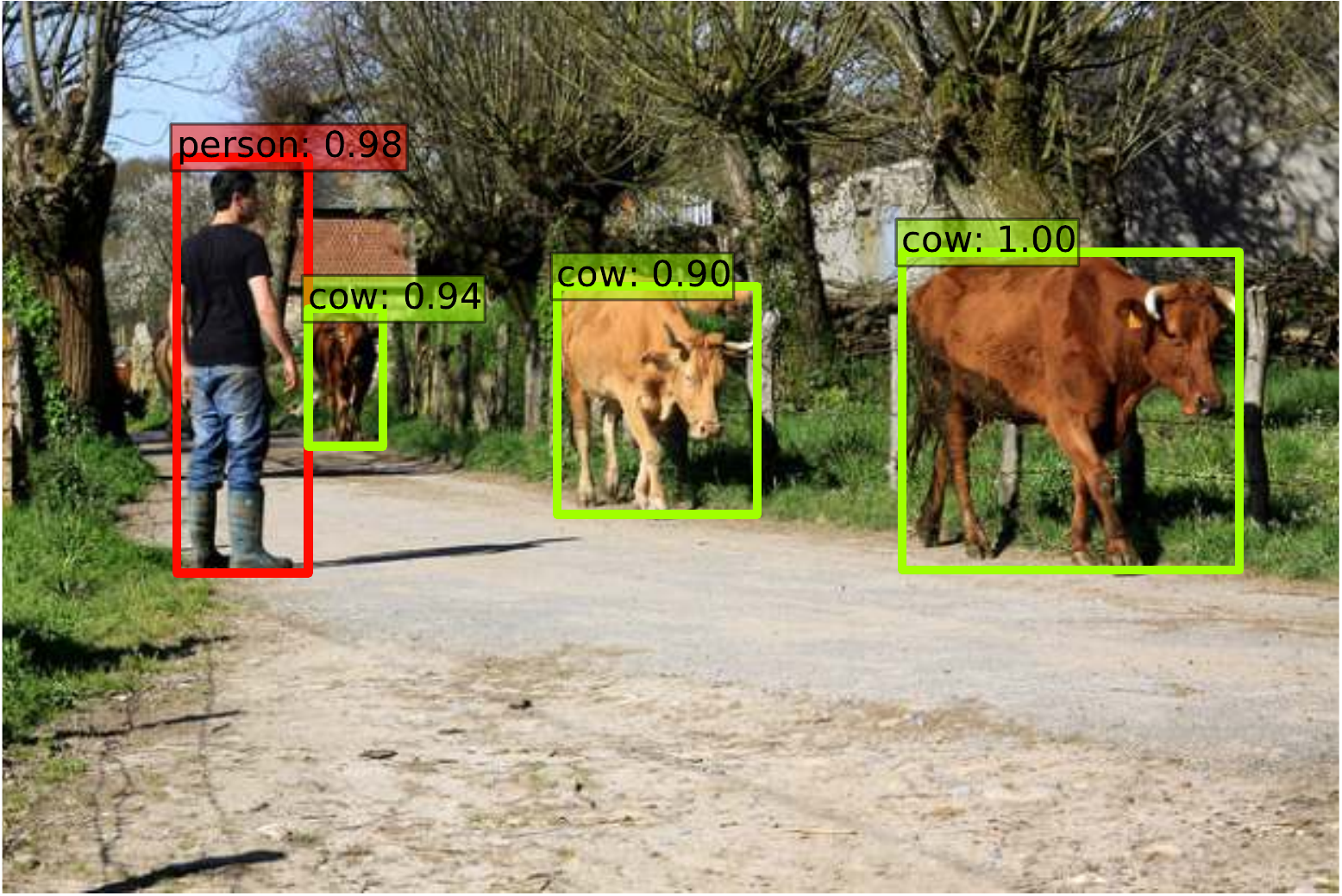}
	\includegraphics[width=0.19\linewidth]{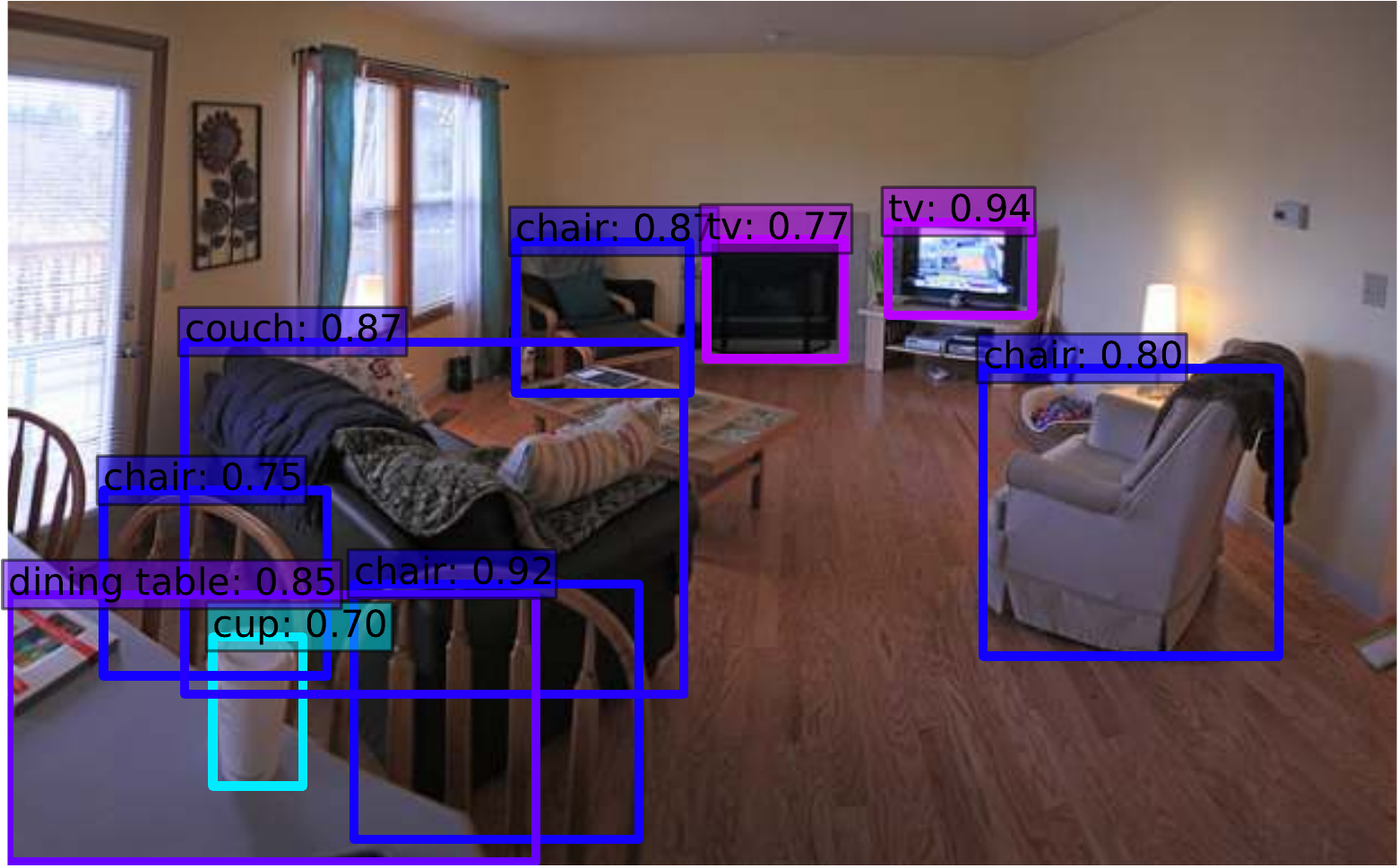}
	\includegraphics[width=0.19\linewidth]{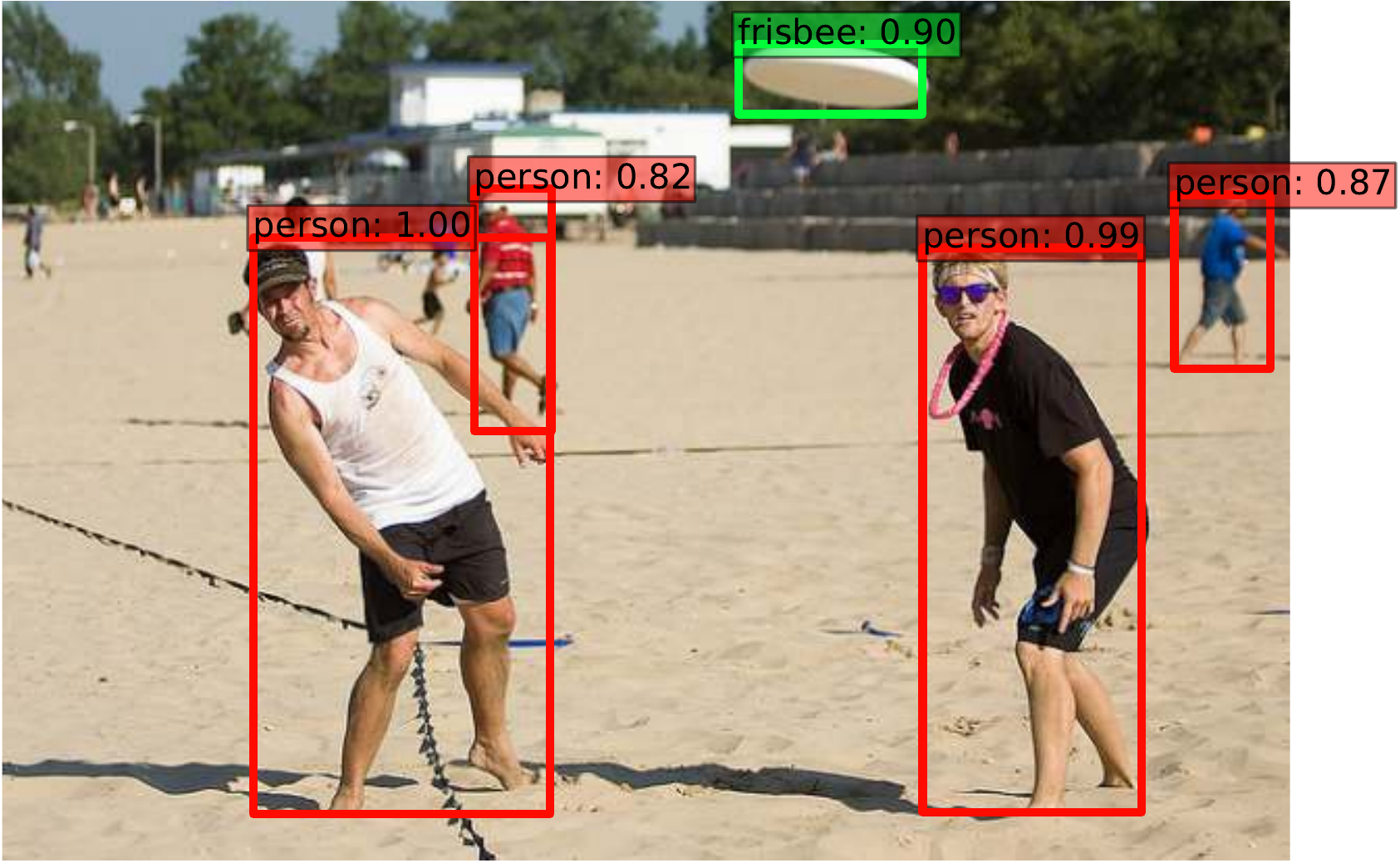}
	\includegraphics[width=0.19\linewidth]{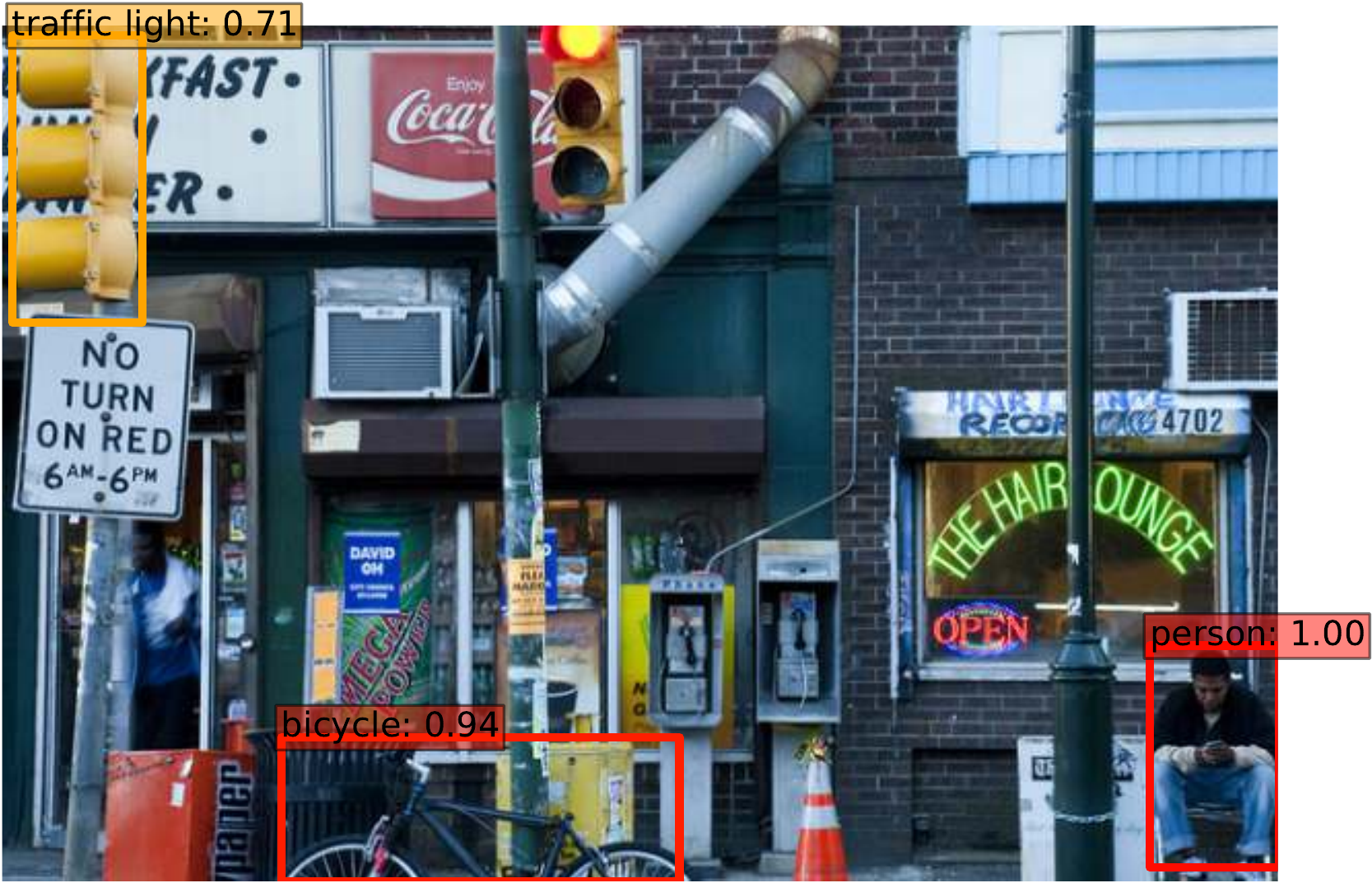}\\
	\includegraphics[trim={0 1.2cm 0 0},clip,width=0.19\linewidth]{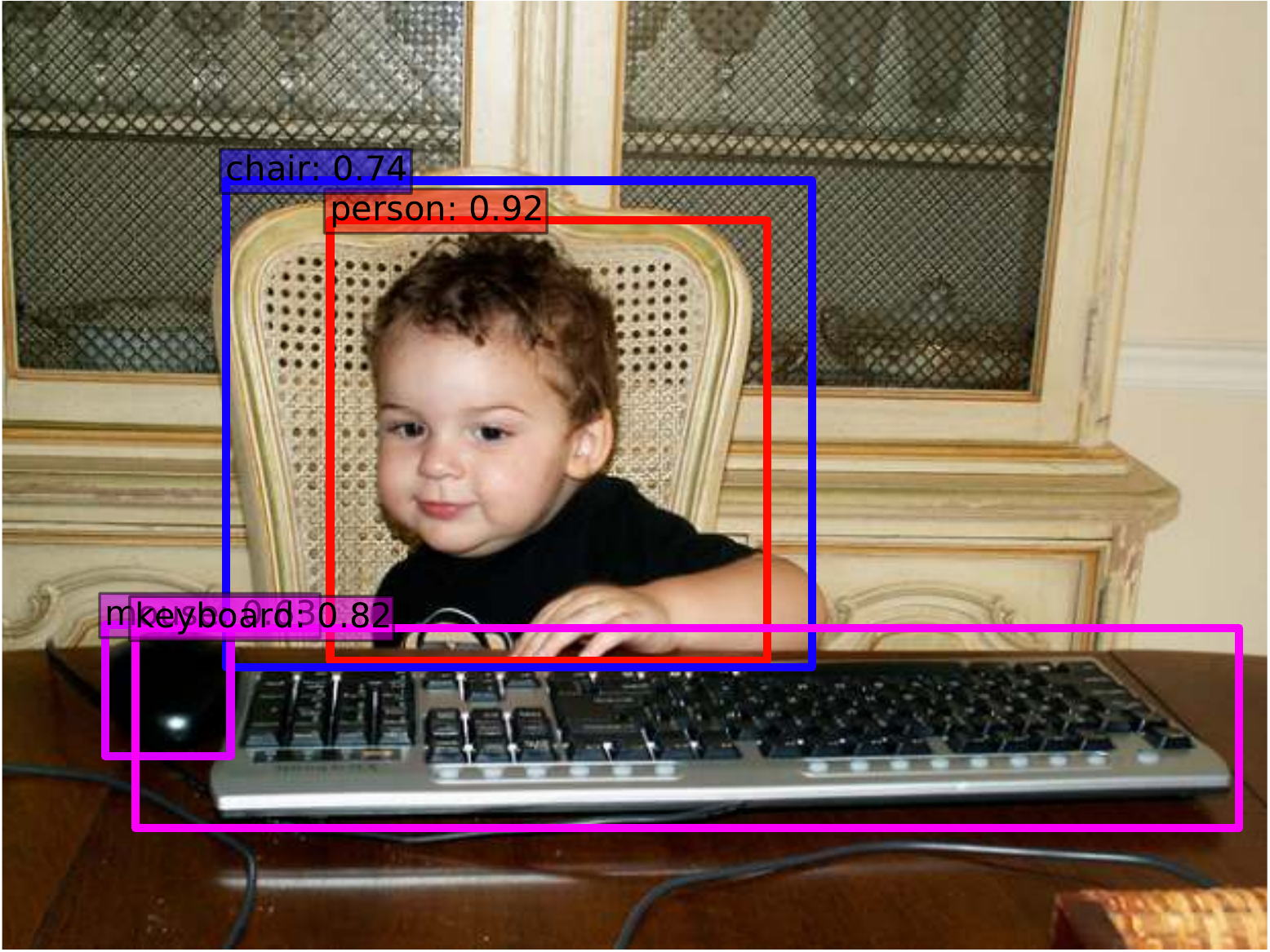}
	\includegraphics[width=0.19\linewidth]{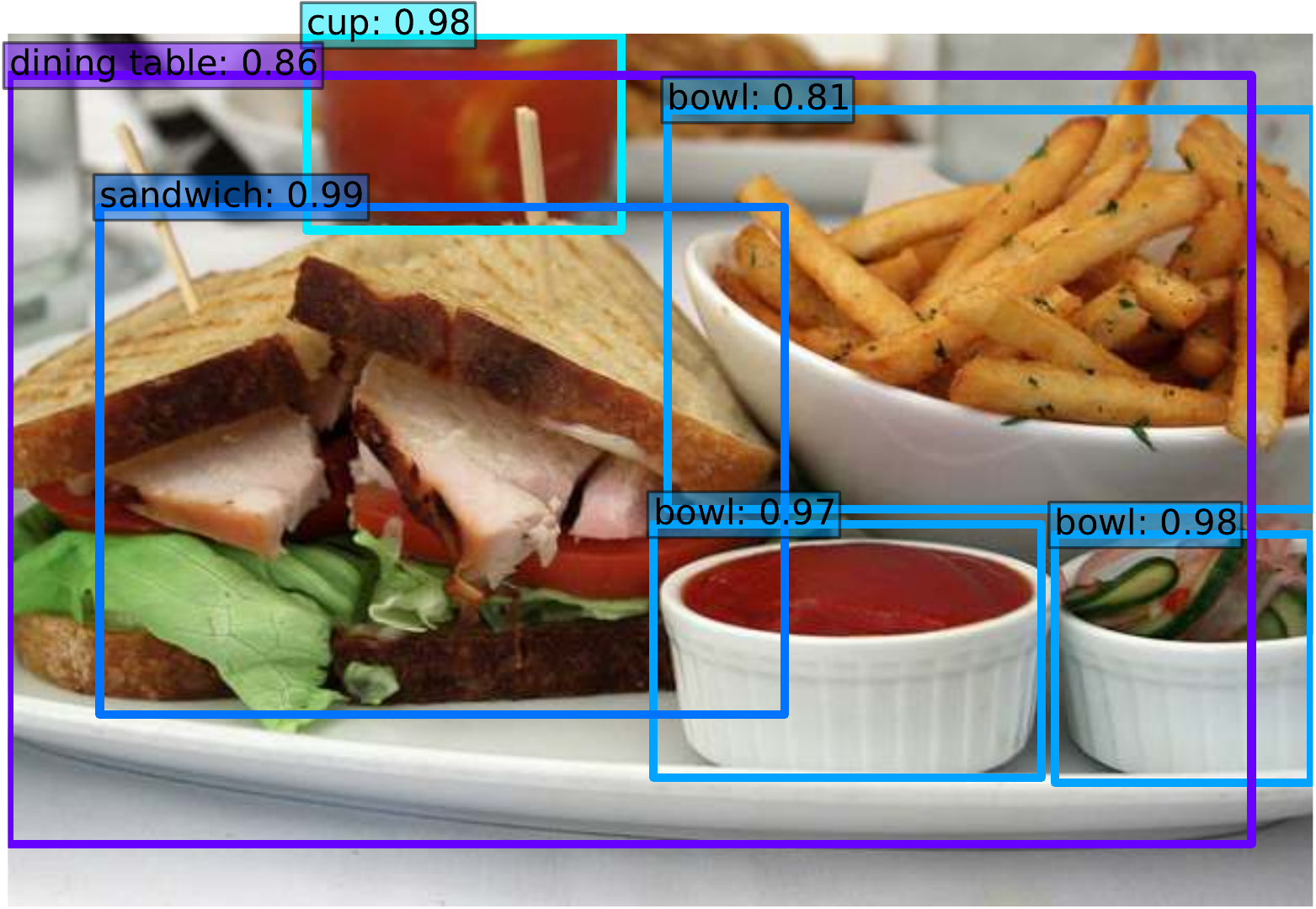}
	\includegraphics[width=0.19\linewidth]{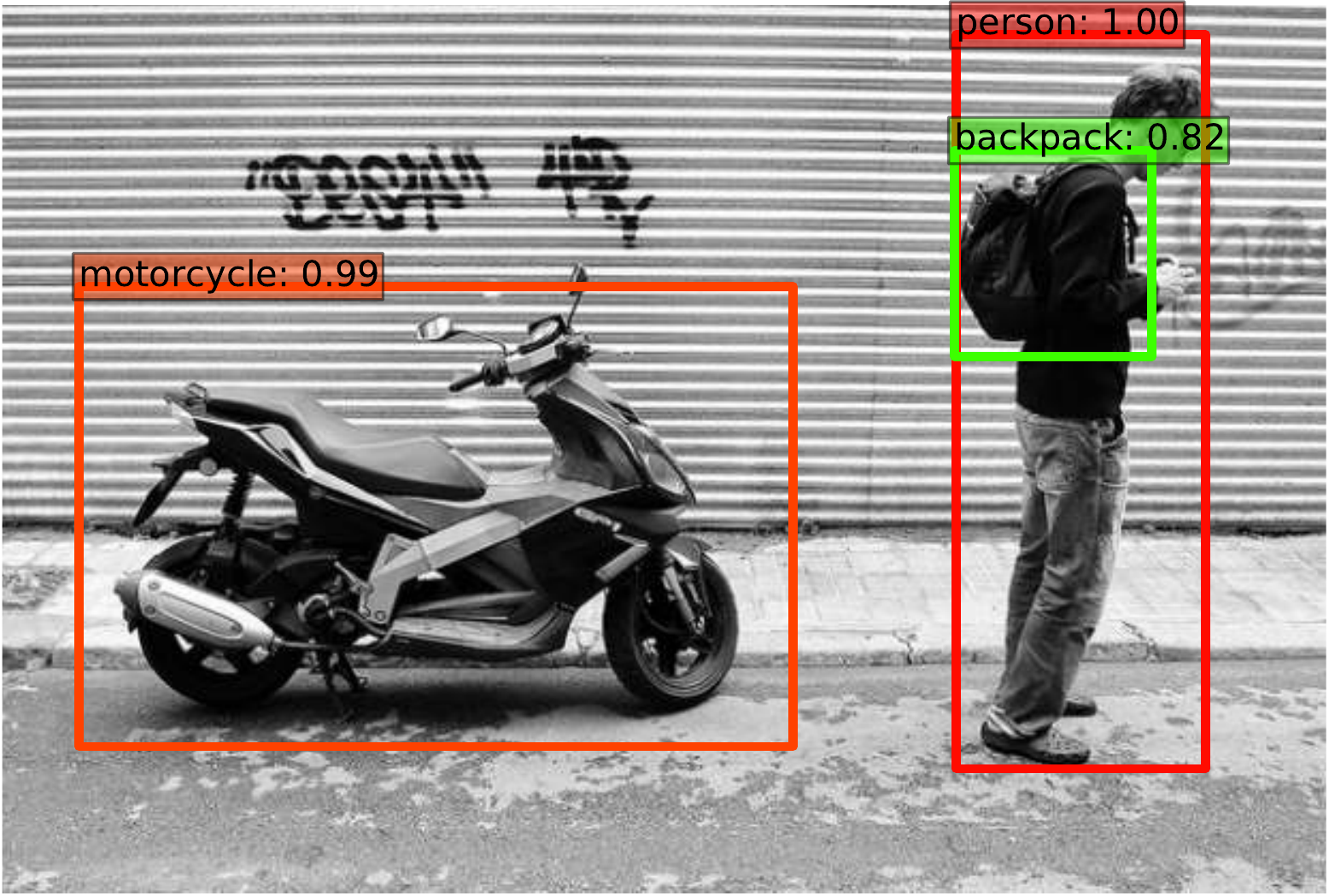}
	\includegraphics[trim={0 1.0cm 0 0},clip,width=0.19\linewidth]{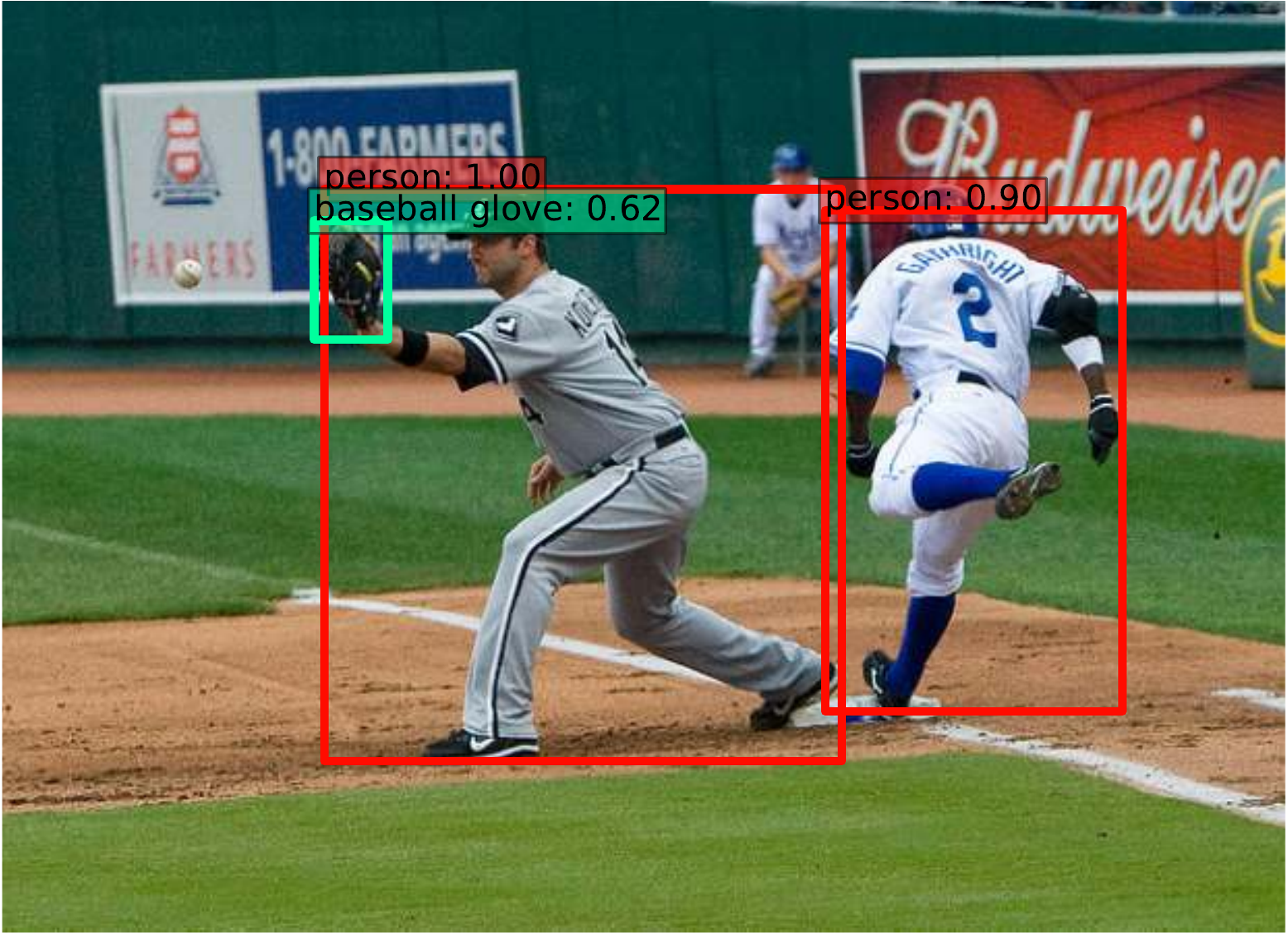}
	\includegraphics[width=0.19\linewidth]{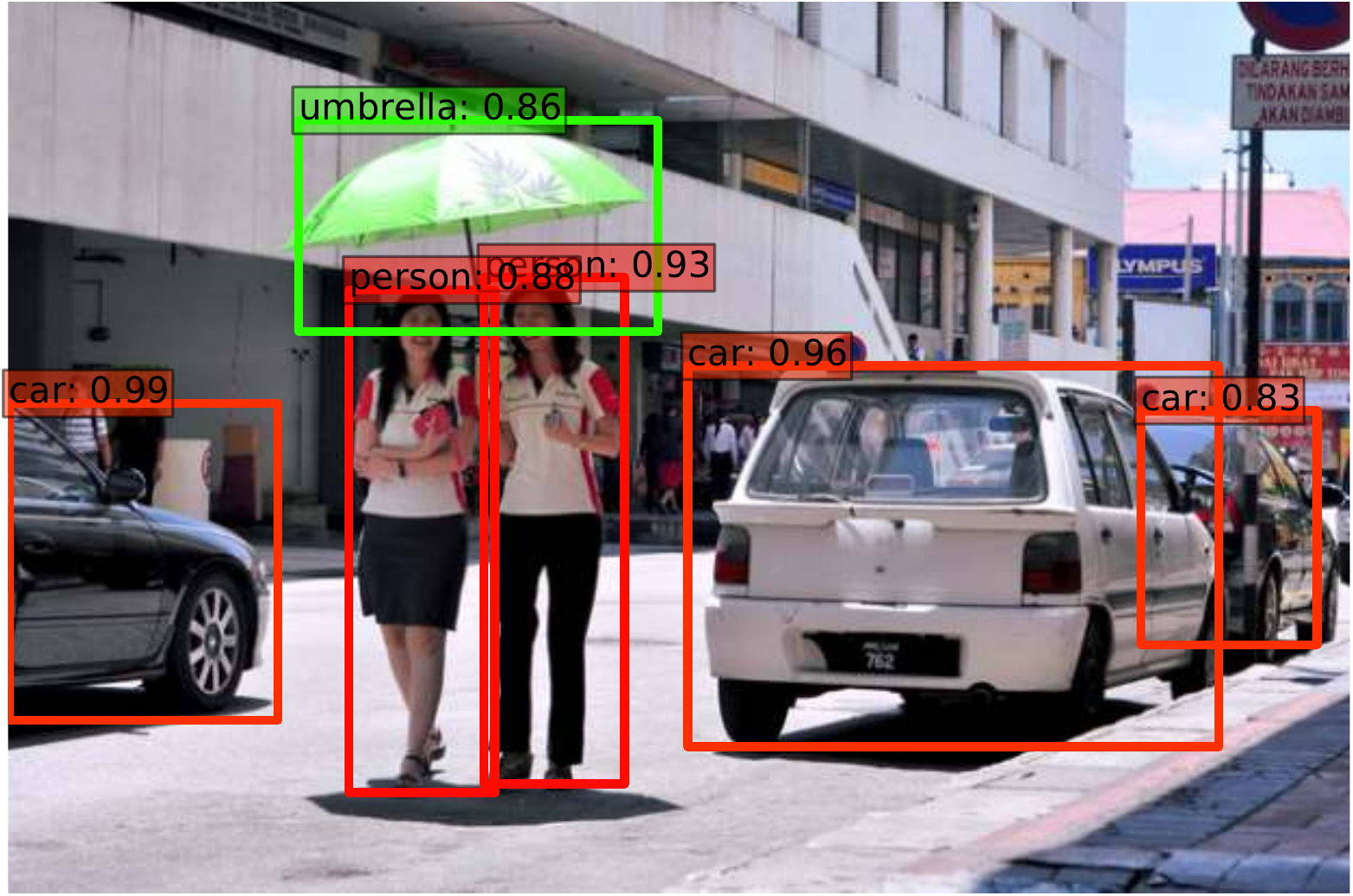}\\
	\includegraphics[trim={0 0.4cm 0 0},clip,width=0.19\linewidth]{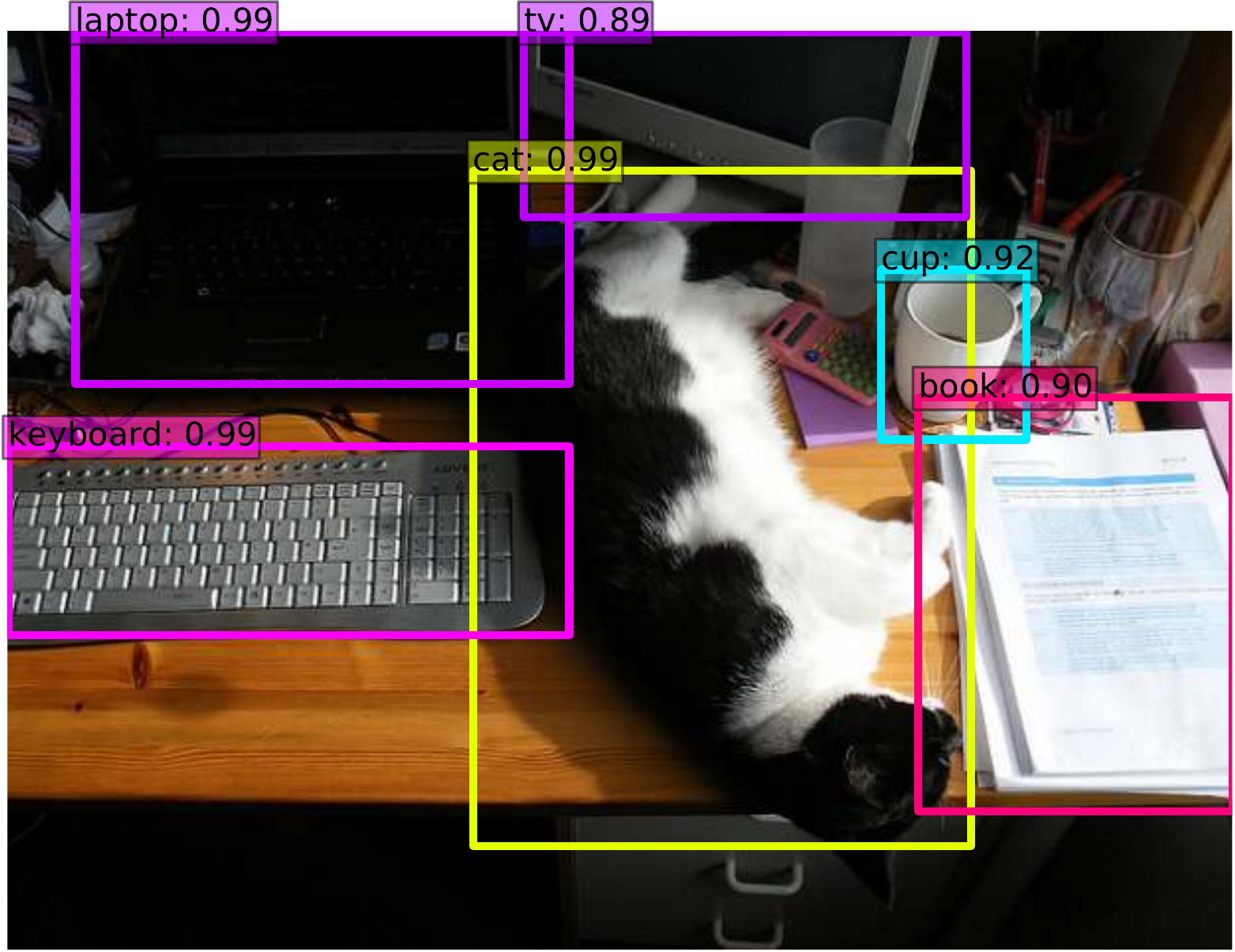}
	\includegraphics[width=0.19\linewidth]{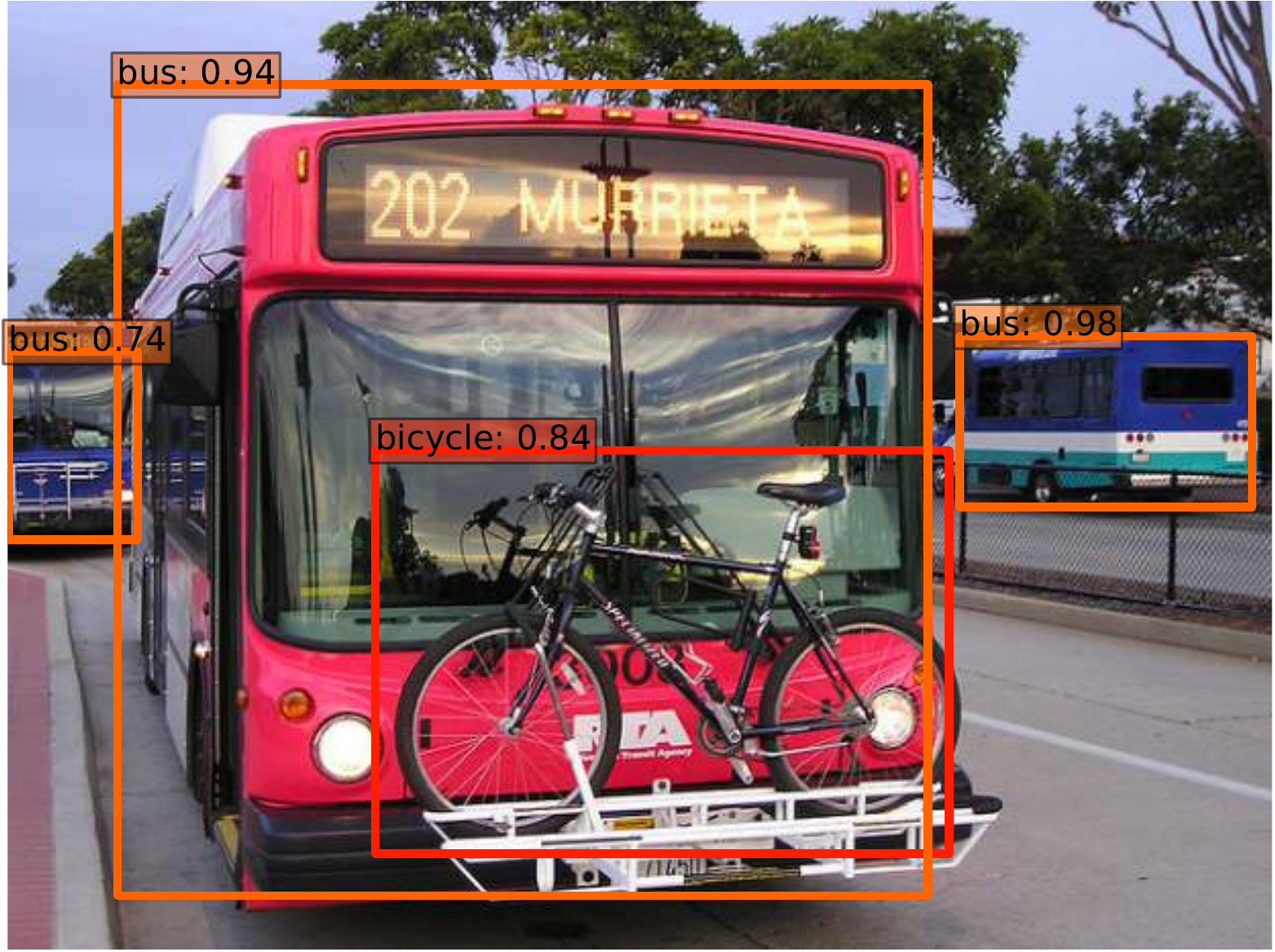}
	\includegraphics[width=0.19\linewidth]{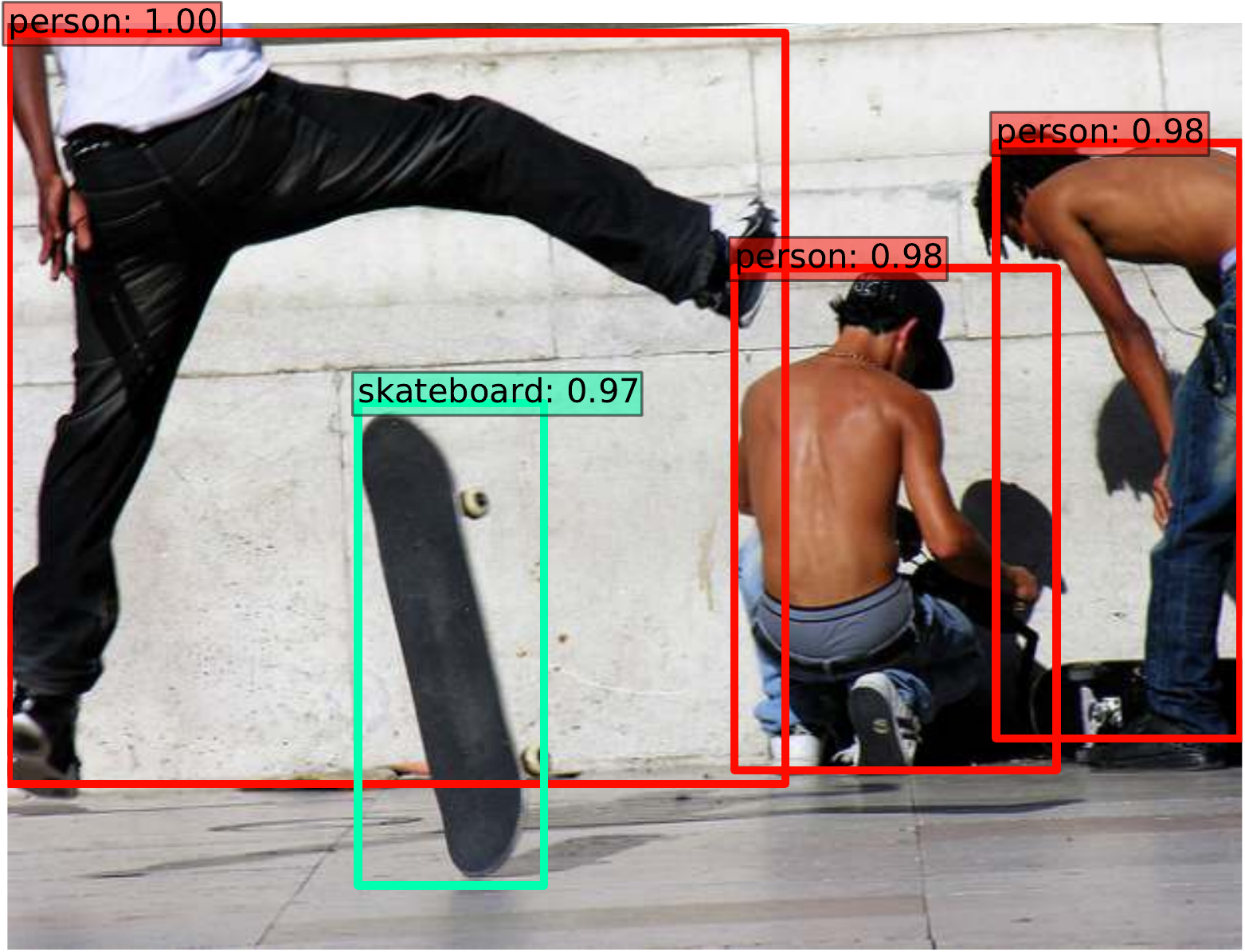}
	\includegraphics[width=0.19\linewidth]{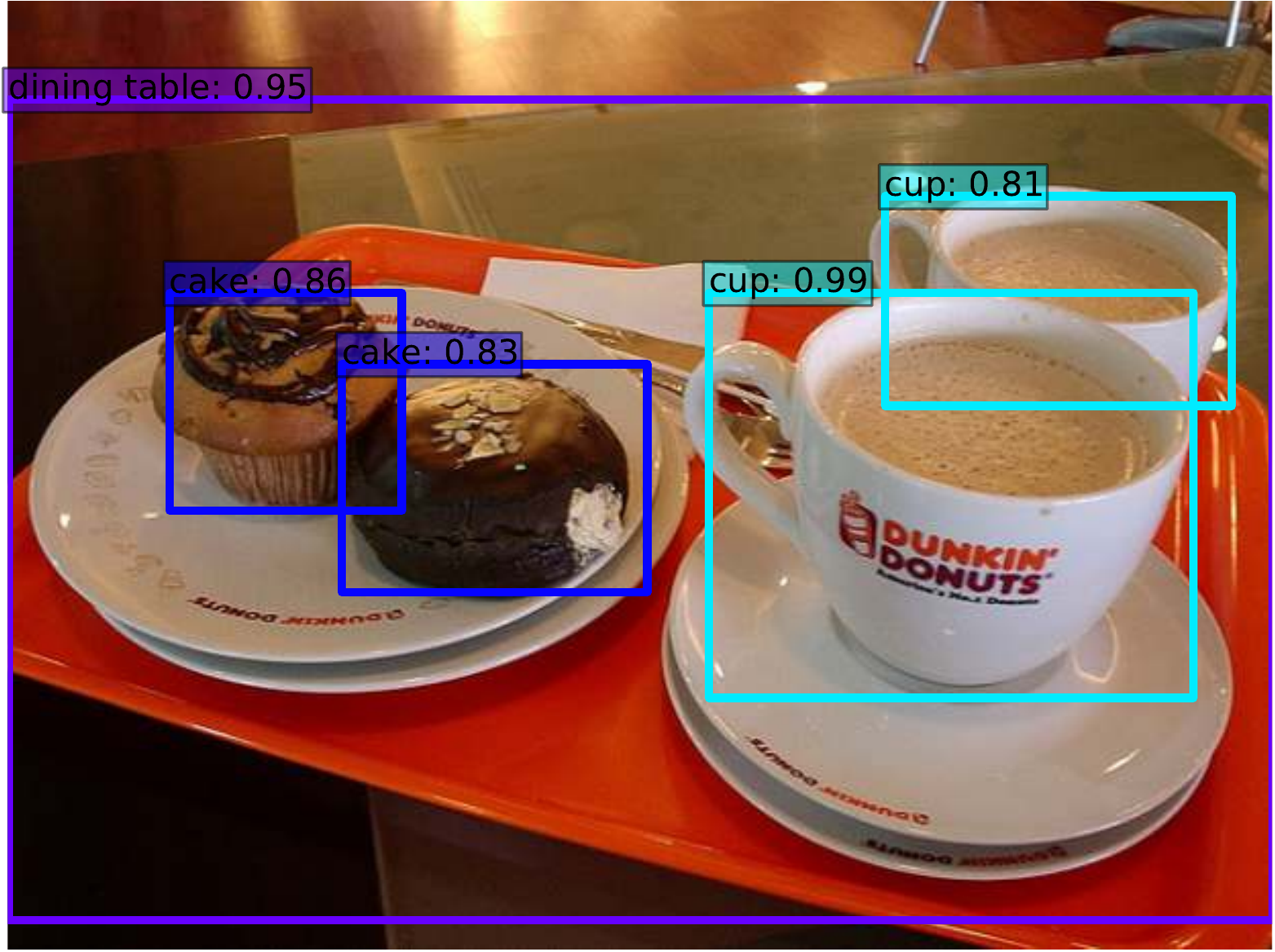}
	\includegraphics[width=0.19\linewidth]{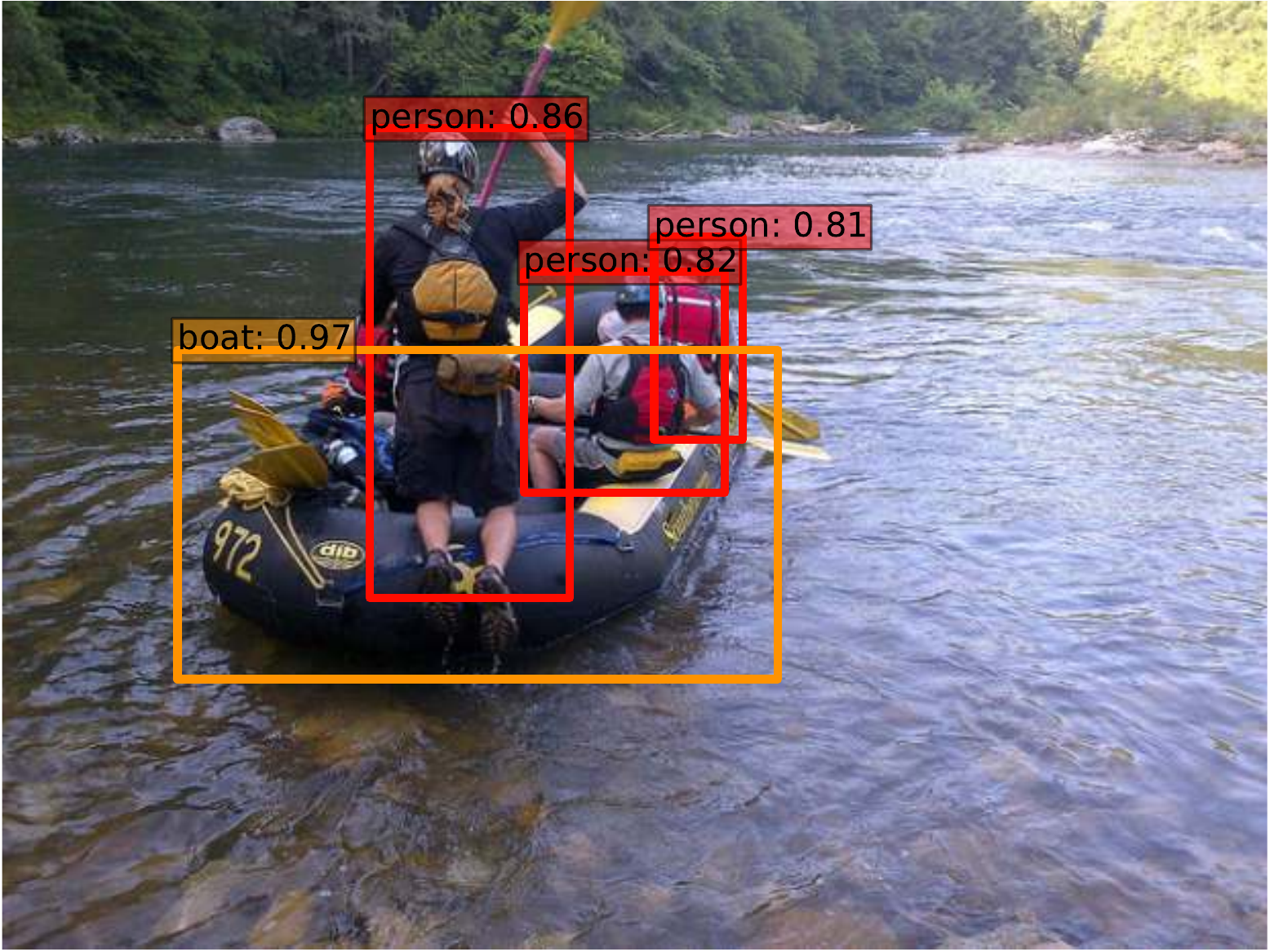}\\
	\includegraphics[trim={0 0 0 0.8cm},clip,width=0.19\linewidth]{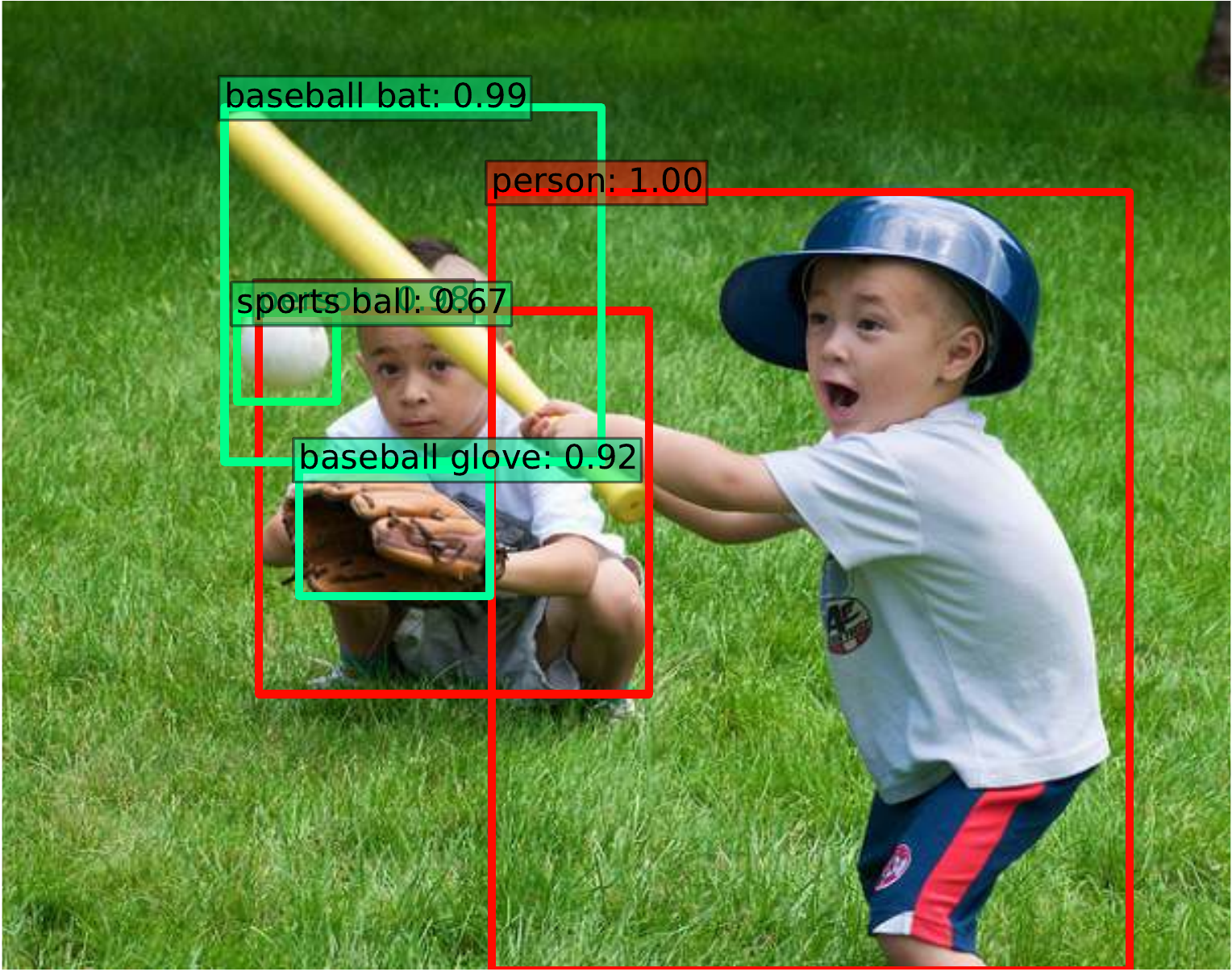}
	\includegraphics[width=0.19\linewidth]{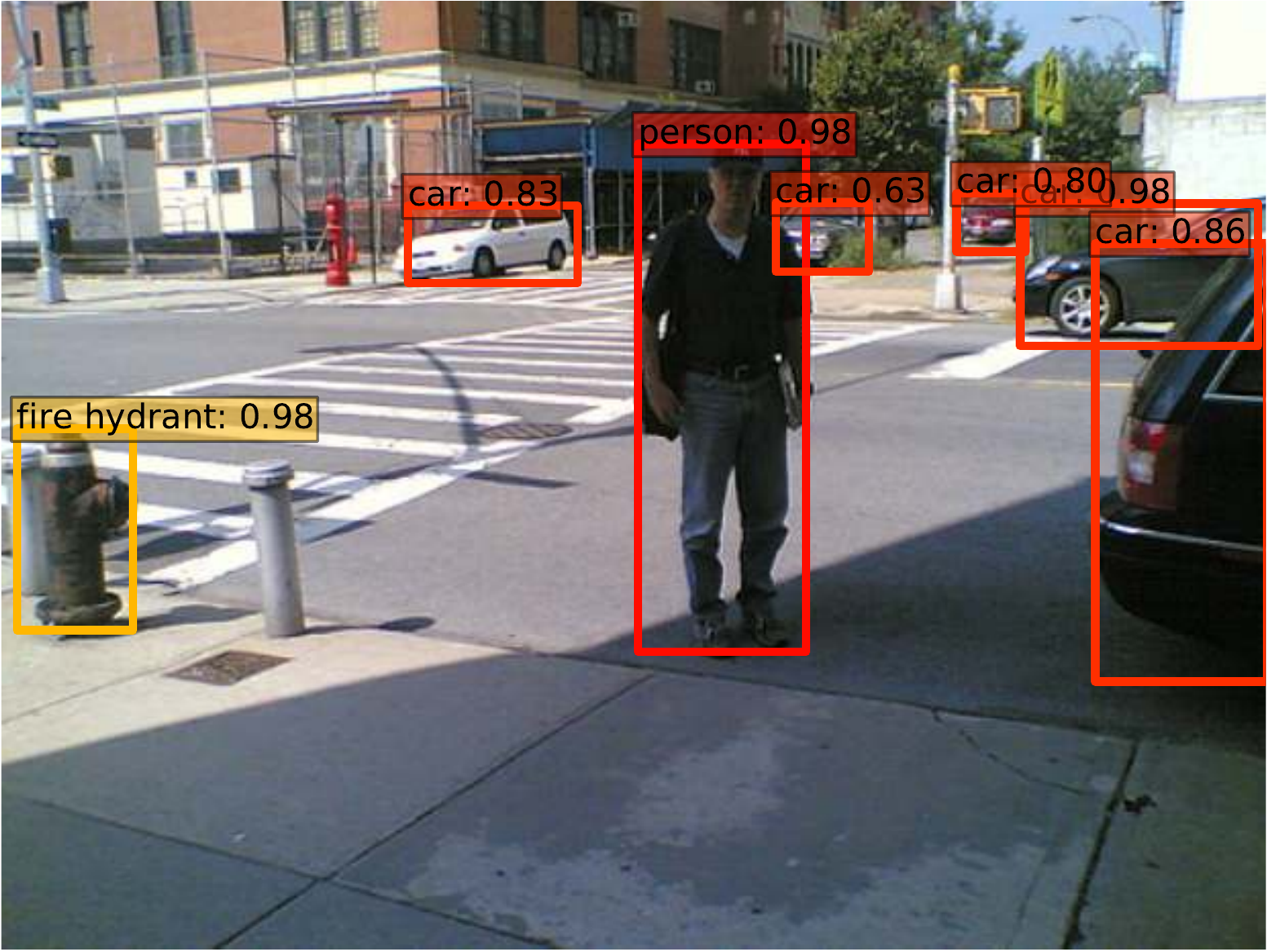}
	\includegraphics[width=0.19\linewidth]{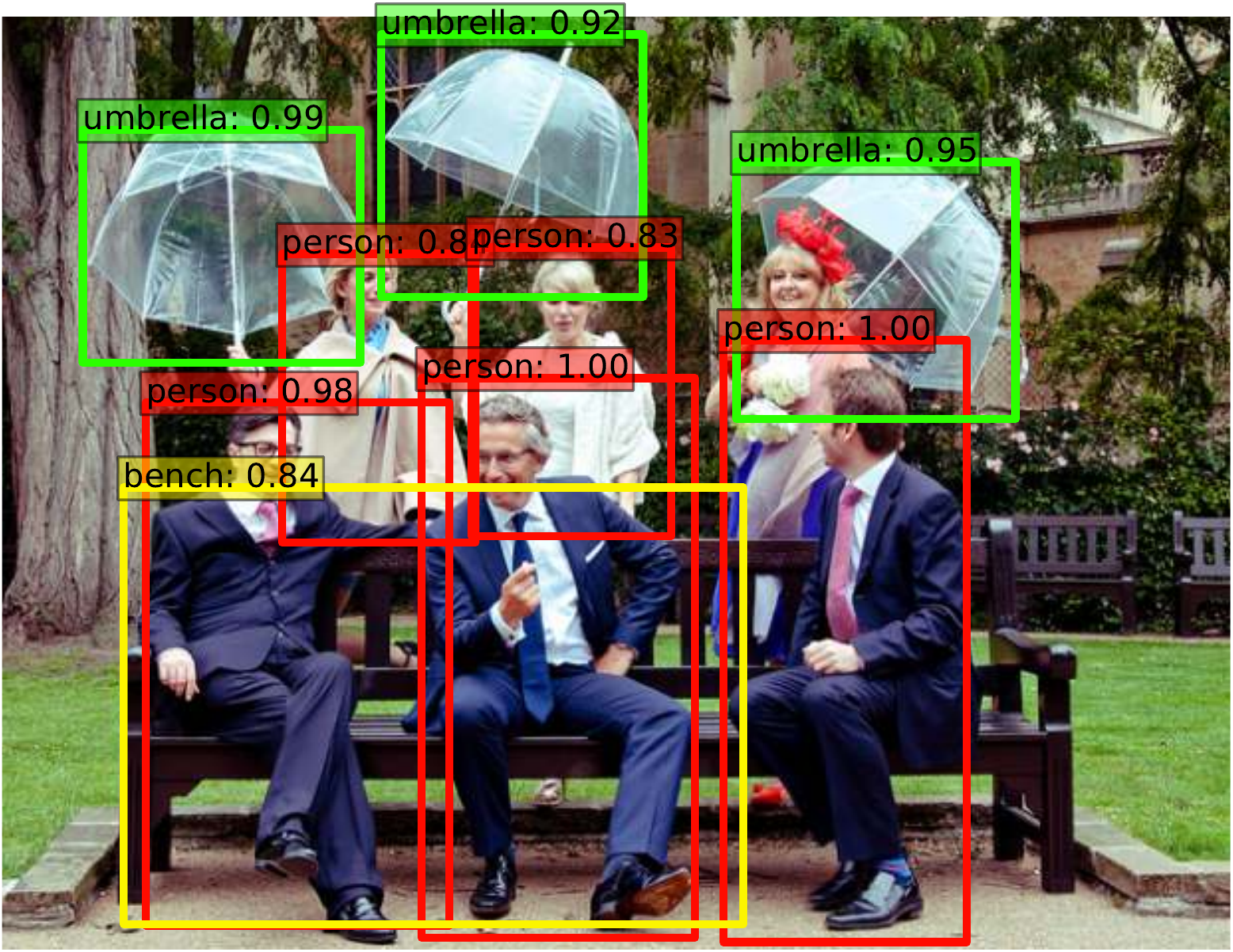}
	\includegraphics[width=0.19\linewidth]{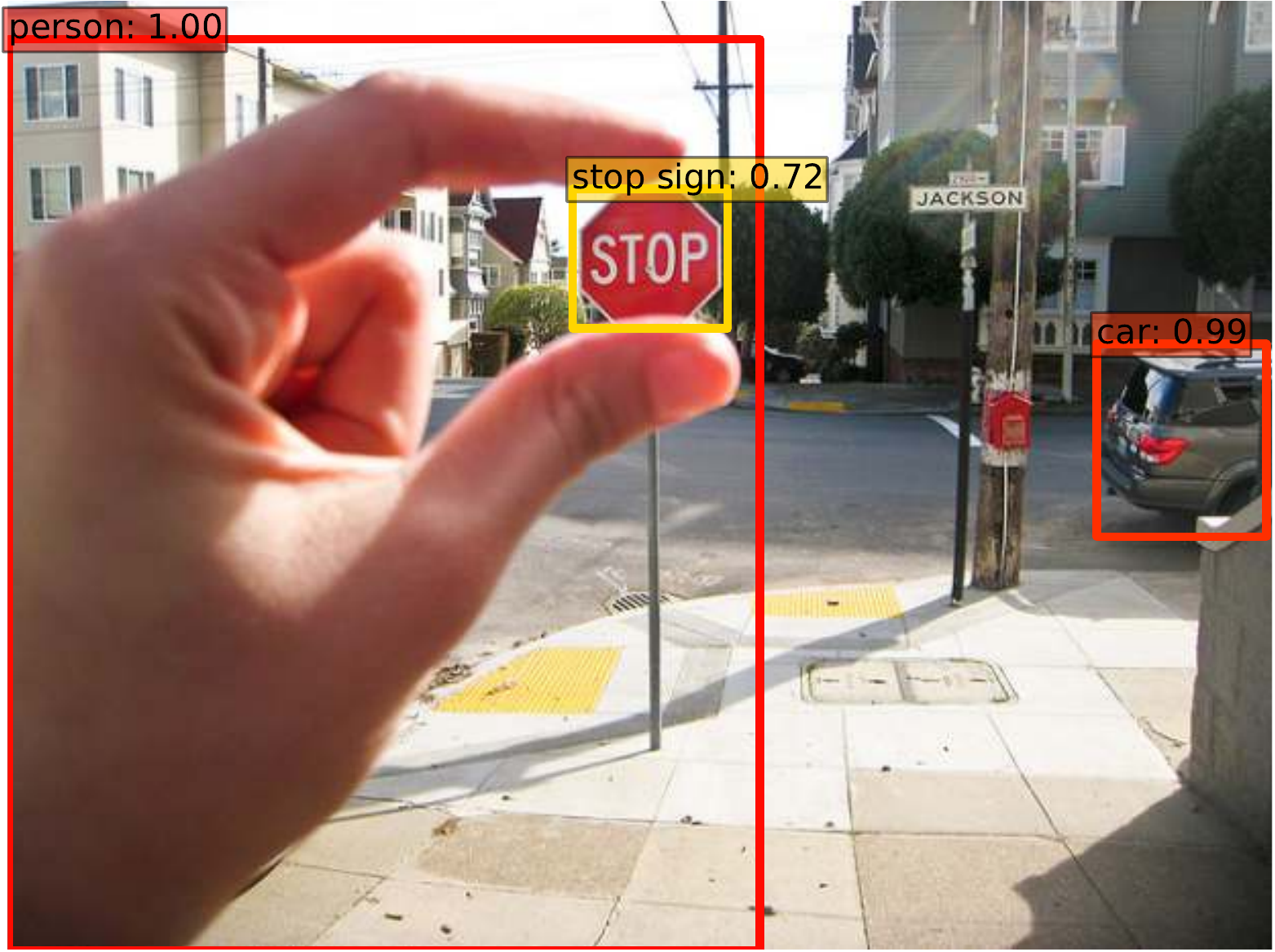}
	\includegraphics[width=0.19\linewidth]{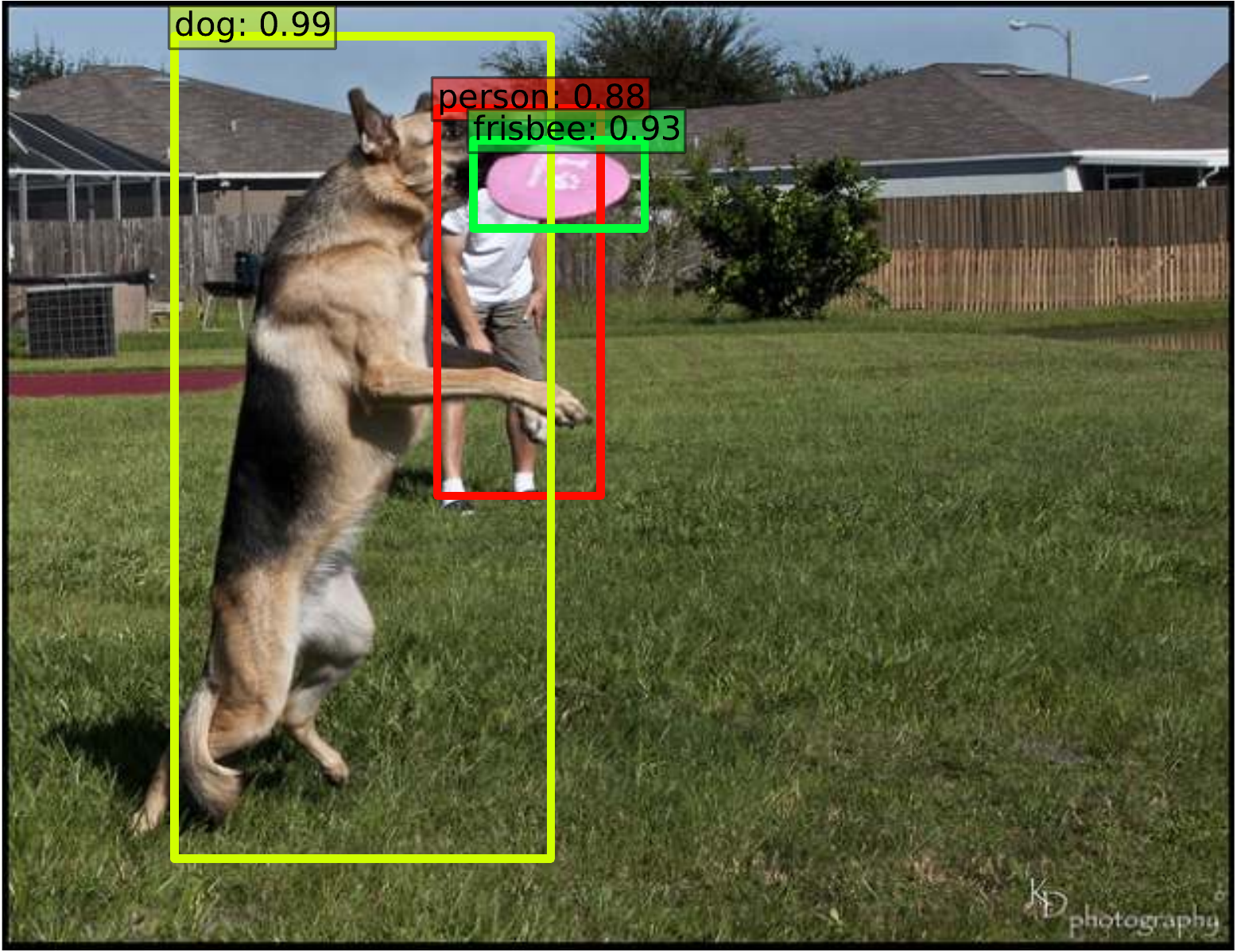}\\
	\includegraphics[trim={0 0 0 0.4cm},clip,width=0.19\linewidth]{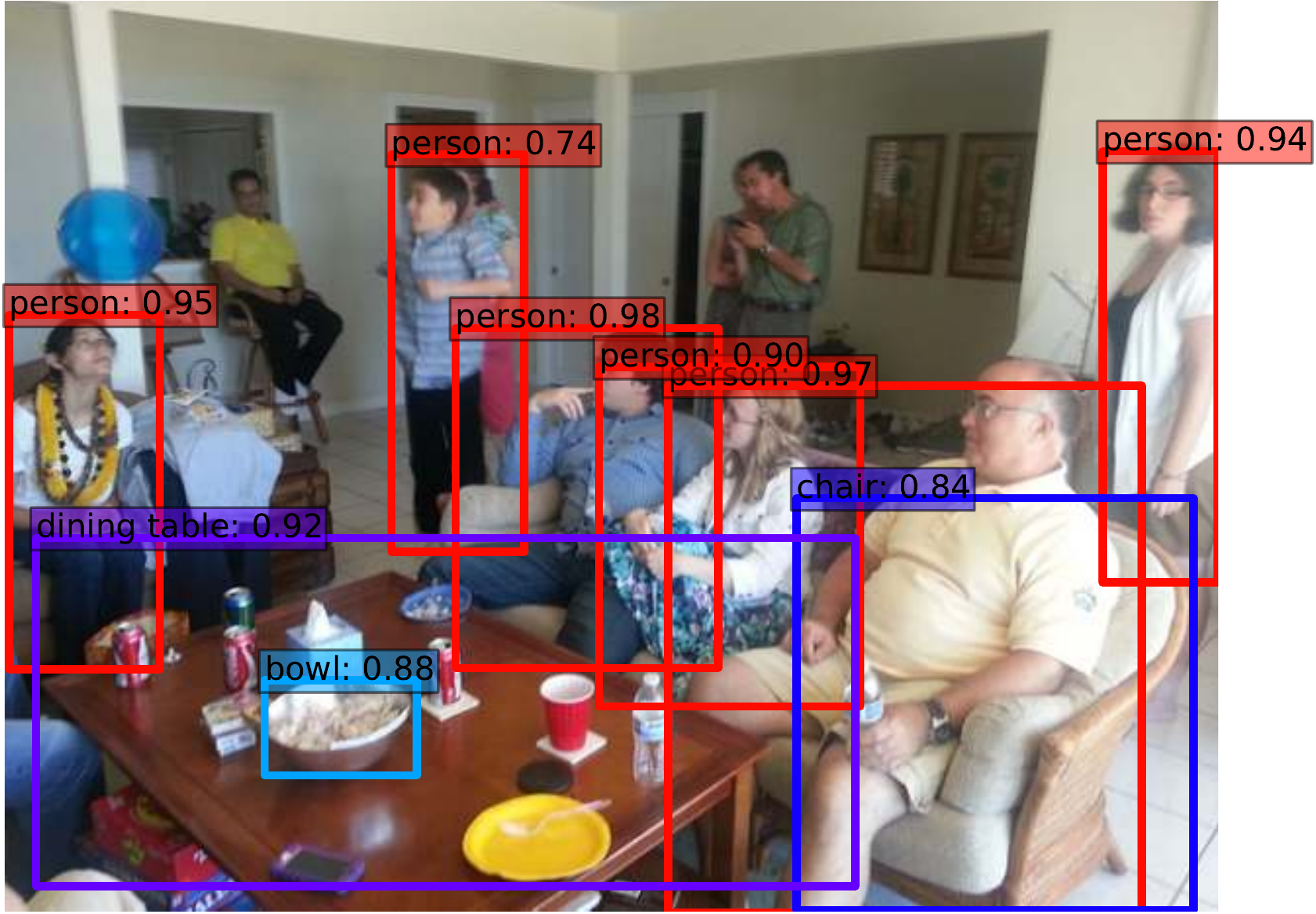}
	\includegraphics[width=0.19\linewidth]{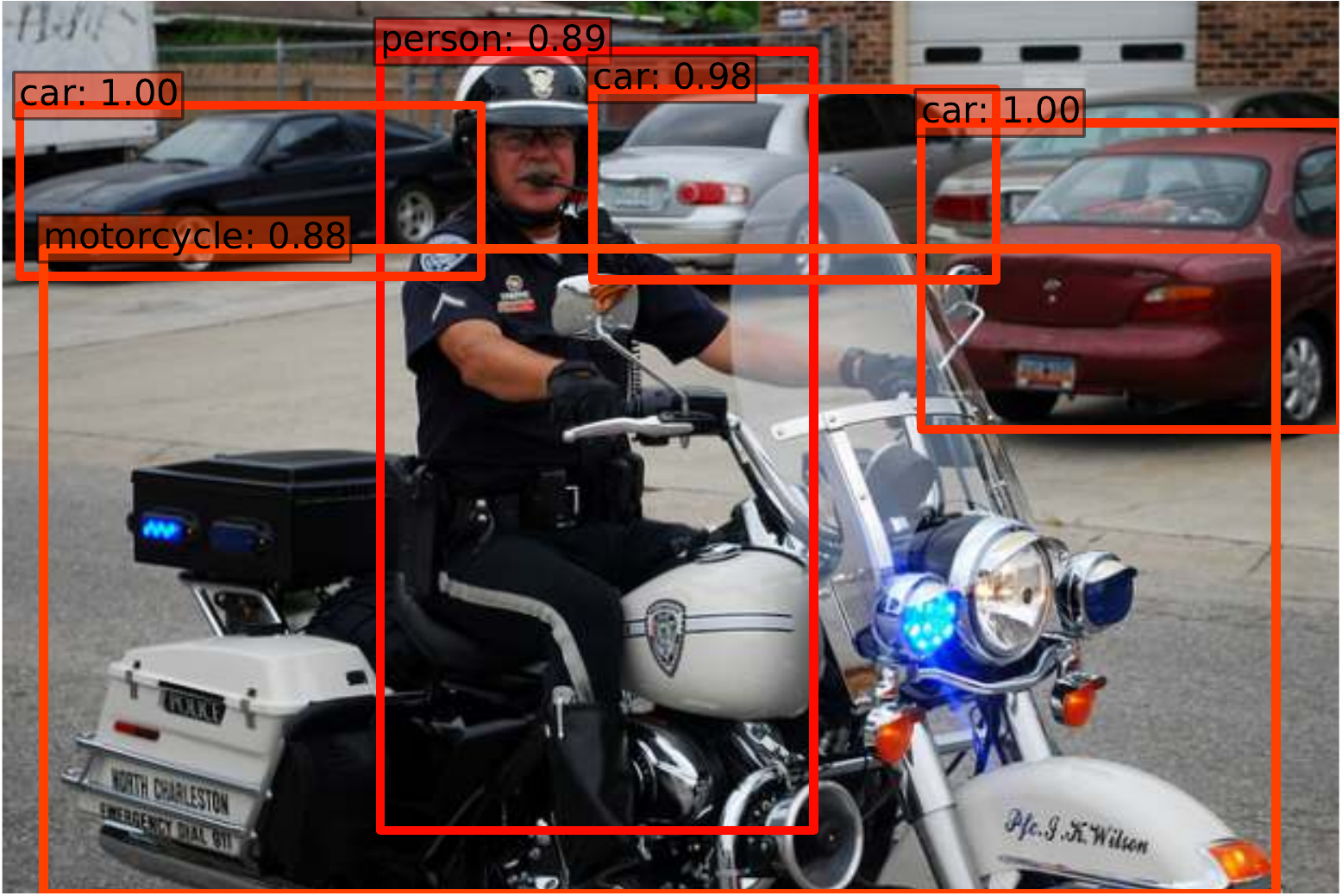}
	\includegraphics[width=0.19\linewidth]{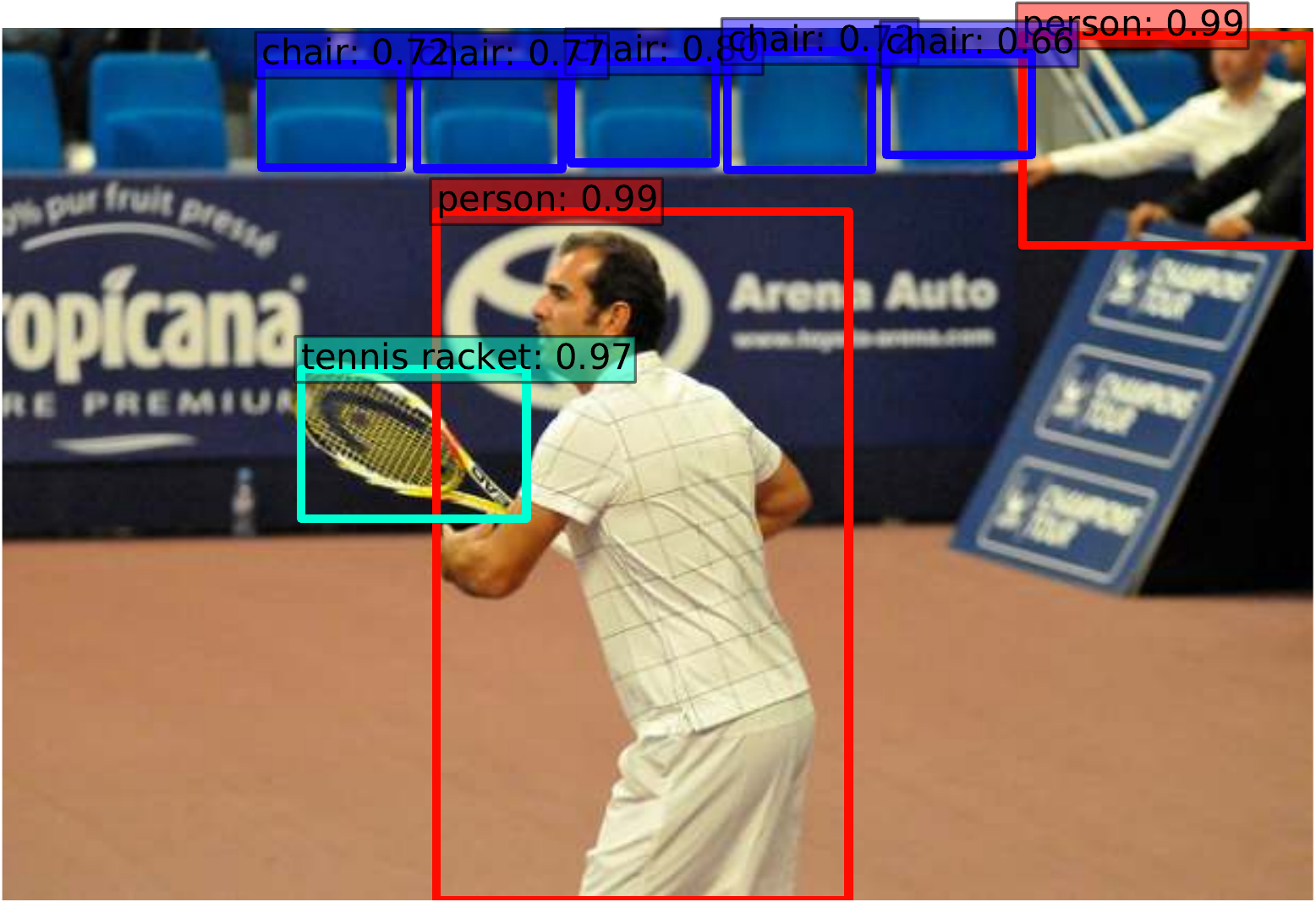}
	\includegraphics[width=0.19\linewidth]{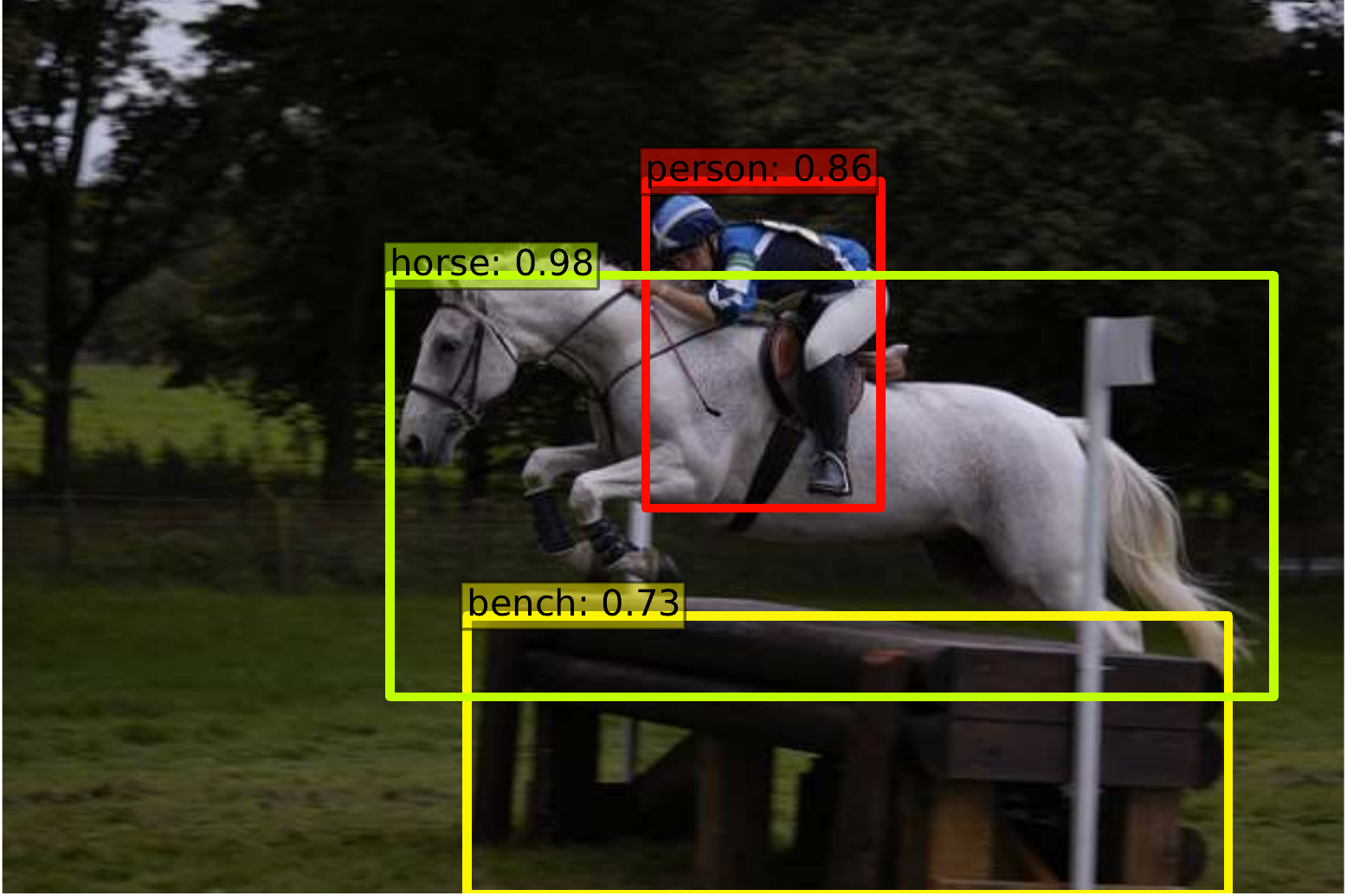}
	\includegraphics[width=0.19\linewidth]{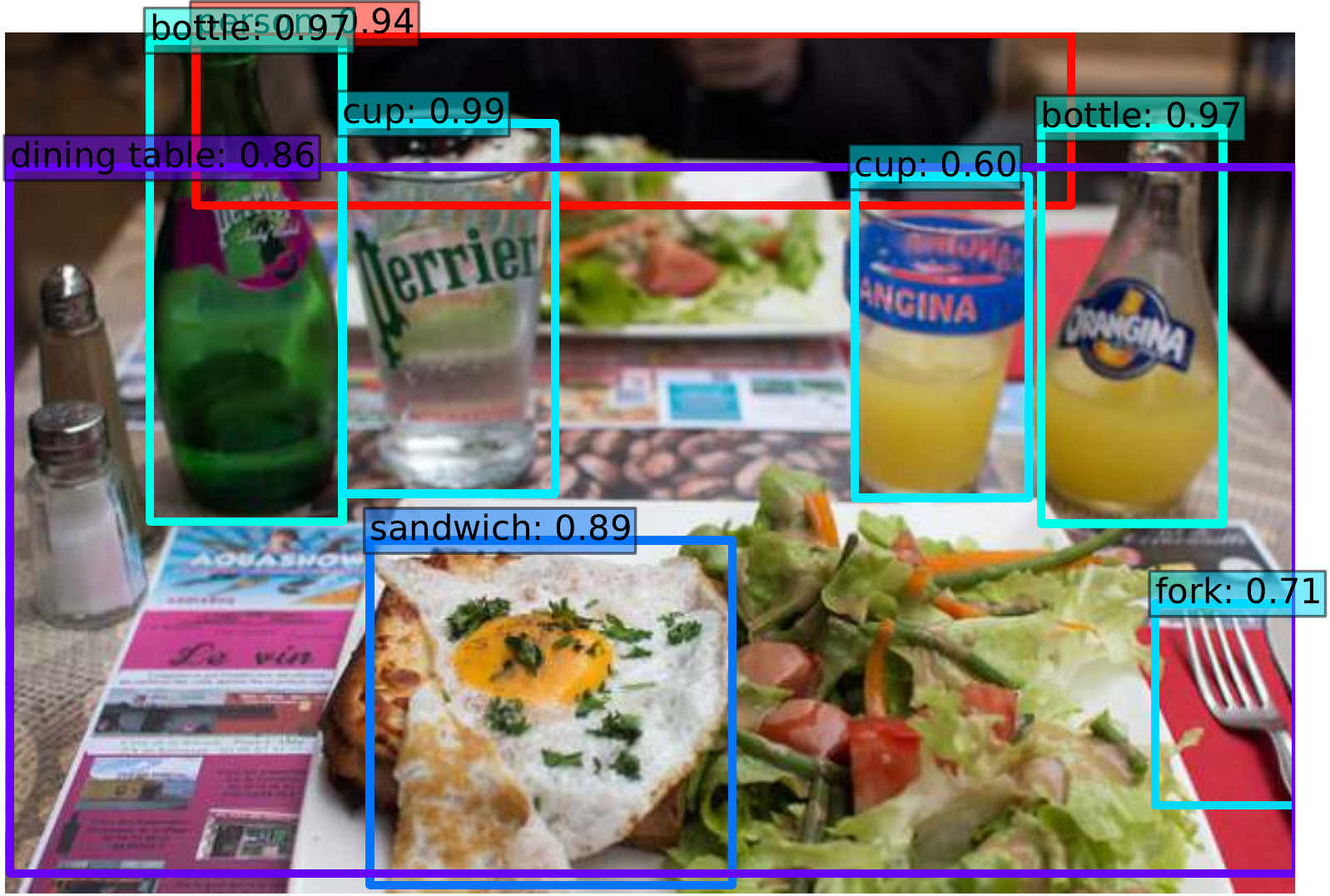}\\
	\includegraphics[trim={0 0 0 0.6cm},clip,width=0.19\linewidth]{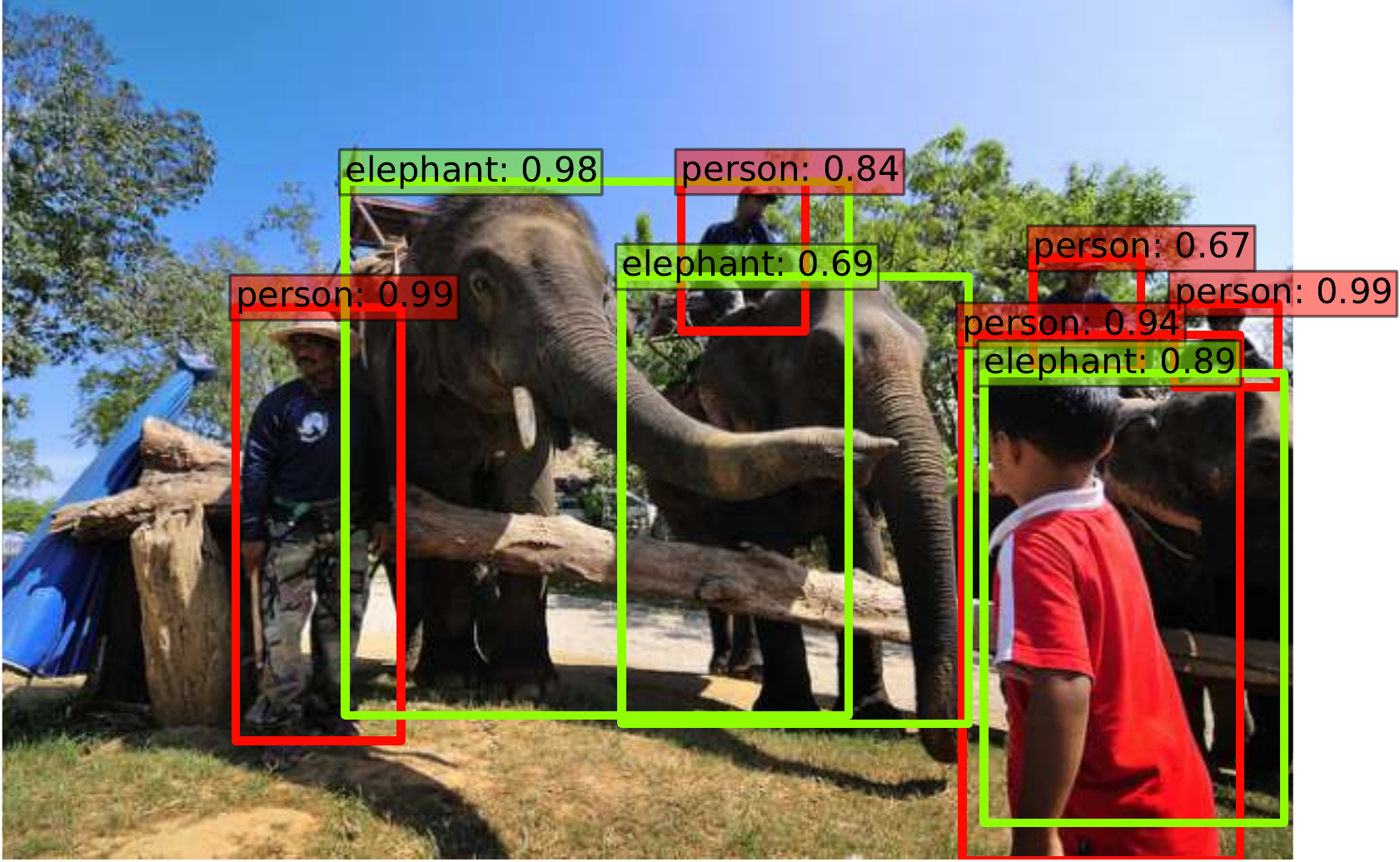}
	\includegraphics[width=0.19\linewidth]{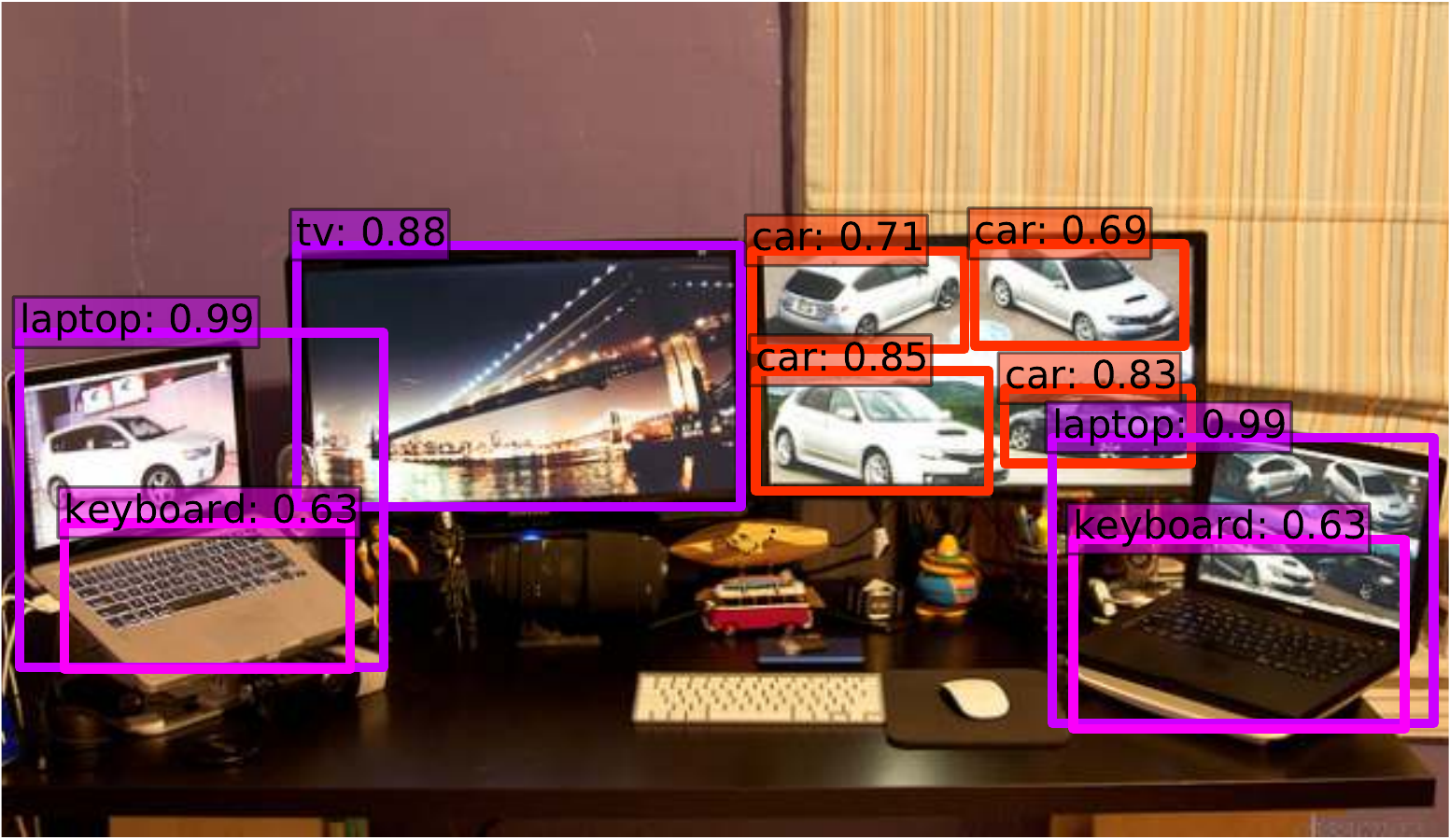}
	\includegraphics[trim={0 0 0 2.2cm},clip,width=0.19\linewidth]{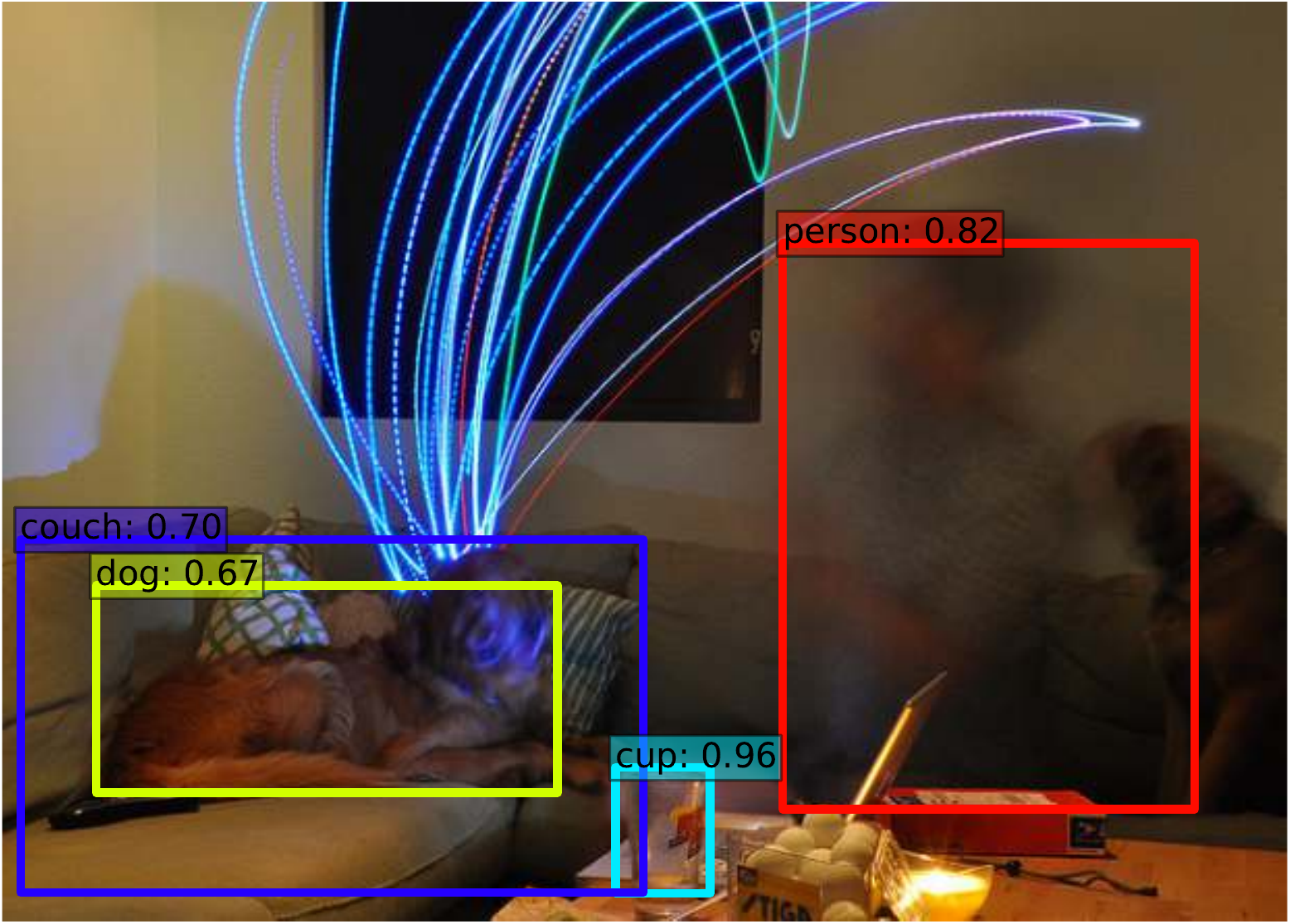}
	\includegraphics[trim={0 1.1cm 0 0},clip,width=0.19\linewidth]{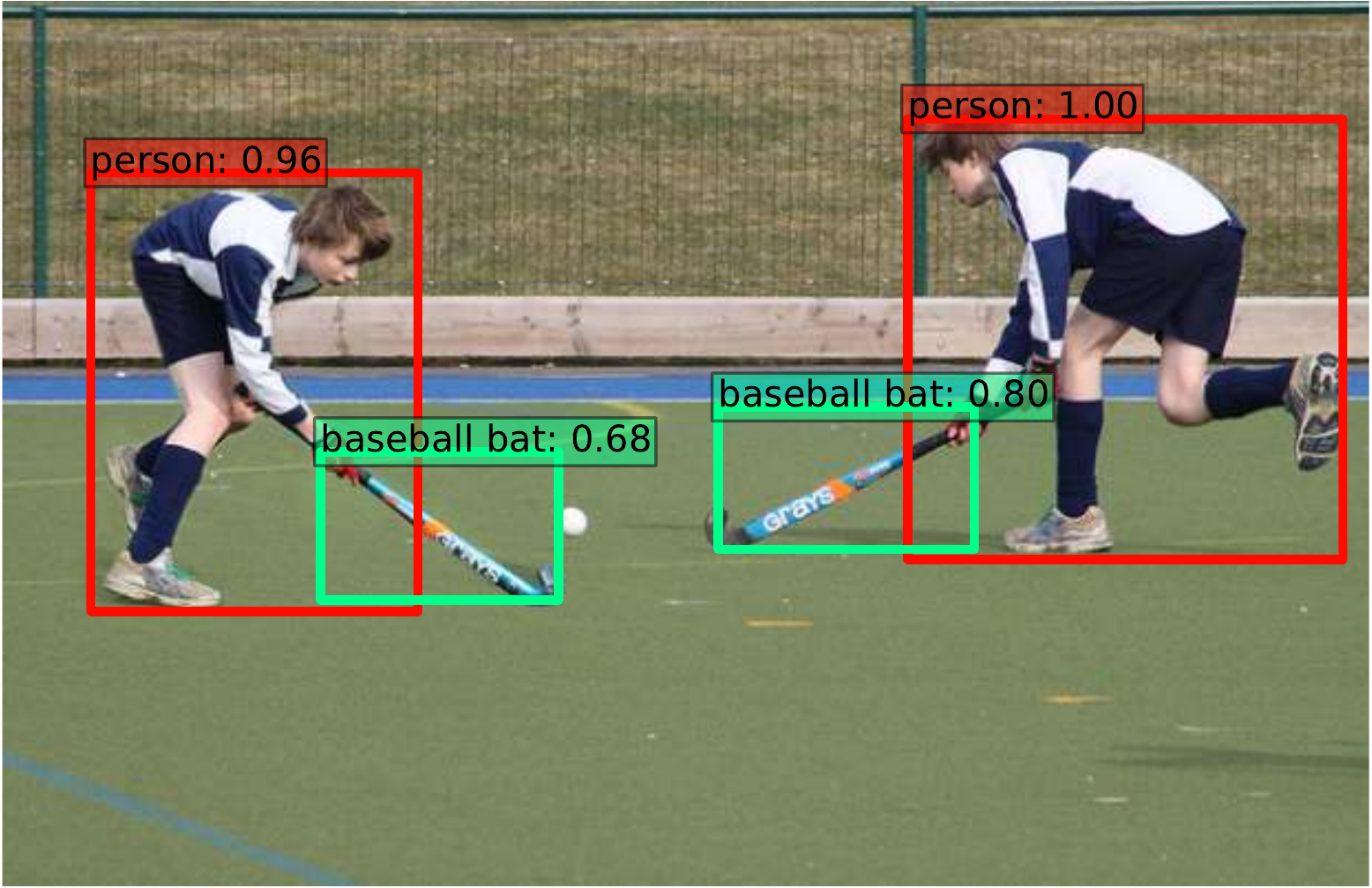}
	\includegraphics[width=0.19\linewidth]{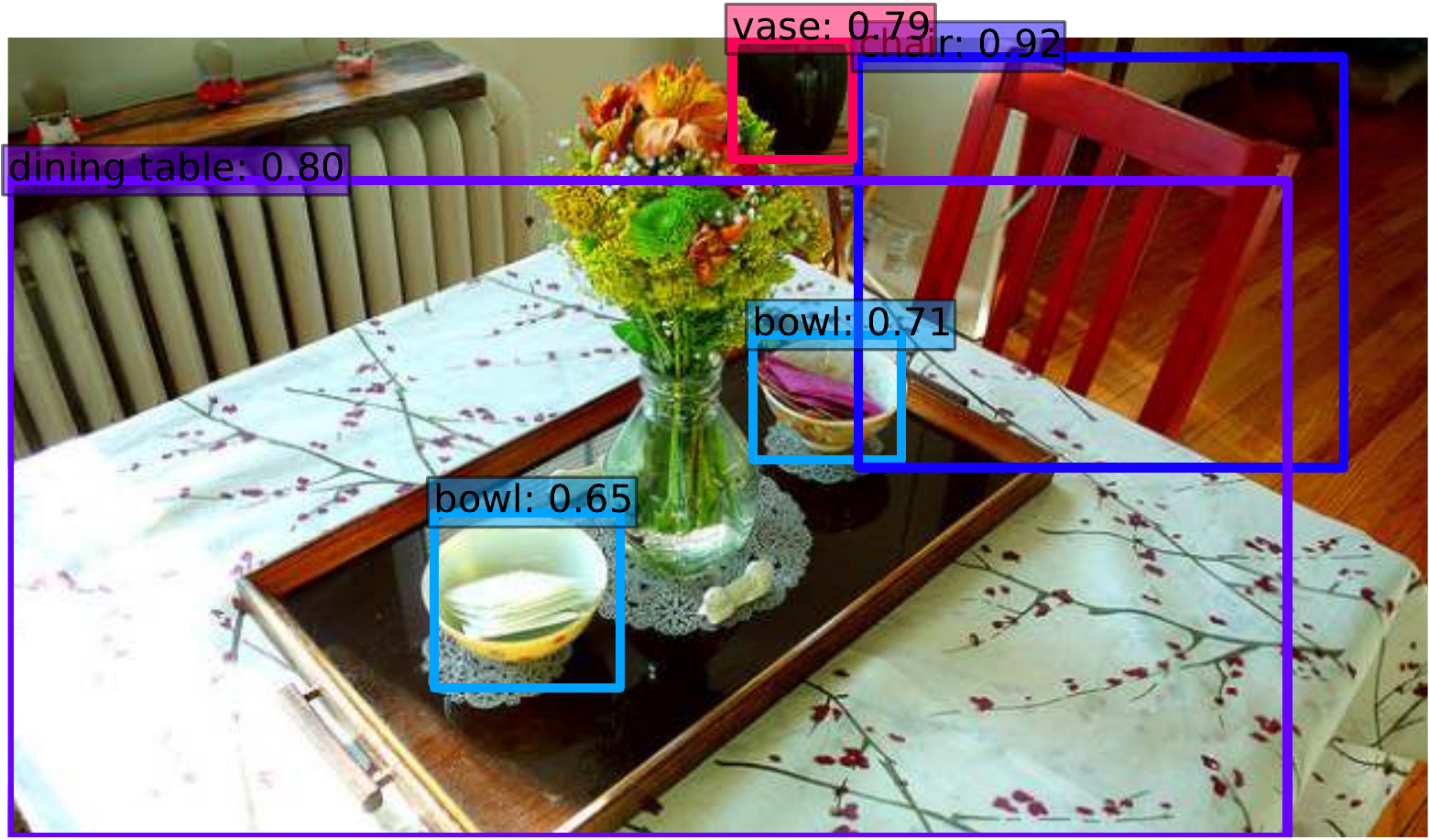}\\
	\includegraphics[width=0.19\linewidth]{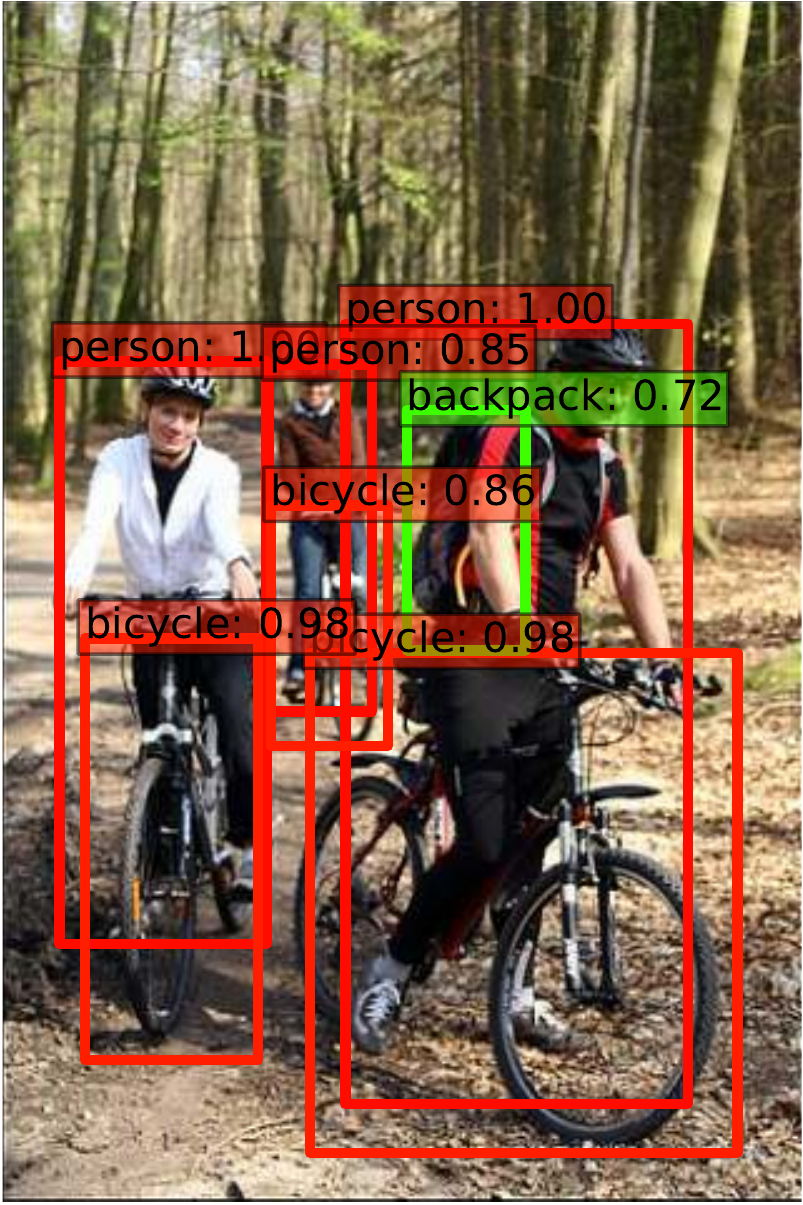}
	\includegraphics[width=0.19\linewidth]{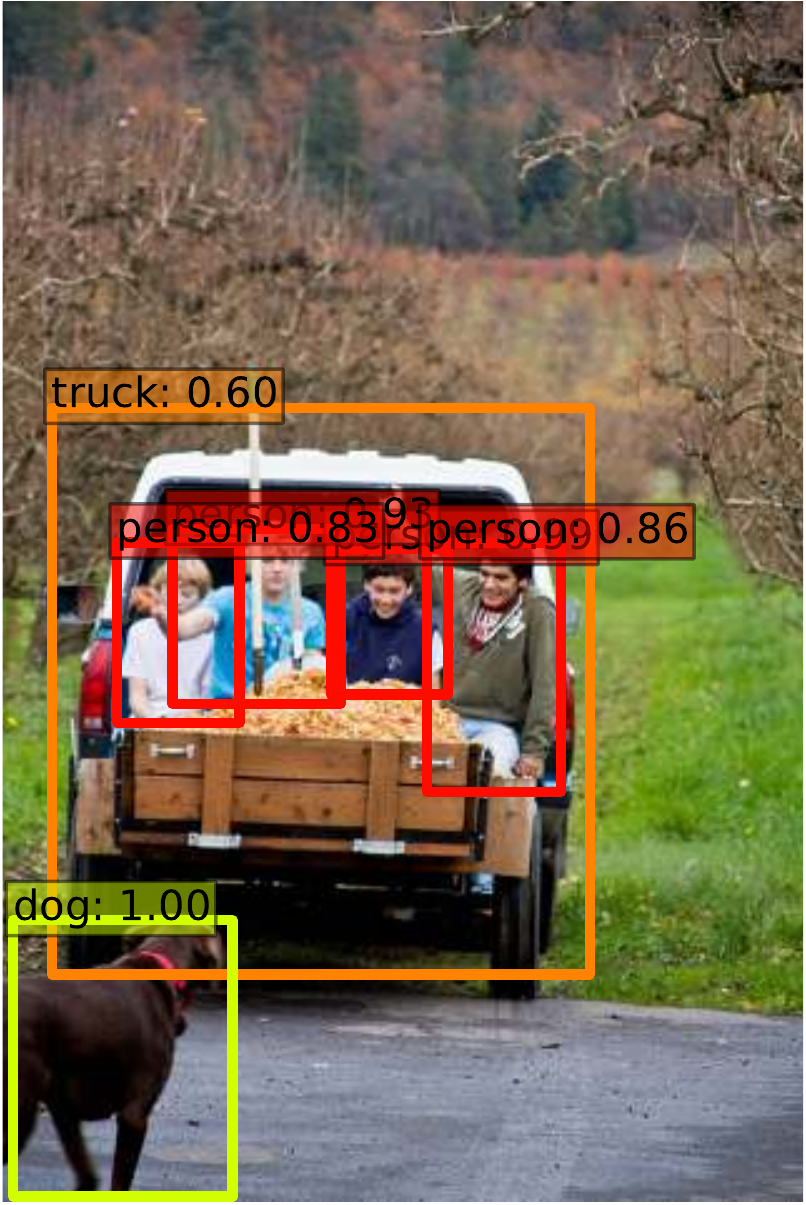}
	\includegraphics[width=0.19\linewidth]{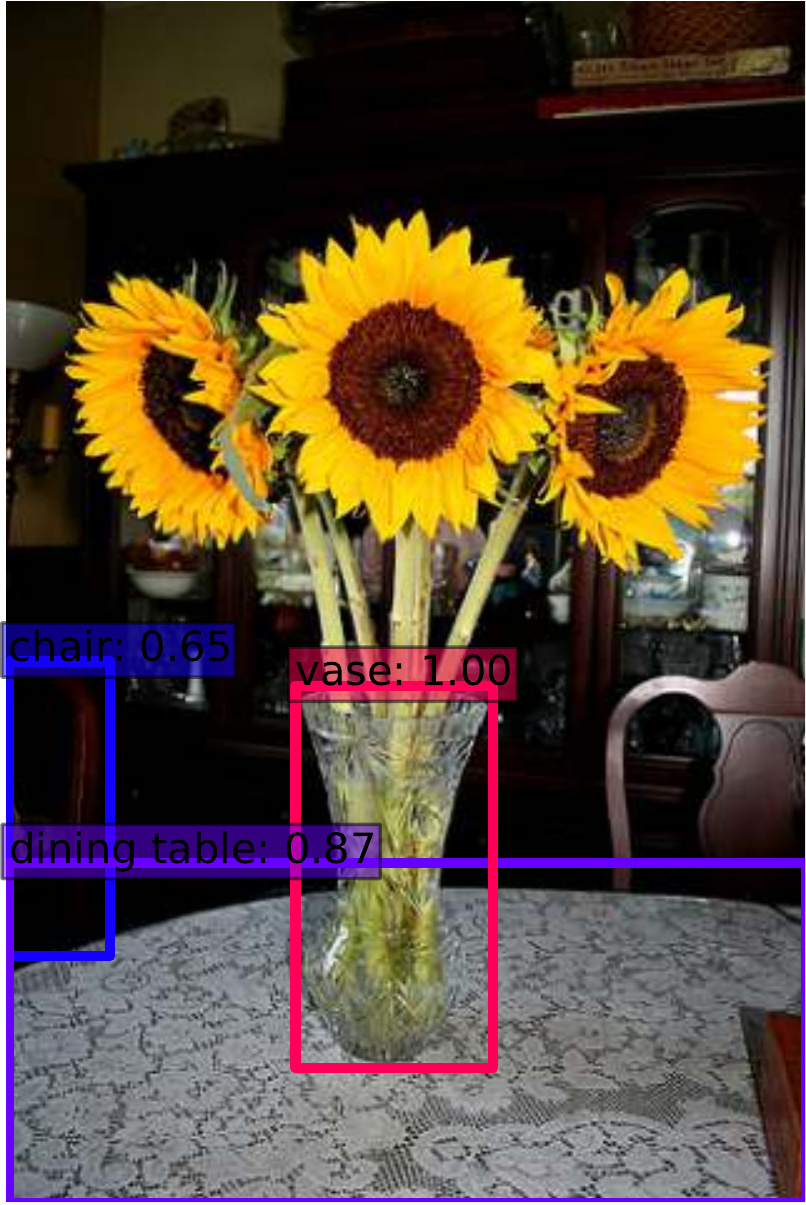}
	\includegraphics[width=0.19\linewidth]{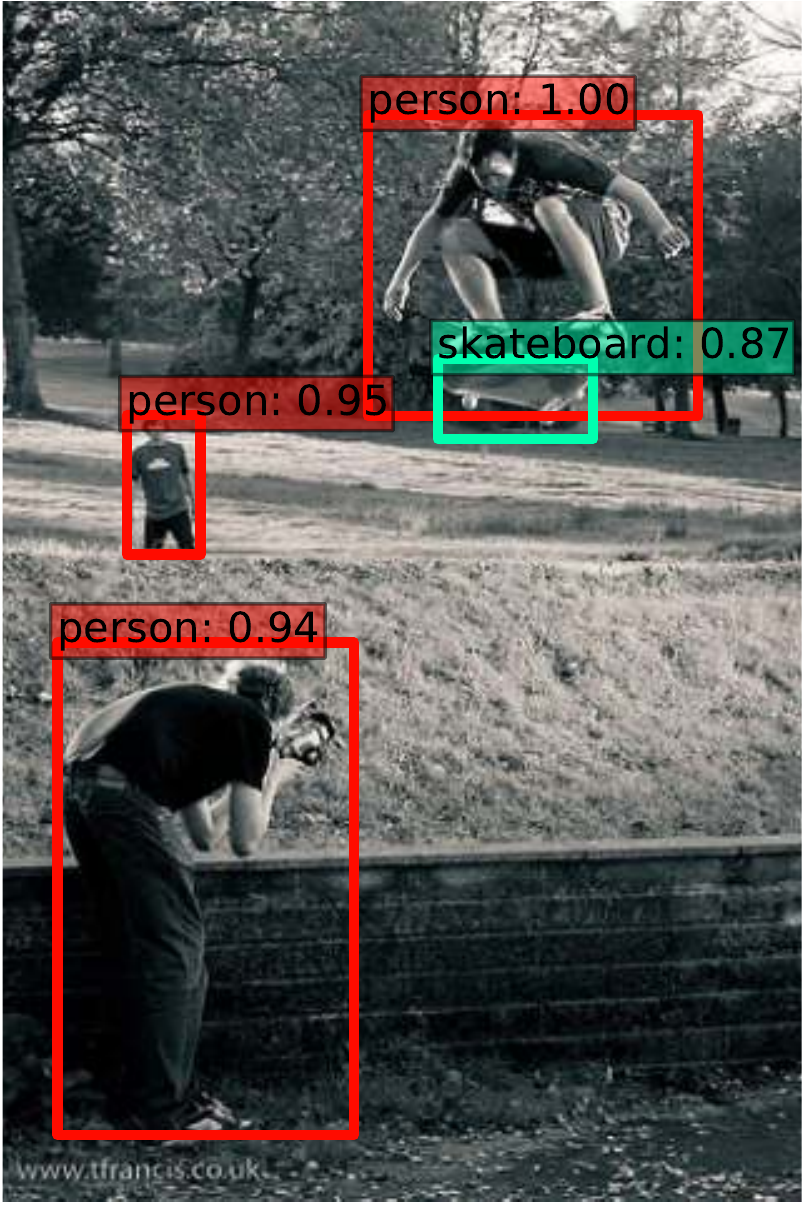}
	\includegraphics[width=0.19\linewidth]{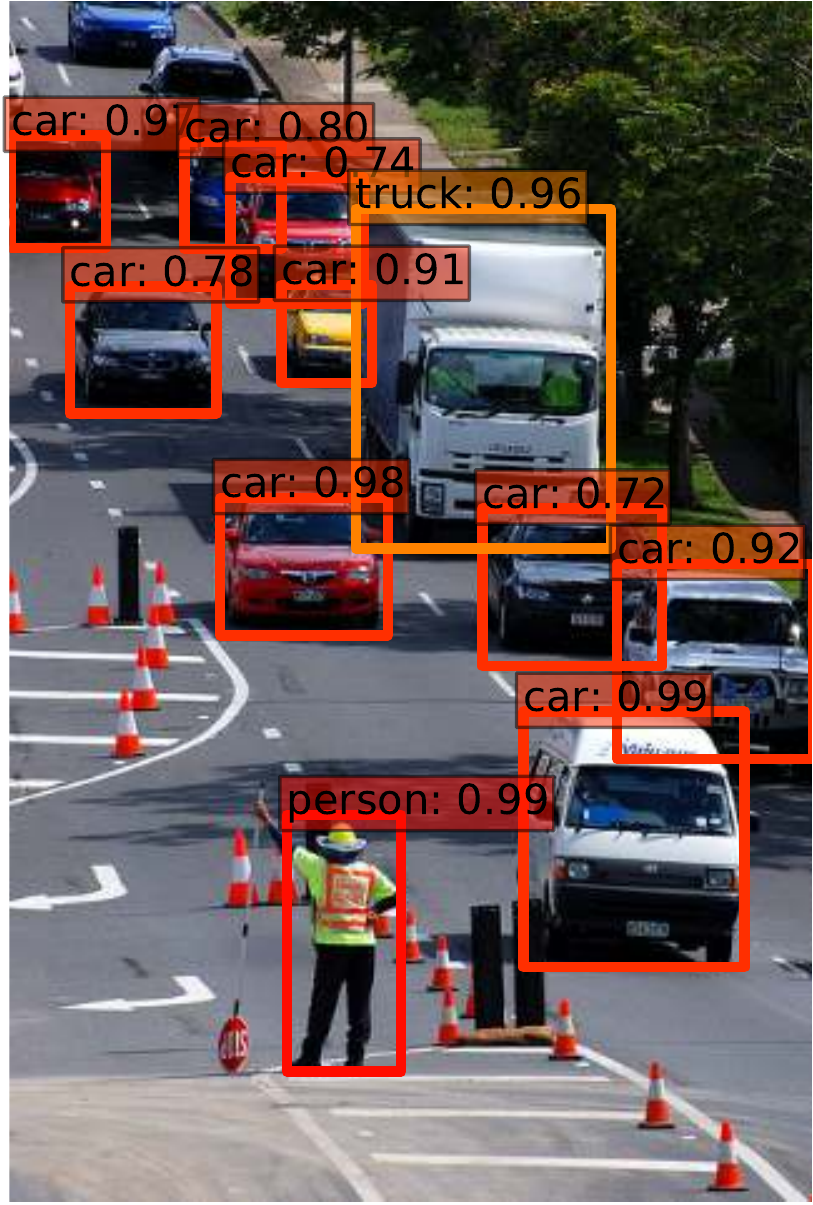}\\
	\caption{\textbf{Detection examples on COCO \texttt{test-dev} with SSD512 model.} We show detections with scores higher than 0.6. Each color corresponds to an object category.}
    \label{fig:coco}
\end{figure}

We use the \texttt{trainval35k}~\cite{bell2015inside} for training. We first train the model with $10^{-3}$ learning rate for 160k iterations, and then continue training for 40k iterations with $10^{-4}$ and 40k iterations with $10^{-5}$. Table~\ref{tab:coco} shows the results on \texttt{test-dev2015}. Similar to what we observed on the PASCAL VOC dataset, SSD300 is better than Fast R-CNN in both mAP@0.5 and mAP@[0.5:0.95]. SSD300 has a similar mAP@0.75 as ION~\cite{bell2015inside} and Faster R-CNN~\cite{cocoleaderboard}, but is worse in mAP@0.5.
%and we conjecture that it is because the image size is too small, which prevents the model from detecting many small objects. But overall, SSD can localize objects more accurately.
By increasing the image size to $512\times 512$, our SSD512 is better than Faster R-CNN~\cite{cocoleaderboard} in both criteria. Interestingly, we observe that SSD512 is 5.3\% better in mAP@0.75, but is only 1.2\% better in mAP@0.5. We also observe that it has much better AP (4.8\%) and AR (4.6\%) for large objects, but has relatively less improvement in AP (1.3\%) and AR (2.0\%) for small objects. Compared to ION, the improvement in AR for large and small objects is more similar (5.4\% vs. 3.9\%). We conjecture that Faster R-CNN is more competitive on smaller objects with SSD because it performs two box refinement steps, in both the RPN part and in the Fast R-CNN part. In Fig.~\ref{fig:coco}, we show some detection examples on COCO \texttt{test-dev} with the SSD512 model.

% \begin{table}
% 	\centering
%     \setlength{\tabcolsep}{5pt}
%     \begin{tabular}{l|c|C{2.2em}C{2.2em}C{3.5em}}
%     	\multirow{2}{*}{Method} & \multirow{2}{*}{data} & \multicolumn{3}{c}{mean Average Precision} \\
%         & & 0.5 & 0.75 & 0.5:0.95\\
%         \hline
%         Fast R-CNN~\cite{girshick2015fast} & train & 35.9 & - & 19.7\\
%         Fast R-CNN~\cite{bell2015inside} & train & 39.9 & 20.5 & 19.4\\
%         Faster R-CNN~\cite{ren2015faster} & train & 42.1 & - & 21.5\\
%         Faster R-CNN~\cite{ren2015faster} & trainval & 42.7 & - & 21.9\\
%         Faster R-CNN~\cite{cocoleaderboard} & trainval & 45.3 & 24.2 & 23.5\\
%         ION~\cite{bell2015inside} & train & 42.0 & 23.0 & 23.0\\
%         \hline
%         SSD300 & trainval35k & 41.2 & 23.2 & 23.4\\
%         SSD512 & trainval35k & \textbf{46.5} & \textbf{26.8} & \textbf{27.8}\\
% 	\end{tabular}
%     \caption{\textbf{COCO \texttt{test-dev2015} detection results.}}
%     \label{tab:coco}
% \end{table}

\subsection{Preliminary ILSVRC results}
We applied the same network architecture we used for COCO to the ILSVRC DET dataset~\cite{russakovsky2014imagenet}. We train a SSD300 model using the ILSVRC2014 DET \texttt{train} and \texttt{val1} as used in~\cite{girshick2014rich}. We first train the model with $10^{-3}$ learning rate for 320k iterations, and then continue training for 80k iterations with $10^{-4}$ and 40k iterations with $10^{-5}$. We can achieve 43.4 mAP on the \texttt{val2} set~\cite{girshick2014rich}. Again, it validates that SSD is a general framework for high quality real-time detection.

\subsection{Data Augmentation for Small Object Accuracy}
\label{sec:data-aug-new}
\begin{figure*}[htbp]
	\centering
    \includegraphics[width=0.495\linewidth]{figure/300_plots_area_strong}
    \includegraphics[width=0.495\linewidth]{figure/512_plots_area_strong}\\
    \includegraphics[width=0.495\linewidth]{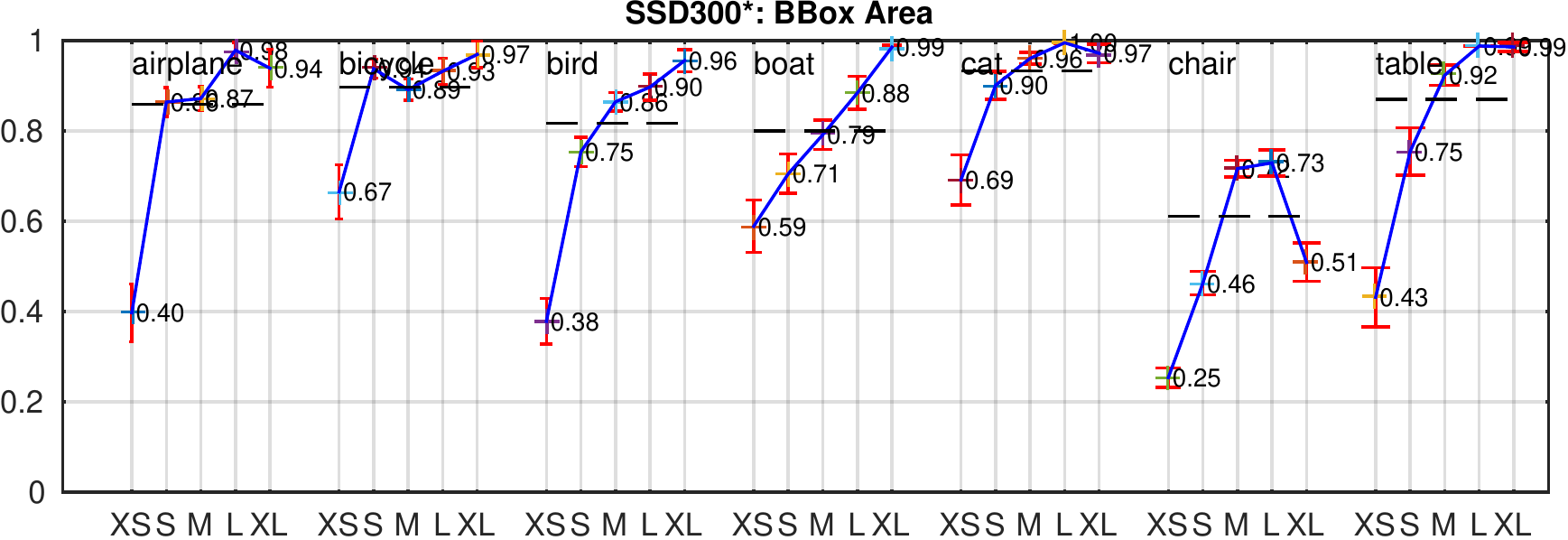}
    \includegraphics[width=0.495\linewidth]{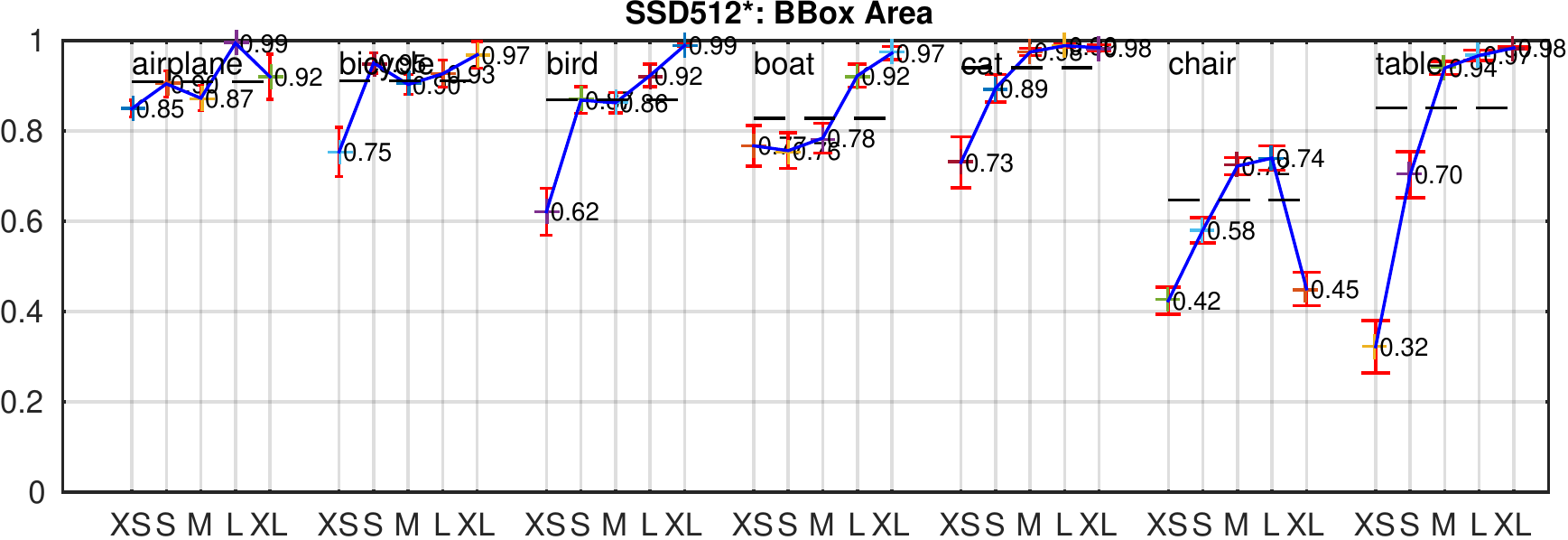}\\
    \caption{\textbf{Sensitivity and impact of object size with new data augmentation on VOC2007 \texttt{test} set using~\cite{hoiem2012diagnosing}.} The top row shows the effects of BBox Area per category for the original SSD300 and SSD512 model, and the bottom row corresponds to the SSD300* and SSD512* model trained with the new data augmentation trick. It is obvious that the new data augmentation trick helps detecting small objects significantly.}
    \label{fig:sensitivityanalysisnew}
\end{figure*}

Without a follow-up feature resampling step as in Faster R-CNN, the classification task for small objects is relatively hard for SSD, as demonstrated in our analysis (see Fig.~\ref{fig:sensitivityanalysis}).
The data augmentation strategy described in Sec.~\ref{sec:ssdtraining} helps to improve the performance dramatically, especially on small datasets such as PASCAL VOC. The random crops generated by the strategy can be thought of as a "zoom in" operation and can generate many larger training examples. To implement a "zoom out" operation that creates more small training examples, we first randomly place an image on a canvas of $16\times$ of the original image size filled with mean values before we do any random crop operation. Because we have more training images by introducing this new "expansion" data augmentation trick, we have to double the training iterations. We have seen a consistent increase of 2\%-3\% mAP across multiple datasets, as shown in Table~\ref{tab:expansion}. In specific, Figure~\ref{fig:sensitivityanalysisnew} shows that the new augmentation trick significantly improves the performance on small objects. This result underscores the importance of the data augmentation strategy for the final model accuracy. 

An alternative way of improving SSD is to design a better tiling of default boxes so that its position and scale are better aligned with the receptive field of each position on a feature map. We leave this for future work. 

\begin{table}[htbp]
	\centering
	\setlength{\tabcolsep}{1pt}
	\begin{tabular}{C{4em}|C{3.4em}C{6.2em}|C{4em}C{7em}|C{5em}C{3em}C{3em}}
    	\multirow{3}{*}{Method} & \multicolumn{2}{c|}{VOC2007 \texttt{test}} & \multicolumn{2}{c|}{VOC2012 \texttt{test}} & \multicolumn{3}{c}{COCO \texttt{test-dev2015}}\\
        & 07+12 & 07+12+COCO & 07++12 & 07++12+COCO & \multicolumn{3}{c}{trainval35k}\\
        & 0.5 & 0.5 & 0.5 & 0.5 & 0.5:0.95 & 0.5 & 0.75\\
        \hline
        SSD300 & 74.3 & 79.6 & 72.4 & 77.5 & 23.2 & 41.2 & 23.4\\
        SSD512 & 76.8 & 81.6 & 74.9 & 80.0 & 26.8 & 46.5 & 27.8\\
        \hline
        SSD300* & 77.2 & 81.2 & 75.8 & 79.3 & 25.1 & 43.1 & 25.8\\
        SSD512* & \textbf{79.8} & \textbf{83.2} & \textbf{78.5} & \textbf{82.2} & \textbf{28.8} & \textbf{48.5} & \textbf{30.3}\\
    \end{tabular}
    \caption{\textbf{Results on multiple datasets when we add the image expansion data augmentation trick.} SSD300* and SSD512* are the models that are trained with the new data augmentation.}
    \label{tab:expansion}
\end{table}

\subsection{Inference time}
Considering the large number of boxes generated from our method, it is essential to perform non-maximum suppression (nms) efficiently during inference. By using a confidence threshold of 0.01, we can filter out most boxes. We then apply nms with jaccard overlap of 0.45 per class and keep the top 200 detections per image. This step costs about 1.7 msec per image for SSD300 and 20 VOC classes, which is close to the total time (2.4 msec) spent on all newly added layers. We measure the speed with batch size 8 using Titan X and cuDNN v4 with Intel Xeon E5-2667v3@3.20GHz.
%In the future, we could potentially learn to do nms following~\cite{hosang2015convnet} to further speed up, especially for large number of classes.

Table~\ref{tab:inference} shows the comparison between SSD, Faster R-CNN\cite{ren2015faster}, and YOLO\cite{redmon2015you}. Both our SSD300 and SSD512 method outperforms Faster R-CNN in both speed and accuracy. Although Fast YOLO\cite{redmon2015you} can run at 155 FPS, it has lower accuracy by almost 22\% mAP. To the best of our knowledge, SSD300 is the first real-time method to achieve above 70\% mAP. Note that about 80\% of the forward time is spent on the base network (VGG16 in our case). Therefore, using a faster base network could even further improve the speed, which can possibly make the SSD512 model real-time as well.
\begin{table}
	\centering
    \setlength{\tabcolsep}{5pt}
    \begin{tabular}{l|c|c|c|c|c}
		Method & mAP & FPS & batch size & \# Boxes & Input resolution \\
        \hline
        Faster R-CNN (VGG16) & 73.2 & 7 & 1 & $\sim6000$ & $\sim1000\times600$\\
        \hline 
        Fast YOLO & 52.7 & 155 & 1 & 98 & $448\times448$\\
        YOLO (VGG16) & 66.4 & 21 & 1 & 98 & $448\times448$\\
        \hline
        SSD300 & 74.3 & 46 & 1 & 8732 & $300\times300$ \\
        SSD512 & 76.8 & 19 & 1 & 24564 & $512\times512$ \\
        SSD300 & 74.3 & 59 & 8 & 8732 & $300\times300$ \\
        SSD512 & 76.8 & 22 & 8 & 24564 & $512\times512$ \\
	\end{tabular}
    \caption{\textbf{Results on Pascal VOC2007 \texttt{test}.} SSD300 is the only real-time detection method that can achieve above 70\% mAP. By using a larger input image, SSD512 outperforms all methods on accuracy while maintaining a close to real-time speed.}
    \label{tab:inference}
\end{table}

%------------------------------------------------------------------------
\section{Related Work}
\label{sec:relatedwork}
There are two established classes of methods for object detection in images, one based on sliding windows and the other based on region proposal classification. Before the advent of convolutional neural networks, the state of the art for those two approaches -- Deformable Part Model (DPM)~\cite{felzenszwalb2008discriminatively} and Selective Search~\cite{uijlings2013selective} -- had comparable performance. However, after the dramatic improvement brought on by R-CNN~\cite{girshick2014rich}, which combines selective search region proposals and convolutional network based post-classification, region proposal object detection methods became prevalent.

The original R-CNN approach has been improved in a variety of ways. The first set of approaches improve the quality and speed of post-classification, since it requires the classification of thousands of image crops, which is expensive and time-consuming. SPPnet~\cite{he2014spatial} speeds up the original R-CNN approach significantly. It introduces a spatial pyramid pooling layer that is more robust to region size and scale and allows the classification layers to reuse features computed over feature maps generated at several image resolutions. Fast R-CNN~\cite{girshick2015fast} extends SPPnet so that it can fine-tune all layers end-to-end by minimizing a loss for both confidences and bounding box regression, which was first introduced in MultiBox~\cite{erhan2014scalable} for learning objectness.

The second set of approaches improve the quality of proposal generation using deep neural networks. In the most recent works like MultiBox~\cite{erhan2014scalable,szegedy2014scalable}, the Selective Search region proposals, which are based on low-level image features, are replaced by proposals generated directly from a separate deep neural network. This further improves the detection accuracy but results in a somewhat complex setup, requiring the training of two neural networks with a dependency between them. Faster R-CNN~\cite{ren2015faster} replaces selective search proposals by ones learned from a region proposal network (RPN), and introduces a method to integrate the RPN with Fast R-CNN by alternating between fine-tuning shared convolutional layers and prediction layers for these two networks.  This way region proposals are used to pool mid-level features and the final classification step is less expensive.  Our SSD is very similar to the region proposal network (RPN) in Faster R-CNN in that we also use a fixed set of (default) boxes for prediction, similar to the anchor boxes in the RPN.  But instead of using these to pool features and evaluate another classifier, we simultaneously produce a score for each object category in each box. Thus, our approach avoids the complication of merging RPN with Fast R-CNN and is easier to train, faster, and straightforward to integrate in other tasks.

Another set of methods, which are directly related to our approach, skip the proposal step altogether and predict bounding boxes and confidences for multiple categories directly. OverFeat~\cite{sermanet2013overfeat}, a deep version of the sliding window method, predicts a bounding box directly from each location of the topmost feature map after knowing the confidences of the underlying object categories. YOLO~\cite{redmon2015you} uses the whole topmost feature map to predict both confidences for multiple categories and bounding boxes (which are shared for these categories). Our SSD method falls in this category because we do not have the proposal step but use the default boxes. However, our approach is more flexible than the existing methods because we can use default boxes of different aspect ratios on each feature location from multiple feature maps at different scales. If we only use one default box per location from the topmost feature map, our SSD would have similar architecture to OverFeat~\cite{sermanet2013overfeat}; if we use the whole topmost feature map and add a fully connected layer for predictions instead of our convolutional predictors, and do not explicitly consider multiple aspect ratios, we can approximately reproduce YOLO~\cite{redmon2015you}.

\section{Conclusions}
\label{sec:futurework}
This paper introduces SSD, a fast single-shot object detector for multiple categories. A key feature of our model is the use of multi-scale convolutional bounding box outputs attached to multiple feature maps at the top of the network. This representation allows us to efficiently model the space of possible box shapes. We experimentally validate that given appropriate training strategies, a larger number of carefully chosen default bounding boxes results in improved performance. We build SSD models with at least an order of magnitude more box predictions sampling location, scale, and aspect ratio, than existing methods~\cite{redmon2015you,erhan2014scalable}. We demonstrate that given the same VGG-16 base architecture, SSD compares favorably to its state-of-the-art object detector counterparts in terms of both accuracy and speed.  Our SSD512 model significantly outperforms the state-of-the-art Faster R-CNN~\cite{ren2015faster} in terms of accuracy on PASCAL VOC and COCO, while being $3\times$ faster. Our real time SSD300 model runs at 59 FPS, which is faster than the current real time YOLO~\cite{redmon2015you} alternative, while producing markedly superior detection accuracy.

Apart from its standalone utility, we believe that our monolithic and relatively simple SSD model provides a useful building block for larger systems that employ an object detection component. A promising future direction is to explore its use as part of a system using recurrent neural networks to detect and track objects in video simultaneously.

\section{Acknowledgment}
\noindent This work was started as an internship project at Google and continued at UNC. We would like to thank Alex Toshev for helpful discussions and are indebted to the Image Understanding and DistBelief teams at Google. We also thank Philip Ammirato and Patrick Poirson for helpful comments. We thank NVIDIA for providing GPUs and acknowledge support from NSF 1452851, 1446631, 1526367, 1533771.
\bibliographystyle{splncs}
\bibliography{ssd}

\end{document}